\documentclass[11pt,letterpaper]{article}
\usepackage{caption}
\usepackage{subcaption}
\usepackage{fullpage}
\usepackage{times}
\usepackage{array}
\usepackage{ragged2e}
\usepackage{multirow}
\usepackage{fancyhdr, amssymb}
\usepackage[ruled,vlined]{algorithm2e}
\include{pythonlisting}
\usepackage[dvipsnames]{xcolor}
\usepackage[margin = 1 in]{geometry}
\usepackage{graphicx}
\usepackage{outlines}
\usepackage{amsmath}
\DeclareMathOperator*{\argmax}{argmax}

\usepackage{graphicx}

\usepackage[english]{babel} 
\usepackage{blindtext}
\usepackage{xltabular}
\usepackage{arydshln} 
\usepackage{wrapfig}
\usepackage{enumitem}
\usepackage{floatrow}
\usepackage{xfrac}
\usepackage[font={footnotesize}, labelfont={bf}, labelsep=space, justification=justified, skip=0pt]{caption}
\usepackage[hang,flushmargin]{footmisc} 
\usepackage[colorlinks=true, allcolors=blue]{hyperref}

\usepackage[
backend=biber,
style=numeric-comp,
sorting=none,
maxnames = 10
]{biblatex}
\newcommand{\cmt}[1]{} 
\newcommand{\fbest}{$y^*$}
\newcommand{\fobj}{$f(\boldsymbol{x})$}
\newcommand{\yobj}{$y(\boldsymbol{x}) $}

\newcommand{\MEI}{\texttt{LMGP\textsubscript{EI}}}
\newcommand{\MPI}{\texttt{LMGP\textsubscript{PI}}}
\newcommand{\BOT}{\texttt{BoTorch}}
\newcommand{\LMGP}{\texttt{LMGP\textsubscript{CA}}}

\newcommand{\potential}{\texttt{Double-well Potential}}
\newcommand{\rosen}{\texttt{Rosenbrock}}
\newcommand{\borehole}{\texttt{Borehole}}
\newcommand{\wing}{\texttt{Wing}}
\newcommand{\nta}{\texttt{NTA}}
\newcommand{\hoip}{\texttt{HOIP}}

\title{Multi-Fidelity Cost-Aware Bayesian Optimization}
\date{\vspace{-5ex}}
\usepackage{authblk}
\author[1]{Zahra Zanjani Foumani}
\author[1]{Mehdi Shishehbor}
\author[1]{Amin Yousefpour}
\author[1]{Ramin Bostanabad \thanks{\noindent Corresponding Author: Raminb@uci.edu \\\href{https://gitlab.com/S3anaz/multi-fidelity-cost-aware-bayesian-optimization/-/tree/main/}{Gitlab repository: https://gitlab.com/S3anaz/multi-fidelity-cost-aware-bayesian-optimization/-/tree/main/}}}
\affil[1]{Department of Mechanical and Aerospace Engineering, University of California, Irvine}
 
\begin{document}
    \pagenumbering{arabic}
    \sloppy
    \maketitle
    
    \noindent \textbf{Abstract}\\
Bayesian optimization (BO) is increasingly employed in critical applications such as materials design and drug discovery. An increasingly popular strategy in BO is to forgo the sole reliance on high-fidelity data and instead use an ensemble of information sources which provide inexpensive low-fidelity data. The overall premise of this strategy is to reduce the overall sampling costs by querying inexpensive low-fidelity sources whose data are correlated with high-fidelity samples. Here, we propose a multi-fidelity cost-aware BO framework that dramatically outperforms the state-of-the-art technologies in terms of efficiency, consistency, and robustness. We demonstrate the advantages of our framework on analytic and engineering problems and argue that these benefits stem from our two main contributions: $(1)$ we develop a novel acquisition function for multi-fidelity cost-aware BO that safeguards the convergence against the biases of low-fidelity data, and $(2)$ we tailor a newly developed emulator for multi-fidelity BO which enables us to not only simultaneously learn from an ensemble of multi-fidelity datasets, but also identify the severely biased low-fidelity sources that should be excluded from BO.  

    \section{Introduction} \label{sec: intro}
Bayesian optimization (BO) is an iterative and sample-efficient global optimization technique that has been successfully applied to a wide range of applications including materials discovery \cite{griffiths2020constrained, RN1375, RN1424, RN1794}, design of chemical systems such as catalysts \cite{ebikade2020active}, hyperparameter tuning in machine learning (ML) models \cite{snoek2012practical}, robot motion control \cite{burger2020mobile}, and updating internet-scale software systems \cite{matsubara2016data}. 
While BO is very effective, the total optimization cost can still be high if only an expensive source (e.g., experiments or costly simulations) is sampled during the optimization. To reduce the overall data collection costs in such scenarios an increasingly popular strategy is to formulate \textit{multi-fidelity} (MF) methods that use multiple data sources which typically have different levels of accuracy and cost. Assuming low-fidelity (LF) sources are cheaper to query, the overall premise of these methods is to reduce the total sampling costs by leveraging the correlations between low- and high-fidelity (HF) data.
In this paper, we propose a \textit{multi-fidelity cost-aware (MFCA) BO framework} that optimizes an expensive objective function using an ensemble of data sources with arbitrary levels of accuracy and cost. We provide a new perspective on probabilistic learning from multiple sources which endows our framework with five major advantages over existing MF BO techniques: $(a)$ safeguarding the convergence against the biases of the LF sources even if they are extremely inexpensive to query (i.e., if the majority of the samples are LF), $(b)$ learning the relative fidelity of the sources rather than requiring a priori determination of such relations by the user, $(c)$ dispensing with the assumptions that aim to relate the fidelity and cost of a data source, and $(d)$ improving numerical stability and efficiency.

\begin{figure*}[!b] 
    \centering
    \includegraphics[page=1, width = 1\textwidth]{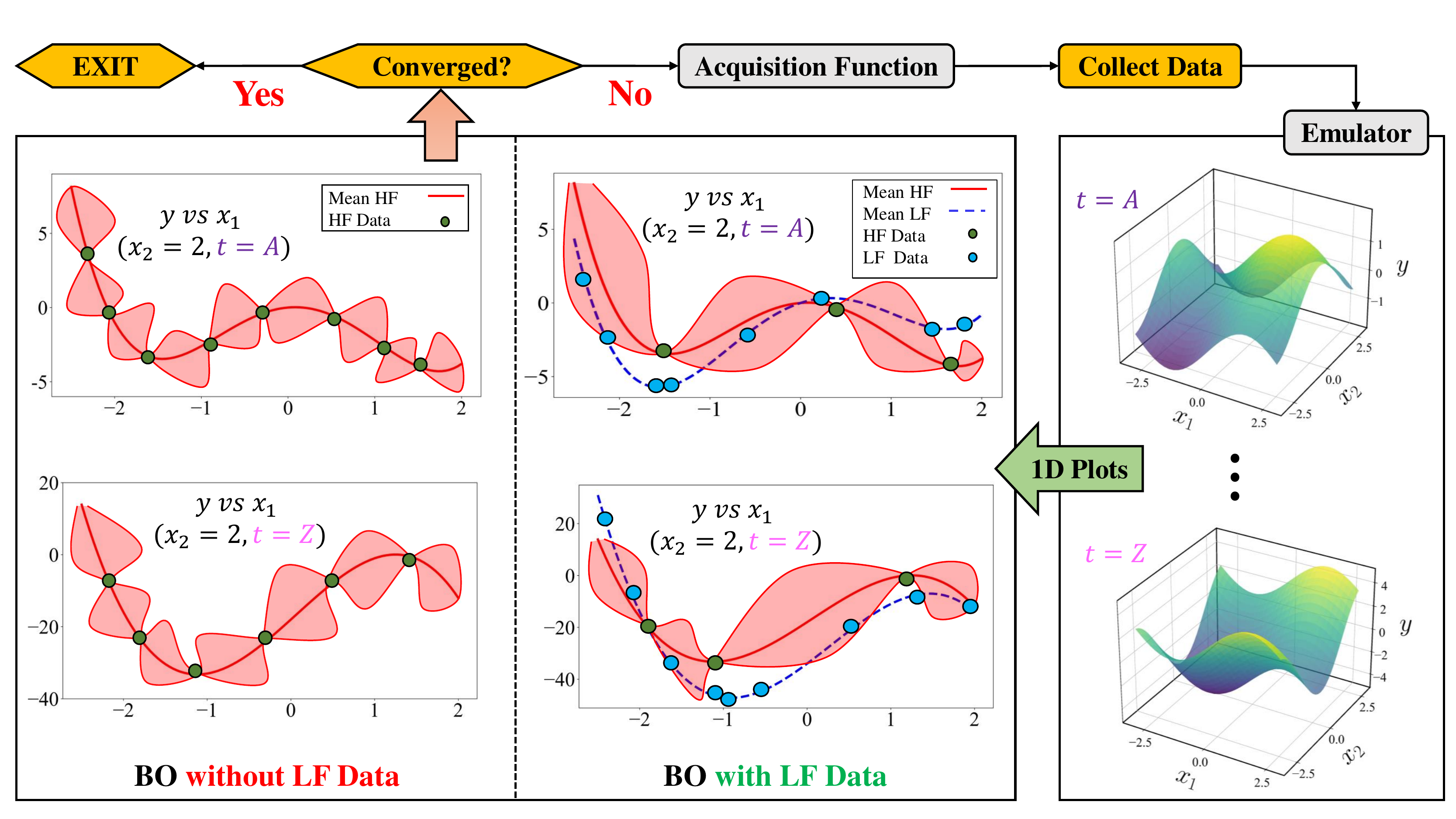}
    \vspace{-0.3cm}
    \caption{\textbf{Bi-fidelity BO:} The left box of the figure shows two different scenarios (\textcolor{red}{without LF data} and \textcolor{ForestGreen}{with LF Data}) for a function with two numerical inputs ($x_1$ , $x_2$) and one categorical input ($t$) that has two levels ($A$ and $Z$). In each scenario, $x_2$ is fixed to simplify the visualizations and the only difference between the upper and lower plots is the level of $t$. In the left panel only HF data is used in BO while in the right panel two data sources are available where one of them has a lower fidelity and sampling cost. In the multi-fidelity case, the correlations between the data sources are used to decrease the overall sampling costs by reducing the reliance on the HF source.}
    \label{fig: bo}
\end{figure*}

As schematically demonstrated in Figure \ref{fig: bo} and detailed in Section \ref{sec: background-bo}, BO has two main ingredients that interact sequentially to optimize a black-box and expensive-to-evaluate objective function. These two ingredients include an emulator (i.e., a probabilistic surrogate) and an acquisition function (AF). The optimization process starts by fitting the emulator to a small initial dataset that is typically obtained via design of experiments. The emulator is next used in the AF to determine the candidate input(s) whose corresponding output(s) must be obtained by querying the expensive function. Given the new sample(s), the training dataset is updated and the entire fitting-searching-sampling process is repeated until a convergence criterion is met (e.g., resources are exhausted). 

While many emulators such as Bayesian neural networks are available, Gaussian processes (GPs) are typically used in BO since they very efficiently learn from small data, are easy and fast to train, provide prediction uncertainties, and have interpretable parameters \cite{RN332, RN434, RN681, RN649}. Use of GPs in BO has increased even more because of the recent works that enable them to handle categorical variables \cite{RN1075, RN1214, RN1215}, high-dimensional inputs \cite{RN1228, RN1255}, large datasets \cite{RN1573, RN1479, RN1270, RN893}, and non-stationary noise \cite{RN1788, RN340}. As for the AF, there are many choices \cite{RN1392, RN1393} such as expected improvement (EI), probability of improvement (PI), and knowledge gradient (KG). The primary difference among these AFs is that they select the candidate input(s) while taking different approaches for balancing exploitation (i.e., sampling based on the best predictions of the emulator) and exploration (i.e., sampling to reduce prediction uncertainty). This selection involves integration which can sometimes be analytically computed (e.g., when GP and EI are chosen as the emulator and AF, respectively). 

To optimally use an ensemble of information sources in BO, two conditions must be met: the emulator should leverage the cross-source correlations (which are hidden in the datasets) to more accurately surrogate \textit{all} the data sources (esp. the HF one) and the AF should appropriately calculate the value or \textit{utility} of a to-be-sampled data point based on its source and evaluation cost. Satisfying these two conditions in many realistic applications is nontrivial for the following reasons:
\begin{itemize}
    \item The global optima (input and output) of LF sources differ from those of the HF source, see Figure \ref{fig: 1D-approach} for a one dimensional illustration.
    \item Some LF sources have major biases (which are a priori unknown to the analyst) and must be excluded from the search process from the very beginning. Including such LF sources increases the overall sampling costs and may also result in convergence to an incorrect solution.
    \item If highly cheap LF sources are available, a naively designed AF chooses to sample from them very frequently since the information value of a candidate point is inversely scaled by the cost of its source. This heavily biased sampling can force BO to converge to the optima of those sources rather than the HF source.
\end{itemize}

As reviewed in Section \ref{sec: background-bo}, existing multi-fidelity BO technologies partially address these challenges by ad hoc tuning and making simplifying assumptions. These assumptions often include presuming a direct relationship between fidelity level and sampling cost, assuming LF sources are always useful, or manually adjusting the sampling costs (e.g., converting the $1000/50/1$ cost ratio between three sources to $1/0.5/0.1$). These manual changes are quite laborious and result in either convergence to an incorrect solution or higher overall costs compared to single-fidelity (SF) BO which solely leverages the HF source.

In this paper, we provide new perspectives for learning from multi-fidelity sources in the context of BO. In particular, we argue that $(a)$ the emulator must fuse the multi-source data in a nonlinearly learnt manifold to maximally leverage cross-source correlations and also indicate trustworthy LF sources that do not deteriorate BO's performance, and $(b)$ the AF should use the available information on the LF sources (HF source) solely for exploration (exploitation). As demonstrated in Section \ref{sec: results}, these contributions endow our framework with significant performance improvements over existing technologies. 

The rest of the paper is organized as follows. We review the relevant technical background in Section \ref{sec: background} and introduce our approach in Section \ref{sec: approach}. We test the performance of our MFCA BO framework against the state-of-the-art on a set of analytic and two engineering problems in Section \ref{sec: results} and finally conclude the paper in section \ref{sec: conclusion}. We also provide a nomenclature and list of symbols in Appendix \ref{sec: nomenclature}. The \href{https://gitlab.com/S3anaz/multi-fidelity-cost-aware-bayesian-optimization/-/tree/main/}{Gitlab repository} associated with this project hosts supplementary materials.  


    \section{Technical Background} \label{sec: background}
In this Section, we first provide some background on latent map Gaussian process (LMGP) which is one of the key components of our MFCA BO framework. Then, we elaborate on the two main ingredients of BO in Section \ref{sec: background-bo} where we also review some of the existing methods for handling MF data.

\subsection{Latent Map Gaussian Processes (LMGPs)} \label{sec: background-LMGP}
For metamodeling via GPs, one assumes that the training data comes from a multivariate normal distribution with parametric mean and covariance functions and then uses closed-form conditional distribution formulas for prediction. Below, we first detail the process for estimating these parameters when the input space contains categorical and/or numerical variables. Then, we provide the prediction formulas.

Assume the training data is a realization from a GP and that the following relation holds:

\begin{equation} 
    \begin{split}
        y(\boldsymbol{x})=\boldsymbol{m}(\boldsymbol{x}) \boldsymbol{\beta}+\xi(\boldsymbol{x})
    \end{split}
    \label{eq: GP-prior}
\end{equation}

\noindent where $\boldsymbol{x}=[x_1, x_2, \ldots, x_{dx}]^T$ is the input vector, $y(\boldsymbol{x})$ is the output/response, $\boldsymbol{m}(\boldsymbol{x})=\left[m_1(\boldsymbol{x}), \ldots, m_{d\beta}(\boldsymbol{x})\right]$ are a set of parametric basis functions, $\boldsymbol{\beta} = [\beta_1, \beta_2,..., \beta_{d\beta}]^T$ are unknown coefficients, and $\xi(\boldsymbol{x})$ is a zero-mean GP whose covariance function or kernel is:

\begin{equation} 
    \begin{split}
        \operatorname{cov}\left(\xi(\boldsymbol{x}), \xi\left(\boldsymbol{x}^{\prime}\right)\right)=c\left(\boldsymbol{x}, \boldsymbol{x}^{\prime}\right)=\sigma^2 r\left(\boldsymbol{x}, \boldsymbol{x}^{\prime}\right)
    \end{split}
    \label{eq: GP-Cov1}
\end{equation}

\noindent where $\sigma^{2}$ is the variance of the process and $r(.,.)$ is a parametric correlation function. A common choice for $r(.,.)$ is: 

\begin{equation} 
    \begin{split}
        r(\boldsymbol{x}, \boldsymbol{x}^{\prime})=\exp \{-\sum_{i=1}^{dx} 10^{\omega_{i}}(x_{i}-x_{i}^{\prime})^{2}\}=\exp \{-(\boldsymbol{x}-\boldsymbol{x}^{\prime})^{T} 10^{\boldsymbol{\Omega}}(\boldsymbol{x}-\boldsymbol{x}^{\prime})\}
    \end{split}
    \label{eq: GP-Corelation1}
\end{equation}

\noindent where $\boldsymbol{\omega}=[\omega_{1}, \ldots, \omega_{dx}]^{T},-\infty<\omega_{i}<\infty$ are the scale parameters and $\boldsymbol{\Omega}=\operatorname{diag}(\boldsymbol{\omega})$.
GP modeling largely depends on the choice of the correlation function which measures the weighted distance between any two inputs, see Eq. \ref{eq: GP-Corelation1}. As recently motivated in \cite{oune2021latent}, to directly use GPs for MF modeling we must extend them such that they can handle categorical inputs. This extension primarily relies on reformulating $r(.,.)$ and can be accomplished in different ways. In this paper, we employ LMGPs \cite{oune2021latent} since $(1)$ they have been shown to outperform other approaches, and $(2)$ they provide a visualizable and interpretable manifold which can be used to detect discrepancies among data sources (this manifold helps us to exclude highly biased sources from BO). 

Let us denote categorical inputs via $ \boldsymbol{t}=[t_1, \ldots, t_{dt}]^T$ where variable $ t_i $ has $ l_i $ distinct levels. For example, $ t_1=\{Male, Female\} $ and $ t_2=\{Persian, American, Spanish\} $ are two categorical inputs where $l_1=2$ and $l_2=3$. To handle mixed inputs, $\boldsymbol{u}=[x_1, ..., x_{dx}, t_1, ..., t_{dt}]^T$, LMGP learns a unified parametric function that maps each combination of categorical variables to a point in a manifold or latent space\footnote{Multiple mapping functions can also be used and we leverage this in Section \ref{sec: BO-LMGP} where two mapping functions are learned to build two manifolds.}. This mapping function can be incorporated into any standard correlation function such as the Gaussian which is reformulated as follows for mixed inputs:
\begin{equation} 
    \begin{split}
        r(\boldsymbol{u}, \boldsymbol{u}^{\prime})=         
        \exp \{-(\boldsymbol{x}-\boldsymbol{x}^{\prime})^T \boldsymbol{\Omega}(\boldsymbol{x}-\boldsymbol{x}^{\prime}) - \|\boldsymbol{z}(\boldsymbol{t})-\boldsymbol{z}(\boldsymbol{t}^{\prime})\|_2^2\}
    \end{split}
    \label{eq: LMGP-Corelation}
\end{equation}
or equivalently,
\begin{equation} 
    \begin{split}
        r(\boldsymbol{u}, \boldsymbol{u}^{\prime}) = 
        \exp \{-\sum_{i=1}^{dx} 10^{\omega_i}(x_i-x_i^{\prime})^2\} \times 
        \exp \{-\sum_{i=1}^{dz}(z_i(\boldsymbol{t})-z_i(\boldsymbol{t}^{\prime}))^2\} 
    \end{split}
    \label{eq: LMGP-Corelation-extended}
\end{equation}

\noindent where $\|\cdot\|_2$ denotes the Euclidean 2-norm and $\boldsymbol{z}(\boldsymbol{t})=\left[z_1(\boldsymbol{t}), \ldots, z_{dz}(\boldsymbol{t})\right]_{1 \times dz}$ is the location in the learned manifold corresponding to the specific combination of the categorical variables denoted by $\boldsymbol{t}$. To find these latent points, LMGP first assigns a unique prior representation (a unique vector) to each combination of categorical variables. Then, it learns a linear transformation\footnote{More complex transformations based on, e.g., deep neural networks can also be used.} that maps these unique vectors into a compact manifold (aka latent space) with dimensionality $dz$:
\begin{equation} 
    \begin{split}
        \boldsymbol{z(t)=\zeta(t) A}
    \end{split}
    \label{eq: LMGP-zeta}
\end{equation}
\noindent where $\boldsymbol{t}$ denotes a specific combination of the categorical variables, $\boldsymbol{z(t)}$ is the $1 \times dz$ posterior latent representation of $\boldsymbol{t}$, $\boldsymbol{\zeta}(\boldsymbol{t})$ is the unique prior vector representation of $\boldsymbol{t}$, and $\boldsymbol{A}$ is a rectangular matrix that maps $\boldsymbol{\zeta(t)}$ to $\boldsymbol{z(t)}$.
There are various methods for constructing the prior vectors $\boldsymbol{\zeta} $ and we refer the reader to \cite{oune2021latent} for more details. In this paper, we use grouped one-hot encoding which makes $\boldsymbol{\zeta}(\boldsymbol{t})$ and $\boldsymbol{A}$ to be of sizes $1 \times \sum_{i=1}^{dt} l_{i}$ and $\sum_{i=1}^{dt} l_{i} \times dz$, respectively. 
For instance, in the above example the grouped one-hot encoded version of the combination $\boldsymbol{t}=[Female, American]^T$ is $\boldsymbol{\zeta(\boldsymbol{t})}=[0,1,0,1,0]$ where the first two numbers encode the levels of $t_1$ while the last three numbers indicate the levels of $t_2$.


To emulate via LMGP, the hyperparameters ($\boldsymbol{\beta,A, \omega} \text {, and } \sigma^{2}$) must be estimated via the training data. To find these estimates, we utilize maximum likelihood estimation (MLE) due to its computational efficiency over fully Bayesian techniques. 
The MLE estimates the hyperparameters such that they maximize the likelihood of the $n$ training data being generated by $y(\boldsymbol{x})$, that is:

\begin{equation} 
    \begin{split}
        [\widehat{\boldsymbol{\beta}}, \hat{\sigma}, \widehat{\boldsymbol{\omega}}, \widehat{\boldsymbol{A}}] = 
        \underset{\boldsymbol{\beta}, \sigma^2, \omega, \boldsymbol{A}}{\operatorname{argmax}}\left|2 \pi \sigma^2 \boldsymbol{R}\right|^{-\frac{1}{2}} \times \exp \left\{\frac{-1}{2}(\boldsymbol{y}-\boldsymbol{M} \boldsymbol{\beta})^T\left(\sigma^2 \boldsymbol{R}\right)^{-1}(\boldsymbol{y}-\boldsymbol{M} \boldsymbol{\beta})\right\}
    \end{split}
    \label{eq: LMGP-parameter1}
\end{equation}

\noindent or equivalently:
\begin{equation} 
    \begin{split}
        [\widehat{\boldsymbol{\beta}}, \hat{\sigma}, \widehat{\boldsymbol{\omega}}, \widehat{\boldsymbol{A}}]=\underset{\boldsymbol{\beta}, \sigma^2, \omega, \boldsymbol{A}}{\operatorname{argmin}} \frac{n}{2} \log \left(\sigma^2\right)+\frac{1}{2} \log (|\boldsymbol{R}|)+\frac{1}{2 \sigma^2}(\boldsymbol{y}-\boldsymbol{M} \boldsymbol{\beta})^T \boldsymbol{R}^{-1}(\boldsymbol{y}-\boldsymbol{M} \boldsymbol{\beta}),
    \end{split}
    \label{eq: LMGP-parameter2}
\end{equation}

\noindent where $\log (\cdot)$ is the natural logarithm, $|\cdot|$ denotes the determinant operator, $\boldsymbol{y}=[y^1, \ldots, y^n]^T$ is the $n \times 1$ vector of outputs in the training data, $\boldsymbol{R}$ is the $n \times n$  correlation matrix with the $(i, j)^{t h} \text { element } R_{i j}=r(\boldsymbol{x}^i, \boldsymbol{x}^j) \text { for } i, j=1, \ldots, n$ , and $\boldsymbol{M}$ is the $n \times d\beta$ matrix with the $(i, j)^{th}$ element $M_{ij}=m_j(\boldsymbol{x}^i) \text { for } i=1, \ldots, n \text { and } j=1, \ldots, d\beta$.

Using the method of profiling \cite{bostanabad2018leveraging}, Eq. \ref{eq: LMGP-parameter2} can be simplified to:
\begin{equation} 
    \begin{split}
        [\widehat{\boldsymbol{\omega}}, \widehat{\boldsymbol{A}}]=\underset{\boldsymbol{\omega}, \boldsymbol{A}}{\operatorname{argmin}} \hspace{3mm}n \log \left(\hat{\sigma}^2\right)+\log (|\boldsymbol{R}|)=\underset{\boldsymbol{\omega}, \boldsymbol{A}}{\operatorname{argmin}} \hspace{3mm}L,
    \end{split}
    \label{eq: LMGP-omega-A}
\end{equation}
\noindent where $\hat{\sigma}^2 = \frac{1}{n}(\boldsymbol{y}-\boldsymbol{M} \hat{\boldsymbol{\beta}})^T \boldsymbol{R}^{-1}(\boldsymbol{y}-\boldsymbol{M} \hat{\boldsymbol{\beta}})$ and $\hat{\boldsymbol{\beta}}=[\boldsymbol{M}^T \boldsymbol{R}^{-1} \boldsymbol{M}]^{-1}[\boldsymbol{M}^T \boldsymbol{R}^{-1} \boldsymbol{y}]$. Eq. \ref{eq: LMGP-omega-A} can be efficiently solved via a gradient-based optimization technique \cite{RN783, RN649}. 

Once the hyperparameters are estimated, the conditional distribution formulas are used to predict the response distribution at the arbitrary point $\boldsymbol{x}^*$. The mean and variance of this normal distribution are:
\begin{equation} 
    \begin{split}        
    \mathbb{E}[y(\boldsymbol{x}^*)]=\mu(\boldsymbol{x}^*)=
    \boldsymbol{m}(\boldsymbol{x}^*) \widehat{\boldsymbol{\beta}}+\boldsymbol{r}^T(\boldsymbol{x}^*) \boldsymbol{R}^{-1}(\boldsymbol{y}-\boldsymbol{M} \widehat{\boldsymbol{\beta}})
    \end{split}
    \label{eq: LMGP-predict}
\end{equation}

\begin{equation} 
    \begin{split}        
    \operatorname{cov}(y(\boldsymbol{x}^*), y(\boldsymbol{x}^*))=\sigma^2(\boldsymbol{x}^*)=
    \hat{\sigma}^2(1 -
    \boldsymbol{r}^T(\boldsymbol{x}^*) \boldsymbol{R}^{-1} \boldsymbol{r}(\boldsymbol{x}^*) +
    \boldsymbol{g}(\boldsymbol{x}^*)(\boldsymbol{M}^T \boldsymbol{R}^{-1} \boldsymbol{M})^{-1} \boldsymbol{g}(\boldsymbol{x}^*)^T
    )
    \end{split}
    \label{eq: LMGP-posterio-cov}
\end{equation}
\noindent where $\mathbb{E}$ denotes expectation, $m(\boldsymbol{x}^*)=[m_1(\boldsymbol{x}^*), \ldots, m_{d\beta}(\boldsymbol{x}^*)]$, $\boldsymbol{r}(\boldsymbol{x}^*)$ is an $n \times 1$ vector with the $i^{t h}$ element $r(\boldsymbol{x}^i, \boldsymbol{x}^*)$,  and $\boldsymbol{g}(\boldsymbol{x}^*)=\boldsymbol{m}(\boldsymbol{x}^*)-\boldsymbol{M}^T \boldsymbol{R}^{-1} \boldsymbol{r}(\boldsymbol{x}^*)$.


\subsection{Bayesian Optimization (BO)} \label{sec: background-bo}
BO is increasingly used to optimize expensive-to-evaluate (and typically black-box) functions. As opposed to gradient-based or heuristic optimization techniques that only rely on function evaluations (or predictions of a surrogate), BO leverages the probabilistic predictions of a sequentially updated emulator. In this paper, we focus on single-response functions over unconstrained and bounded domains but note that BO can also handle multi-response (or multi-task) \cite{wang2016new}, composite \cite{astudillo2019bayesian}, or constrained problems \cite{gelbart2014bayesian}. 

As summarized in Algorithm \ref{alg: BO-algorithm}, BO has an iterative nature where an emulator is first fitted to some initial training data. This emulator is then queried via the AF which calculates the expected utility of any point in the input space, i.e., $\mathbb{E}[I(\boldsymbol{x})]$ where $I(\boldsymbol{x})$ is the user-defined utility function. These queries are used in the auxiliary optimization problem\footnote{Gradient-based optimization techniques are almost always used at this stage.} which aims to find the candidate point with the maximum expected utility in the input space. Once this point is found, the expensive function is queried and the resulting (input, output) pair is used to update the training data. Given the augmented dataset, the above process is repeated until a convergence metric is met.

Except for some special cases, solving the auxiliary optimization problem in Algorithm \ref{alg: BO-algorithm} is highly costly since each evaluation of its objective function (i.e., $\mathbb{E}[I(\boldsymbol{x})]$) amounts to integration which cannot be calculated analytically. Fortunately, the special cases perform quite well in most practical applications and hence are used frequently (we also employ them in our framework). 

Algorithm \ref{alg: BO-algorithm} is strictly sequential in that the dataset is augmented with a single sample at each iteration. To leverage parallel computing, one can augment the dataset with a pool of samples which jointly maximize the expected utility \cite{wang2020parallel}. 
Additionally, Algorithm \ref{alg: BO-algorithm} is myopic in that the AF does not consider the effect of the to-be-evaluated sample on the emulator in the future iterations. This myopic nature is addressed in look-ahead AFs such as KG (detailed below) which typically provide improved performance at the expense of significantly increasing the cost of solving the auxiliary optimization problem.

\begin{algorithm} 
    \SetAlgoLined
    \DontPrintSemicolon
    \textbf{Given:} Initial data $\mathcal{D}^{k}=\{(\boldsymbol{x}^i, y^i)\}_{i=1}^k$, expensive black-box function $f(\boldsymbol{x})$\\  
    \textbf{Define:} Utility function $I(\boldsymbol{x})$, stop conditions \\
    \While{stop conditions not met}{
    \begin{enumerate}
        \item Train the GP emulator using $\mathcal{D}^k$
        \item Define the acquisition function $\alpha(\boldsymbol{x})=\mathbb{E}[I(\boldsymbol{x}) \mid \mathcal{D}^k]$
        \item Solve the auxiliary optimization problem: $\boldsymbol{x}^{k+1} = \underset{{\boldsymbol{x} \in \mathbb{X}}}{\argmax} \hspace{2 mm} \alpha(\boldsymbol{x})$
        \item Query $f(\boldsymbol{x})$ at $\boldsymbol{x}^{k+1}$ to obtain $y^{k+1}$
        \item Update data: $\mathcal{D}^{k+1} \leftarrow \mathcal{D}^{k} \cup (\boldsymbol{x}^{k+1}, y^{k+1})$
        \item Update counter: $k \leftarrow k+1$
    \end{enumerate}       
    }
    \textbf{Output:} Updated data $\mathcal{D}^{k}=\left\{\left(\boldsymbol{x}^{i}, y^{i}\right)\right\}_{i=1}^k$, GP emulator 
 
    \caption{Strictly Sequential and Myopic Bayesian Optimization for Maximization}
    \label{alg: BO-algorithm}
\end{algorithm}

\subsubsection{Single-Fidelity Acquisition Functions} \label{sec: background-AFs}
Many different AFs have been proposed for diverse applications and in this Section we review three of the most widely used ones: EI, PI, and KG. While the first two AFs are myopic\footnote{EI does have a look-ahead version \cite{zhang2021two}.}, KG is look-ahead. Any AF calculates the expected value of a user-defined utility function conditioned on the available data $D$, that is:
\begin{equation} 
    \begin{split}
        \alpha(\boldsymbol{x})=\mathbb{E}[I(\boldsymbol{x}) \mid \mathcal{D} ]
    \end{split}
    \label{eq: general-AF}
\end{equation}

Our proposed AF for MFCA BO (see Section \ref{sec: proposed-AF}) leverages EI as well as PI and hence is also myopic. In Section \ref{sec: results} we compare our AF to EI, PI, and KG and indicate that it consistently outperforms them. 

PI is an AF that favors exploitation \cite{wang2017new}, i.e., it rewards samples that improve \fbest ~which is the best function value seen so far. For instance, when maximizing the expensive black-box function \fobj, this AF uses the following utility function:
\begin{equation} 
    \begin{split}
        I_{PI}(\boldsymbol{x})= \begin{cases}1 & y(\boldsymbol{x})>y^* \\ 0 & y(\boldsymbol{x}) \leq y^*\end{cases} 
    \end{split}
    \label{eq: PI-improvement}
\end{equation}
\noindent where \yobj ~is the emulator-based prediction at $\boldsymbol{x}$. Based on Eq. \ref{eq: PI-improvement}, if \yobj ~ is less than \fbest, the point $\boldsymbol{x}$ has zero utility. Assuming a GP is used for emulation, \yobj ~follows a normal distribution whose mean and variance are given in Eq. \ref{eq: LMGP-predict} and Eq. \ref{eq: LMGP-posterio-cov}, respectively. Using the reparameterization trick (see Section \ref{sec: appendix_PI}) we can show that the resulting AF based on $I_{PI}(\boldsymbol{x})$ is \cite{jones2001taxonomy,gutmann2001radial}:
\begin{equation} 
    \begin{split}
        \alpha_{PI}(\boldsymbol{x})=\Phi\left(\frac{\mu(\boldsymbol{x})-y^*}{\sigma(\boldsymbol{x})}\right)
    \end{split}
    \label{eq: PI-AF}
\end{equation}
\noindent where $\mu(\boldsymbol{x})$ and $\sigma(\boldsymbol{x})$ are defined in Eq. \ref{eq: LMGP-predict} and Eq. \ref{eq: LMGP-posterio-cov}, and $\Phi(z)$ is the cumulative density function (CDF) of the standard normal random variable $z$. Eq. \ref{eq: PI-AF} clearly indicates that $\alpha_{PI}(\boldsymbol{x})$ favors exploitation because $\Phi(z)$ is maximized at locations where the predictions are close to $y^*$ and have small uncertainty. 

In contrast to PI which discards the magnitude of improvement (regardless of the magnitude of $y(\boldsymbol{x})$, $I_{PI}(\boldsymbol{x})$ is either $0$ or $1$), EI rewards large improvements over \fbest ~by adopting the following utility function: 
\begin{equation} 
    \begin{split}
        I_{EI}(\boldsymbol{x})=\max (y(\boldsymbol{x})-y^*, 0)
    \end{split}
    \label{eq: EI-utility}
\end{equation}
The corresponding AF can now be obtained by substituting Eq. \ref{eq: EI-utility} in Eq. \ref{eq: general-AF} and using the reparameterization trick (see Section \ref{sec: appendix_PI} for the details):
\begin{equation} 
    \begin{split}
        \alpha_{EI}(\boldsymbol{x})=(\mu(\boldsymbol{x})-y^*)\Phi(\frac{\mu(\boldsymbol{x})-y^*}{\sigma(\boldsymbol{x})})+\sigma(\boldsymbol{x}) \phi(\frac{\mu(\boldsymbol{x})-y^*}{\sigma(\boldsymbol{x})})
    \end{split}
    \label{eq: EI-AF-compl}
\end{equation}

\noindent where $\mu(\boldsymbol{x})$ and $\sigma(\boldsymbol{x})$ are given in Eq. \ref{eq: LMGP-predict} and Eq. \ref{eq: LMGP-posterio-cov}, respectively, and $\phi(z)$ is the probability density function (PDF) of $z$. Eq. \ref{eq: EI-AF-compl} clearly demonstrates that $\alpha_{EI}(\boldsymbol{x})$ strikes a balance between exploration and exploitation when it is used as the objective function of the auxiliary optimization problem in Algorithm \ref{alg: BO-algorithm}: while the second term on the right-hand side directly deals with uncertainty and hence encourages exploration, the first term favors exploitation \cite{schonlau1998global, lyu2017efficient}.  

Another widely used AF is KG which, unlike PI and EI, is look-ahead because it chooses $\boldsymbol{x}^{k+1}$ (see Algorithm \ref{alg: BO-algorithm}) based on the effect of the yet-to-be-seen observation (i.e., $y^{k+1}$ which follows a normal distribution) on the optimum value predicted by the emulator. Following the terminology and setup of Algorithm \ref{alg: BO-algorithm}, this AF quantifies the expected utility of $\boldsymbol{x}$ at iteration $k+1$ as:
\begin{equation} 
    \begin{split}
    \alpha_{KG}(\boldsymbol{x}) = 
    \mathbb{E}_{p(y \mid \boldsymbol{x}, \mathcal{D}^k)}[\max{\mu^{k+1}}(\boldsymbol{x})] -
    \max{\mu^k}(\boldsymbol{x})
    \label{eq: KG-AF-untract}
    \end{split}
\end{equation}
\noindent where $\max{\mu^k}(\boldsymbol{x})$ denotes the maximum mean prediction of the GP trained on $\mathcal{D}^k$. The expectation operation in Eq. \ref{eq: KG-AF-untract} appears due to the fact that $y^{k+1}$ is not observed yet and $\alpha_{KG}(\boldsymbol{x})$ is relying on the predictive distribution provided by the GP that is trained on $\mathcal{D}^k$. This expectation cannot be calculated analytically and hence a Monte Carlo estimate is used in practice:
\begin{equation} 
    \begin{split}
        \alpha_{KG}(\boldsymbol{x}) \approx 
        \frac{1}{M} \sum_{m=1}^M \max{{\mu^{k+1}}^m(\boldsymbol{x})} -
        \max{\mu^k}(\boldsymbol{x})
        \end{split}
    \label{eq: KG-AF}
\end{equation}
\noindent where $\max{{\mu^{k+1}}^m(\boldsymbol{x})}$ is calculated by first drawing a sample at $\boldsymbol{x}$ from the GP that is trained on $\mathcal{D}^k$ and then retraining the GP on $\mathcal{D}^{k} \cup (\boldsymbol{x}, y^m)$ where $y^m$ is response of the drawn sample. In practice, a small value must be chosen for $M$ since maximizing $\alpha_{KG}(\boldsymbol{x})$ over the input space at each iteration of BO is very expensive. We refer the readers to \cite{frazier2008knowledge, balandat2020botorch} for more information on KG and its implementation.

\subsubsection{Existing Multi-fidelity BO Techniques} \label{sec: background-MFBO}
As stated in Section \ref{sec: intro}, the overall computational efficiency of BO can be increased by leveraging inexpensive LF datasets. MF BO has been successfully used in many applications such as hyperparameter tuning \cite{zhu2020accelerating, lindauer2019boah, cho2020basic, wu2020practical}, finding Pareto fronts in multi-objective optimizations \cite{phiboon2021experiment, sun2022correlated, belakaria2020multi}, and solving non-linear state-space models \cite{dahlin2014approximate, imani2019mfbo}. For MFBO, both the emulator and the AF must accommodate the multi-source and unbalanced\footnote{LF sources contribute more samples to the training data since they are typically much cheaper to query from.} nature of the data. 

Co-Kriging which is an extension of Kriging (or GP) is a popular emulator that handles MF data by reformulating the covariance function in Eq. \ref{eq: GP-Cov1} as follows (assuming there are three data sources denoted by $A$, $B$, and $C$):
\begin{equation} 
    \begin{split}
        \operatorname{cov}([y_A(\boldsymbol{x}), y_B(\boldsymbol{x}), y_C(\boldsymbol{x})]^T, [y_A(\boldsymbol{x}^{\prime}), y_B(\boldsymbol{x}^{\prime}), y_C(\boldsymbol{x}^{\prime})]^T) =
        \Sigma \otimes r\left(\boldsymbol{x}, \boldsymbol{x}^{\prime}\right)
    \end{split}
    \label{eq: Co-krig-cov}
\end{equation}
where $\otimes$ denotes the Kronecker product and $\Sigma$ is a symmetric positive-definite matrix of size $3 \times 3$. This reformulation assumes that all the responses (regardless of the source) follow a multi-variate normal distribution and that the matrix $\Sigma_{ij}$ captures the overall correlation between sources $i$ and $j$\footnote{GPs can handle multi-response datasets in a similar manner, see \cite{bonilla2007multi}.}. While this method can fuse any number of data sources, it fails to accurately capture cross-source correlations since the matrix $\Sigma$ has insufficient learning capacity. 

Another well-known MF emulation method is that of Kennedy and O'Hagan who fuse bi-fidelity datasets by learning a discrepancy function that aims to explain the differences between HF and LF sources. While this bi-fidelity emulator has proved useful in a wide range of applications \cite{stainforth2005uncertainty, zhang2019numerical}, it has some major drawbacks such as inability to jointly learn from more than two sources, numerical issues, and assuming a priori additive relation between the discrepancy function and the two data sources. 

The bi-fidelity approach of Kennedy and O'Hagan can be viewed as a special case of hierarchical MF modeling where it is assumed that the relative accuracy between all the data sources is known. Space mapping techniques belong to this category, but they are rarely used for sequential sampling, BO, or MF modeling (see \cite{RN1134} for a bi-fidelity example). These techniques are typically employed in solving partial differential equations, particularly to accelerate the convergence of an HF simulation (e.g., based on fine discretization) by initializing it via the results of an LF simulation. 

Upon reformulating the covariance function in Eq. \ref{eq: GP-Cov1}, GPs can also be used for hierarchical MF modeling. For instance, the single-task MF GP of the popular BoTorch package adopts an additive covariance function that relies on introducing two user-defined quantitative features \cite{RN1392, RN1270}. The first feature, denoted by $x_a$, is restricted to the $[0, 1]$ range and assigns a fidelity value to a source based on the user's belief (larger values correspond to higher fidelities). The second feature, denoted by $x_b$, is the iteration fidelity parameter and benefits MF BO specifically in the context of hyperparameter tuning of large machine learning models.
The covariance function directly uses these two additional features as follows:
\begin{equation} 
    \begin{split}             
        \operatorname{cov}(\boldsymbol{x}, \boldsymbol{x}^\prime) =
        c_0(\boldsymbol{x}, \boldsymbol{x}^\prime) +
        e_1(x_{a}, x^\prime_{a}) c_1(\boldsymbol{x}, \boldsymbol{x}^\prime) +
        e_2([x_a,x_b]^T, [ x^\prime_{a}, x^\prime_{b}]^T) c_2(\boldsymbol{x}, \boldsymbol{x}^\prime) +
        e_3(x_{b}, x^\prime_{b}) c_3(\boldsymbol{x},\boldsymbol{x}^\prime)
    \label{eq: Singletaskgp-Cov}
    \end{split}
\end{equation}
\noindent where $c_i(\boldsymbol{x}, \boldsymbol{x}^\prime)$ are Matern kernels\footnote{The parameters of these kernels are endowed with Gamma priors in BoTorch.} that characterize the spatial correlations across the numerical inputs and $e_i(\cdot)$ are user-defined functions that model the cross-source correlations. 
$e_1(x_{a}, x^\prime_{a})$ and $e_3(x_{b}, x^\prime_{b})$ are bias kernels that aim to take the discrepancies among the sources into account while $e_2([x_a,x_b]^T, [ x^\prime_{a}, x^\prime_{b}]^T)$ models the fidelities' interaction kernel (see Appendix \ref{sec: single-GP} for more details).

Despite being useful \cite{RN1790}, Eq. \ref{eq: Singletaskgp-Cov} has some limitations. For instance, it requires a priori knowledge about the exact hierarchy of fidelities and how they should be encoded as a numerical feature (i.e., $x_a$). Additionally, the manually-defined functions $e_i(\cdot)$ are insufficiently flexible to learn complex cross-source relations and they also do not provide any learned metric that quantifies which LF sources are useful for MF BO.

Compared to the few emulators described above, the diversity of the AFs in MF BO is more since they are often tailored to the application, see \cite{shahriari2015taking, RN1392, RN1393, RN1789}. Many of these developments leverage existing AFs that are used in SF BO such as EI, PI, upper confidence bound \cite{srinivas2012information}, Thompson sampling \cite{russo2014learning}, or GP-predictive entropy search \cite{hernandez2014predictive}.
One specific example is most likely EI (MLEI) \cite{pautrat2018bayesian} which is tailored to direct policy search problems where the EI in Eq. \ref{eq: EI-AF-compl} is first scaled by multiple context-specific priors and then the resulting AFs are optimized to determine the next candidate point. 
As another example, Wu et al \cite{wu2020practical} develop trace-aware KG to accelerate the hyperparameter tuning process of machine learning models whose training relies on minimizing the loss function (defined as the expected prediction error on the validation data). MF BO is useful in this process since the evaluation accuracy of the loss function can be controlled by parameters such as the number of iterations and training/validation data points. Correspondingly, trace-aware KG adjusts these parameters to use LF but inexpensive evaluations of the loss function (and it trace) during training. We highlight that EI is also widely used in hyperparameter tuning problems \cite{ponweiser2008multiobjective, zhang2020bayesian, wu2019hyperparameter}. 



    \section{Proposed Approach} \label{sec: approach}
As was previously stated, the emulator and AF are the two fundamental components of any BO framework. In this Section, we first discuss the rationale for using LMGP as the emulator of our MF BO framework in Section \ref{sec: BO-LMGP} and then introduce our novel cost-aware AF in Section \ref{sec: proposed-AF}. We elaborate on the convergence conditions and provide an algorithmic summary of our framework in Section \ref{sec: termination}.

\subsection{Multi-fidelity Emulation via LMGP} \label{sec: BO-LMGP}
\begin{figure*}[!h] 
    \centering
    \includegraphics[page=1, width = 0.45\textwidth]{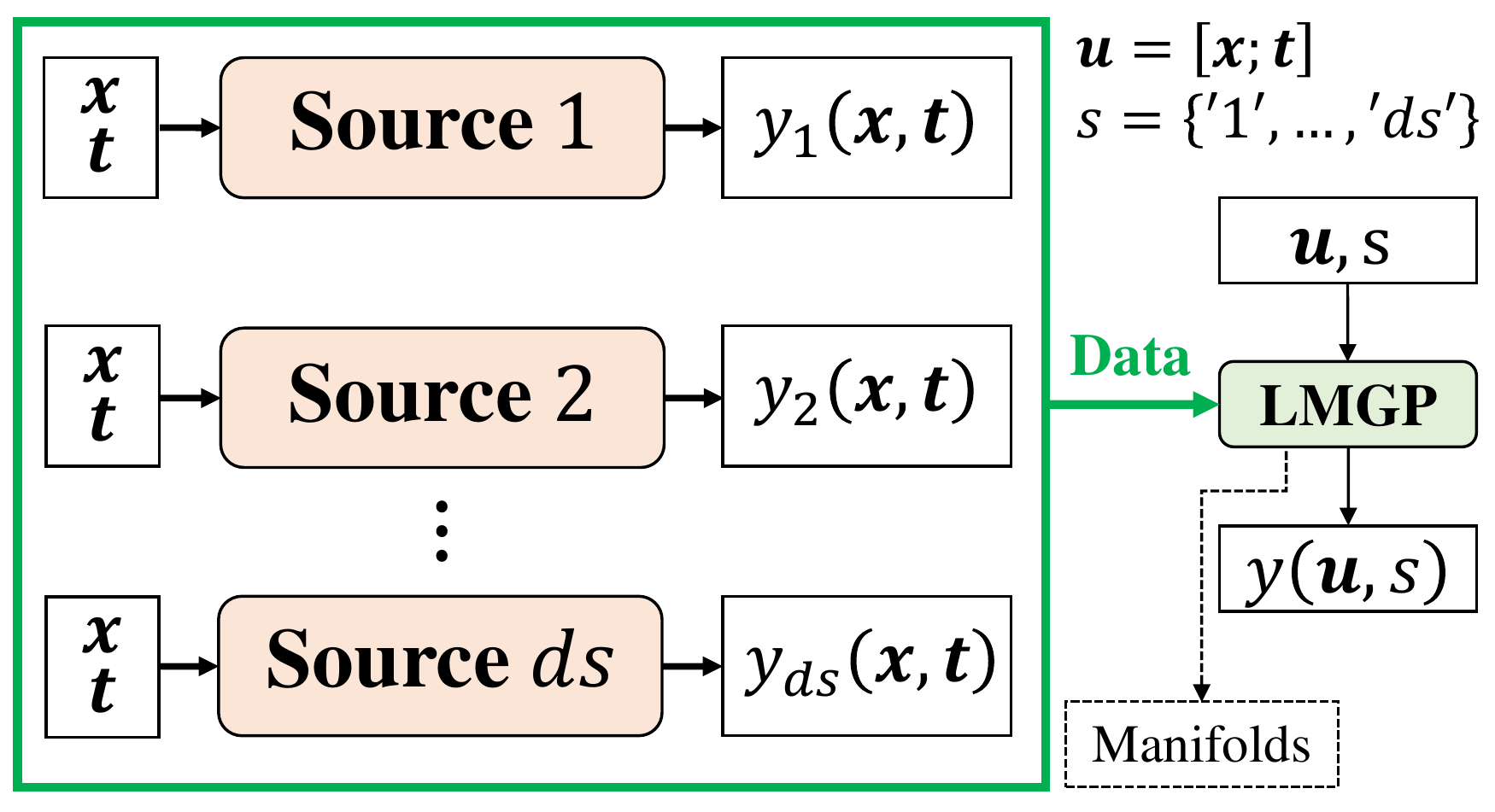}
    \vspace{-0.3cm}
    \caption{\textbf{Multi-fidelity emulation with LMGP:} The training data is built by first augmenting the inputs with the \textit{categorical} feature $s$ that denotes the data source of a sample and then concatenating all the inputs and outputs, see Eq. \ref{fig: LMGP-fidelity-append}. LMGP can use one or more manifolds to encode the categorical variables into a quantitative space. For MF emulation, we recommend using two manifolds to simplify the visualization of cross-source relations: one manifold for $s$ and the other for the rest of the categorical variables, i.e., $\boldsymbol{t}$.}
    \label{fig: LMGP-flowchart}
\end{figure*}

As schematically illustrated in Figure \ref{fig: LMGP-flowchart}, MF emulation via LMGPs is quite straightforward: Assuming there are $ds$ data sources, we augment the inputs with the additional \textit{categorical} variable $s=\{'1', \cdots, 'ds'\}$ whose $j^{th}$ element corresponds to data source $j$ for $j = 1, \cdots, ds$. After this augmentation, the inputs and outputs of all the datasets are concatenated as (following the notation of Figure \ref{fig: LMGP-flowchart}):
\begin{equation} 
    \begin{split}
        \boldsymbol{U}=\left[\begin{array}{cc}
        \boldsymbol{U}_1 & '\mathbf{1}'_{n_{1} \times 1} \\
        \boldsymbol{U}_2 & '\mathbf{2}'_{n_{2} \times 1} \\
        \vdots & \vdots \\
        \boldsymbol{U}_{ds} & '\mathbf{ds}'_{n_{ds} \times 1}
        \end{array}\right] 
        \text { and }
        \boldsymbol{y}=\left[\begin{array}{c}
        \boldsymbol{y}_1 \\
        \boldsymbol{y}_2 \\
        \vdots \\
        \boldsymbol{y}_{ds}
        \end{array}\right]          
    \end{split}
    \label{fig: LMGP-fidelity-append}
\end{equation}
\noindent where the subscripts $1, 2, ..., ds$ correspond to the data sources, $n_j$ is the number of samples obtained from $s(j)$ (i.e., source $j$), $\boldsymbol{U}_j$ and $\boldsymbol{y}_j$ are, respectively, the $n_j \times (dx + dt)$ feature matrix and the ${n_j \times 1}$ vector of responses obtained from $s(j)$, and $'\boldsymbol{j}'$ is a categorical vector of size ${n_j \times 1}$ whose elements are all set to $'j'$. Once the $\{\boldsymbol{U}, \boldsymbol{y}\}$ dataset is built, it is directly fed into LMGP to build an MF emulator. 

\begin{figure*}[!t] 
    \centering
    \includegraphics[page=1, width = 1\textwidth]{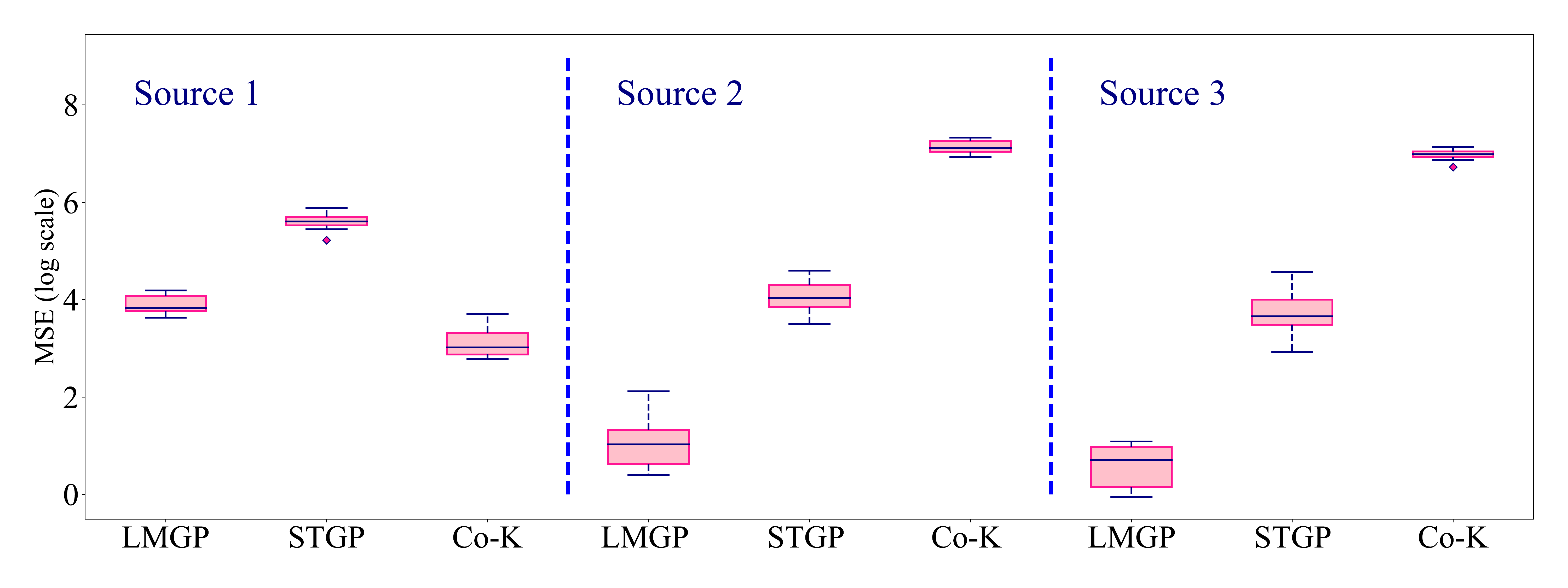}
    \vspace{-0.3cm}
    \caption{\textbf{Emulator comparison:} We compare the prediction accuracy of LMGP against single-task multi-fidelity GP (STGP) and Co-Kriging (Co-K) on an analytic problem with three sources (see Borehole in Table \ref{table: analytic-formulation} where HF, LF3, and LF4 are used). While LMGP and STGP use $(5,5,50)$ initial data for (HF, LF3, LF4) sources, Co-K uses $(50,50,50)$ to achieve comparable performance. 
    Each emulator is trained $10$ times by randomizing the initial data. It is evident that LMGP consistently outperform other methods in emulating all the sources. Prediction accuracy is measured by calculating the mean squared error (MSE, on $log$ scale) on unseen test data.}
    \label{fig: emulator-comparison}
\end{figure*}

We argue that, compared to the existing techniques (see Section \ref{sec: background-MFBO}), LMGPs provide a more flexible and accurate mechanism to build MF emulators, see Figure \ref{fig: emulator-comparison} for a comparison study on an analytic example. This superiority is because LMGP learns the relations between the sources (which are hidden in the combined datasets) in a manifold. This manifold is learned nonlinearly and hence has a much higher representation power than methods that rely on linear operations, e.g., the matrix $\Sigma$ in co-Kriging that linearly scales the correlations, see Eq. \ref{eq: Co-krig-cov}. 
Similarly, LMGP has major advantages over single-task GPs (STGPs) reviewed in Section \ref{sec: background-MFBO} because $(1)$ it does not assume any hierarchy across the data sources, $(2)$ the cross-source relations are encoded via learned latent variables which have significantly higher representation power than a single user-defined scalar variable (see Eq. \ref{eq: Singletaskgp-Cov}) that directly affects the covariance function of the underlying GP and requires knowledge of the relative source fidelities. 

As shown in Figure \ref{fig: emulator-comparison}, LMGPs build MF emulators that more accurately learn all the sources (rather than just the HF source). 
While we can alter the formulations in Section \ref{sec: background-LMGP} such that LMGPs prioritize learning the HF source, we do not believe this is a good general decision in the context of MF BO. The reasoning behind our belief is that the quality of LF predictions (obtained via the emulator) greatly affects the exploration nature of BO and, hence, its convergence behavior. 
The empirical results in Section \ref{sec: results} strongly support this reasoning. 

Another major advantage of LMGPs is that their learned manifold provides an intuitive and visualizable global metric for comparing the relative discrepancies/similarities among the data sources, see Figure \ref{fig: LMGP_ALL4_LS} for four examples with different number of data sources. This manifold is particularly useful in detecting anomalous sources whose data adversely affects MF BO. We demonstrate this in Section \ref{sec: results} with some examples where we use the initial MF data (i.e., before starting the BO iterations) to correctly predict whether SF BO outperforms MF BO. 

Given the importance of identifying relative discrepancies among data sources, we slightly modify the correlation function of LMGP to learn two manifolds where the first one encodes the original categorical variables (collectively denoted by $\boldsymbol{t}$ in Section \ref{sec: background-LMGP}) while the second one encodes the data source identifier (denoted by $s$ in Figure \ref{fig: LMGP-flowchart}). The new correlation function is (compare to Eq. \ref{eq: LMGP-Corelation-extended}):
\begin{align}
    r(\begin{bmatrix}
        \boldsymbol{x}\\
        \boldsymbol{t}\\
        s
    \end{bmatrix} , 
    \begin{bmatrix}
        \boldsymbol{x^\prime}\\
        \boldsymbol{t^\prime}\\
        s^{\prime}
    \end{bmatrix}) = 
    \exp \{-\sum_{i=1}^{dx} 10^{\omega_i}(x_i-x_i^{\prime})^2\ -
    \sum_{i=1}^{dz}(z_i(\boldsymbol{t})-z_i(\boldsymbol{t}^{\prime}))^2 -
    \sum_{i=1}^{dh}(h_i(s)-h_i(s^\prime))^2\}
    \label{eq: LMGP-Corelation-extended2}
\end{align}
\noindent where $\boldsymbol{h}(s) = [h_1, \cdots, h_{dh}]^T$ is the latent representation of data source $s$. Similar to Eq. \ref{eq: LMGP-zeta}, this latent representation is obtained by post-multiplying a prior vector by the parametric matrix $\boldsymbol{A}_h$:
\begin{equation} 
    \begin{split}
        \boldsymbol{h}(s)=\boldsymbol{\zeta}(s) \boldsymbol{A}_h
    \end{split}
    \label{eq: LMGP-zeta2}
\end{equation}
In our studies, we always design the prior by one-hot-encoding the categorical variable $s=\{'1', \cdots, 'ds'\}$ that identifies the data source and estimate all of LMGP's hyperparameters via MLE:
\begin{equation} 
    \begin{split}
        [\widehat{\boldsymbol{\omega}}, \widehat{\boldsymbol{A}}, \widehat{\boldsymbol{A}}_h] = 
        \underset{\boldsymbol{\omega}, \boldsymbol{A}, \boldsymbol{A}_h}{\operatorname{argmin}}
        \hspace{3mm} n \log \left(\hat{\sigma}^2\right)+\log (|\boldsymbol{R}|)
    \end{split}
    \label{eq: LMGP-omega-A2}
\end{equation}
\noindent where the elements of $\boldsymbol{R}$ are now built using Eq. \ref{eq: LMGP-Corelation-extended2}.

We now use Eq. \ref{eq: LMGP-Corelation-extended2} to explain the relation between the latent fidelity representations, i.e., $\boldsymbol{h}(s)$, and the relative fidelity of the data sources. At the same inputs, the correlation between the estimated outputs of sources $s$ and $s^\prime$ is:
\begin{align}
    0 \leq
    corr(y_s(\boldsymbol{x}, \boldsymbol{t}), y_{s^\prime}(\boldsymbol{x}, \boldsymbol{t})) = 
    r(\begin{bmatrix}
        \boldsymbol{x}\\
        \boldsymbol{t}\\
        s
    \end{bmatrix} , \begin{bmatrix}
        \boldsymbol{x}\\
        \boldsymbol{t}\\
        s^\prime
    \end{bmatrix}) = 
    \exp \{0 -
    0 -
    \sum_{i=1}^{d_h}(h_i(s)-h_i(s^\prime))^2\}
    \leq 1
\end{align}
\label{eq: LMGP-Corelation-extended3}
So sources with similar fidelities which provide highly correlated responses, must have similar latent representations, i.e., $\boldsymbol{h}_i(s) \simeq \boldsymbol{h}_i(s^\prime)$. This relation is illustrated in Figure \ref{fig: LMGP_ALL4_LS} where sources with similar fidelities are encoded by close-by points in the manifold. 
 
\subsection{Multi-source Cost-aware Acquisition Function} \label{sec: proposed-AF}
\begin{figure}[!h]
    \centering
    \begin{subfigure}{.5\textwidth}
        \centering
        \includegraphics[width=1\linewidth]{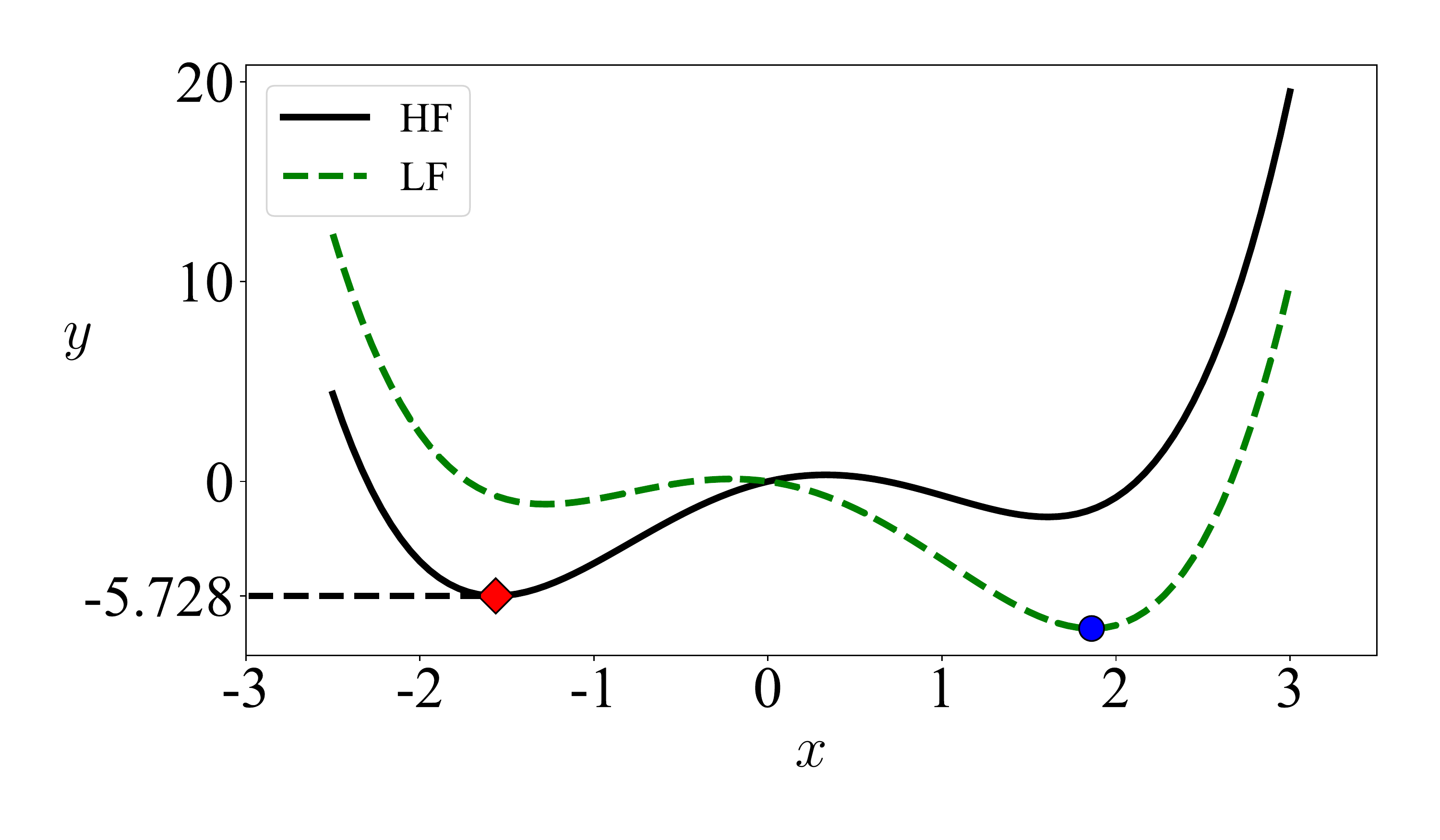}
        \caption{\textbf{Biased LF source}}
        \label{fig: 1d-curvature}
    \end{subfigure}%
    \begin{subfigure}{.5\textwidth}
        \centering
        \includegraphics[width=1\linewidth]{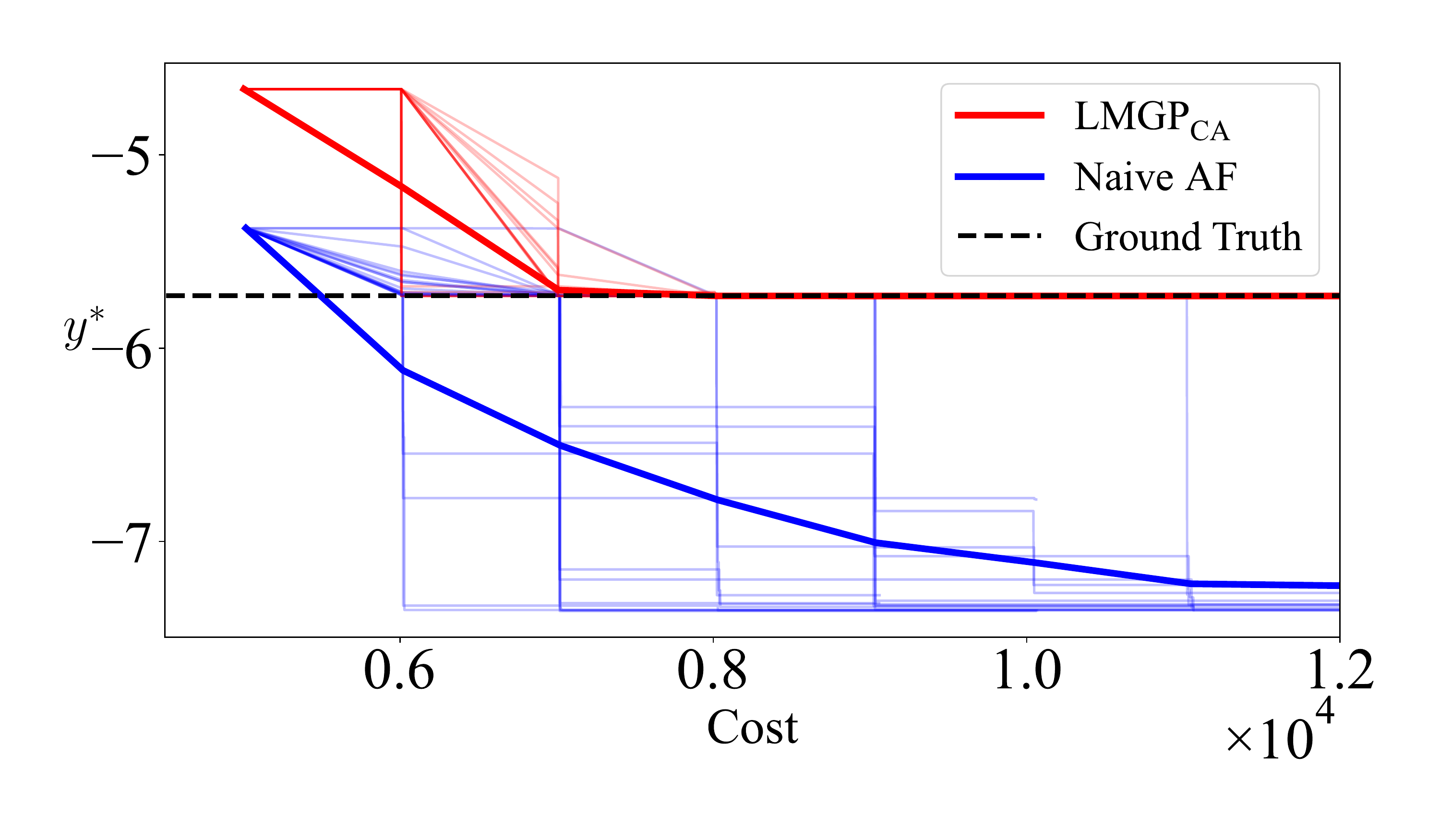}
        \caption{\textbf{Effect of AF on convergence}}
        \label{fig: 1d-convergence}
    \end{subfigure}
    \caption{\textbf{Double-well potential:} The LF source has a systematic bias because its optima do not match with those of the HF source. However, since the two curves have similar trends, an effective AF can leverage this correlation to reduce the overall data acquisition costs. The LF and HF sources have a cost-per-sample of, respectively, $1$ and $1000$ in this example.}
    \label{fig: 1D-approach}
\end{figure}

The choice of AF affects the performance of BO quite significantly. This choice is especially important in MF BO because, in addition to balancing exploration and exploitation, the AF has to consider the biases of LF data and source-dependent sampling costs. 
To demonstrate these challenges, consider the analytic example in Figure \ref{fig: 1d-curvature} where, while the two functions are correlated, the LF source's global optimum (the location and the corresponding $y$ value) is quite different than that of the HF source. Since LF sources are typically much cheaper than the HF source, a naive AF (that merely scales the expected utility based on the cost) forces MF BO to converge to the global optimum of the LF source, see Figure \ref{fig: 1d-convergence} for a one-dimensional example.

Contrary to existing approaches, we argue that the key to addressing the above challenges is to quantify the information value of LF and HF data based on different metrics which are then compared against each other to determine the candidate input and the corresponding source. 
In particular, we propose to use the LF sources exclusively for exploration to leverage their correlations with the HF source while preventing them from dominating the convergence behavior of MF BO. 
Additionally, we propose to exclusively employ the HF source for exploitation to maximally use its trustworthy samples\footnote{These samples may be corrupted via $\epsilon \sim \mathcal{N}(0,\sigma^2) $ where $\sigma^2$ is the (unknown) noise variance.} during optimization.

To develop the AF for the $j^{th}$ LF source with $j = [1, \cdots, ds]$ and $j \neq l$ where $l$ denotes the HF source, we follow Section \ref{sec: background-AFs} and define the improvement (for a maximization problem) as $y_j(\boldsymbol{x})–y_j^*$ where $y_j(\boldsymbol{x})$ denotes the LMGP-based prediction at $\boldsymbol{x}$ for source $j$ and $y_j^*$ is the best function value in the obtained dataset from source $j$. We use the reparameterization trick to rewrite this improvement as:
\begin{equation} 
    \begin{split}
        \frac{y_j(\boldsymbol{x}) - \mu_j(\boldsymbol{x})}{\sigma_j(\boldsymbol{x})} = z \sim \mathcal{N}(0,1) 
        \Rightarrow
        y_j(\boldsymbol{x}) - y_j^*= (\mu_j(\boldsymbol{x})- y_j^*) + \sigma_j(\boldsymbol{x})z
    \end{split}
    \label{eq: LF-rewrite-u}
\end{equation}
\noindent where $\mu_j(\boldsymbol{x})$ and $\sigma_j(\boldsymbol{x})$ are defined in Eq. \ref{eq: LMGP-predict} and Eq. \ref{eq: LMGP-posterio-cov}, respectively. In Eq. \ref{eq: LF-rewrite-u} the ($\mu_j(\boldsymbol{x})- y_j^*$) and $\sigma_j(\boldsymbol{x})z$ terms control the exploitation and exploration aspects of the improvement, respectively. We now define our utility function that focuses on exploration by dropping the first term on the far right-hand-side of Eq. \ref{eq: LF-rewrite-u}:
\begin{equation} 
    \begin{split}
        I_{LF}(\boldsymbol{x}; j)= 
        \begin{cases}
            \sigma_j(\boldsymbol{x})z& y_j(\boldsymbol{x})>y_j^* \\ 0 & y_j(\boldsymbol{x}) \leq y_j^*
        \end{cases}
    \end{split}
    \label{eq: LF-utility}
\end{equation}
which is used for the $j^{th}$ LF source in our framework. 
We obtain the corresponding AF by substituting $I_{LF}(\boldsymbol{x}; j)$ in Eq. \ref{eq: general-AF}:
\begin{equation} 
    \begin{split}
        \alpha_{LF}(\boldsymbol{x}; j)=\int_{-\infty}^{\infty} I_{LF}(\boldsymbol{x}; j) \phi(z) \mathrm{d} z=\int_{-\infty}^{\infty} \sigma_j(\boldsymbol{x})z \phi(z) \mathrm{d} z
    \end{split}
    \label{eq: LF-CA-AF}
\end{equation}
\noindent The integral is zero for $y_j(\boldsymbol{x}) < y_j^*$ so we find the corresponding switch point in terms of $z$: 
\begin{equation} 
    \begin{split}
        y_j(\boldsymbol{x})=y_j^* \Rightarrow \mu_j(\boldsymbol{x})+\sigma_j(\boldsymbol{x}) z=y_j^* \Rightarrow z_{0}=\frac{y_j^*-\mu_j(\boldsymbol{x})}{\sigma_j(\boldsymbol{x})}
    \end{split}
    \label{eq: LF_AF_switch}
\end{equation}
\noindent Inserting Eq. \ref{eq: LF_AF_switch} in Eq. \ref{eq: LF-CA-AF} yields:
\begin{equation} 
    \begin{split}
        \begin{aligned}
            &\alpha_{LF}(\boldsymbol{x}; j)=\int_{z_0}^{\infty} \sigma_j(\boldsymbol{x}) z \phi(z) \mathrm{d} z
            =\int_{z_0}^{\infty} \frac{\sigma_j(\boldsymbol{x}) z}{\sqrt{2 \pi}} e^{-\frac{z^2}{2}} \mathrm{d} z\\
            &=\frac{\sigma_j(\boldsymbol{x})}{\sqrt{2 \pi}} \int_{z_0}^{\infty} z e^{-\frac{z^2}{2}} \mathrm{d} z=
            \frac{\sigma_j(\boldsymbol{x})}{\sqrt{2 \pi}} \int_{z_0}^{\infty}(e^{-\frac{z^2}{2}})^{\prime} \mathrm{d} z\\
            &=-\frac{\sigma_j(\boldsymbol{x})}{\sqrt{2 \pi}}[e^{-\frac{z^2}{2}}]_{z_0}^{\infty}
            =\sigma_j(\boldsymbol{x}) \phi(z_0)
            = \sigma_j(\boldsymbol{x}) \phi(\frac{y_j^* - \mu_j(\boldsymbol{x})}{\sigma_j(\boldsymbol{x})})
        \end{aligned}
    \end{split}
    \label{eq: LF_AF}
\end{equation}
\noindent Comparison between this AF and Eq. \ref{eq: EI-AF-compl} illustrates that our proposed AF for LF sources is the same as the exploration part of EI.

For the HF source, we propose to use PI as the AF because it focuses on exploitation and is computationally efficient (the efficiency is due to the analytic form of PI which dispenses with expensive numerical integration). Assuming source $l$ provides the HF data, this AF is given by:
\begin{equation} 
    \begin{split}
        \alpha_{HF}(\boldsymbol{x}; l)=\Phi(\frac{\mu_l(\boldsymbol{x})-y_l^*}{\sigma_l(\boldsymbol{x})})
    \end{split}
    \label{eq: HF-AF}
\end{equation}

We use $\alpha_{HF}(\boldsymbol{x}; l)$ and $\alpha_{LF}(\boldsymbol{x}; j)$ as defined above in each iteration of BO to solve $ds$ auxiliary optimization problems (assuming there are $ds$ data sources) to find the candidate location with the highest expected utility from each source. We then find the final candidate point and the corresponding source by comparing the cost-normalized version of these $ds$ values. Hence, assuming source $l$ provides HF data among the $ds$ sources, our AF that considers source-dependent costs and fidelities is:
\begin{equation}
    \alpha_{MFCA}(\boldsymbol{x}; j) = 
    \begin{cases}
        \sfrac{\alpha_{LF}(\boldsymbol{x}; j)}{O(j)} & j = [1, \cdots, ds] \And j \neq l \\
        \sfrac{\alpha_{HF}(\boldsymbol{x}; l)}{O(l)} & j = l
    \end{cases}
    \label{eq: MFCA-AF}
\end{equation}
\noindent and
\begin{equation}
    [\boldsymbol{x}^{k+1}, j^{k+1}] = 
    \underset{\boldsymbol{x}, j}{\argmax} \hspace{2 mm} \alpha_{MFCA}(\boldsymbol{x}; j)
    \label{eq: aux-opt MFCA}
\end{equation}
\noindent where $O(j)$ is the cost associated with taking a single sample from source $j$ and $\boldsymbol{x}^{k+1}$ is the point that source $j^{k+1} \in \{1, \cdots, ds \}$ must be evaluated in the current BO iteration. We now point out several important aspects of Eq. \ref{eq: MFCA-AF}.

The AFs in Eq. \ref{eq: LF_AF} and Eq. \ref{eq: HF-AF} quantify the value of a sample by comparing it to the best available sample for the corresponding source (and not the best sample across all the sources). 
The advantage of this source-wise comparison in each of the AFs is that it encourages sampling from sources that provide larger values (which is desirable for a maximization problem). 
However, this formulation enables LF sources whose optima are larger than the HF source to dominate the optimization process where not only more samples are taken from these LF sources (which may also cause numerical issues), but also the converged solution does not belong to the profile of the HF source. This issue is exacerbated once the AFs are divided by the data collection costs (see Eq. \ref{eq: MFCA-AF}) since LF samples are (typically) much cheaper than HF data. 

The abovementioned issues are addressed with three mechanisms in our MFCA approach. Firstly, we always report the points sampled from the HF source as the final optimization history (this choice ensures that the final solution indeed belongs to the profile of the HF source but it does not guarantee global optimality\footnote{BO does not guarantee global optimality anyways.}). 
Secondly, we always use the fidelity manifold of the LMGP that is trained on the \textit{initial} data to detect the LF sources that should not be used in BO due to their severe discrepancy (we demonstrate the benefit of this exclusion with examples in Section \ref{sec: results}). 
Thirdly, we have designed the core of our AFs for the HF and LF sources based on, respectively, the CDF and the scaled PDF of the standard normal variable $z$. As detailed below and empirically shown in Section \ref{sec: results}, the intricate relation between these two functions during the optimization reduces the effect of LF sources' biases on the convergence. 

\begin{figure*}[!t] 
	\centering
	\includegraphics[width = 0.5\textwidth]{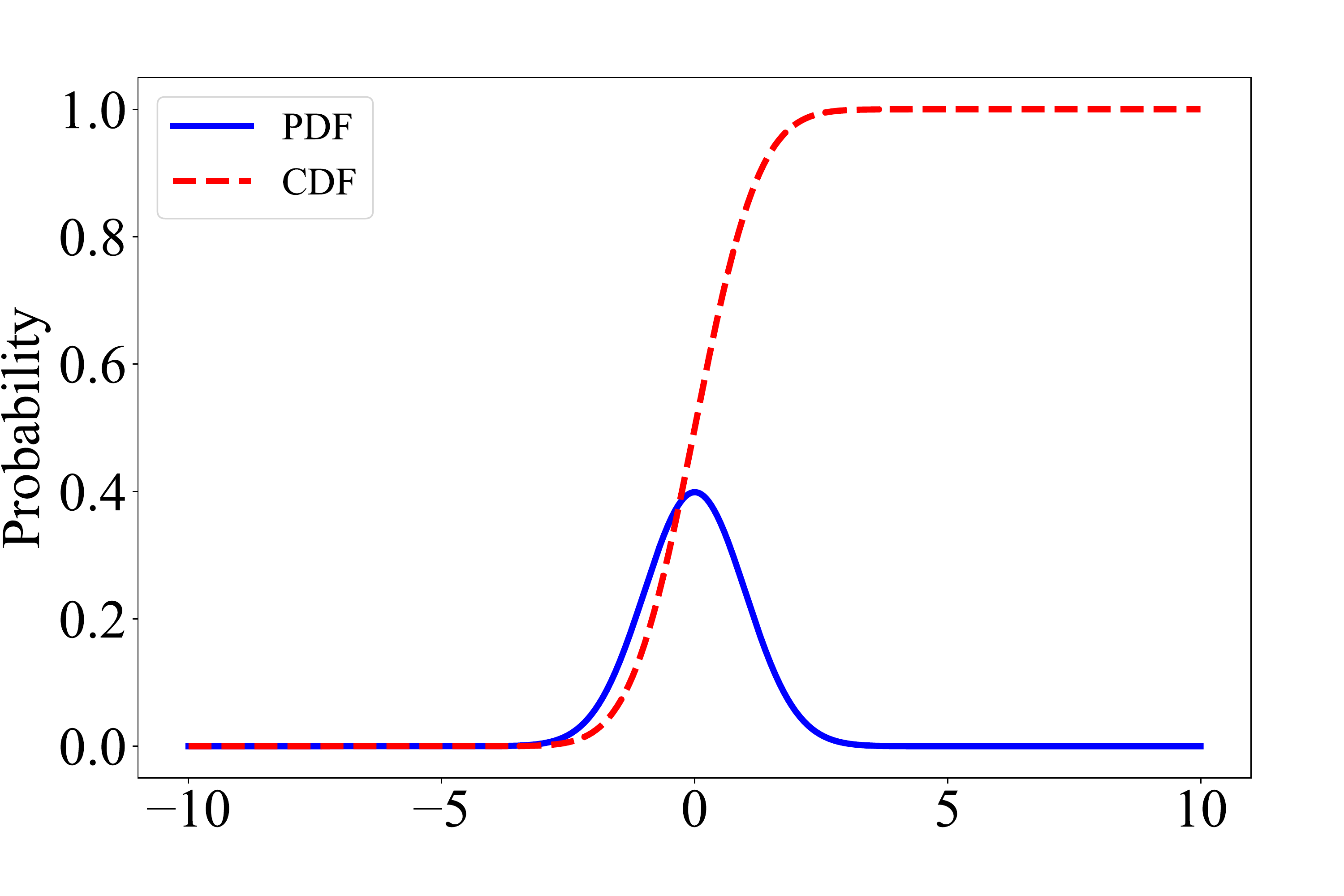}
	\caption{\textbf{Standard normal variable:} The PDF and CDF of $z \sim \mathcal{N}(0, 1)$ have comparable magnitudes up to the mean of $z$ (i.e., $0$) but increasingly differ after $0$.}
    \label{fig: PDF_vs_CDF}
\end{figure*}

As illustrated in Figure \ref{fig: PDF_vs_CDF}, $\phi(z)$ and $\Phi(z)$ have comparable values up to $\mathbb{E}[z]=0$ but the ratio $\Phi(z)/\phi(z)$ increasingly grows as $z$ realizes larger values. This trend indicates that if an HF candidate point sufficiently improves $y_l^*$, then Eq. \ref{eq: aux-opt MFCA} queries the HF source at that point to obtain a new sample for the next BO iteration. The frequency of this query during the optimization process is controlled by the data collection cost and $\sigma_j(\boldsymbol{x})$, see Eq. \ref{eq: LF_AF} and Eq. \ref{eq: HF-AF}, respectively. 
If HF samples are highly costly, the auxiliary optimization in Eq. \ref{eq: aux-opt MFCA} reduces the sampling frequency. However, unlike existing approaches such as BoTorch, this reduction does not translate into "never sampling from the HF source" (even if the cost ratios are as large as $1000$, see Section \ref{sec: results}) because $\Phi(z)/\phi(z)$ can be quite large. 
Regarding the $\sigma_j(\boldsymbol{x})$ term in Eq. \ref{eq: LF_AF}, we note that it encourages exploring the regions where LMGP provides highly uncertain predictions for an LF source. This scenario happens when an LF source is rarely sampled and there are insufficient correlations between that source and other sources. 

In summary, our proposed AF in Eq. \ref{eq: MFCA-AF}, while involving intricate interactions between the fidelities and costs, has a simple form which is analytic (and hence computationally efficient) and interpretable. As we illustrate in Section \ref{sec: results} this AF, combined with LMGP, dramatically improves the performance of our MFCA BO framework.  

\subsection{Convergence Metric} \label{sec: termination}
Similar to AFs, convergence criteria of MF BO techniques are traditionally tailored to the application since many factors (e.g., budget constraints, numerical issues, or convergence history) affect the results. We believe the emulator and AF of our framework alleviate many of the convergence issues associated with MF BO and hence use two simple convergence metrics: overall costs and maximum number of iterations without improvement. The former is a rather generic metric but it can result in considerably high number of iterations if one of the LF sources is extremely inexpensive to query. For this reason, we recommend using additional metrics (such as the second one above) that track convergence.

Algorithm \ref{alg: MFBO-algorithm} summarizes our framework for MFCA BO. Compared to Algorithm \ref{alg: BO-algorithm}, the major differences are in the choice of the emulator and AF which now can handle multi-source data that have different levels of fidelity and cost. In addition, a pre-processing step is added which leverages the fidelity manifold of LMGP to detect the LF sources that must be excluded from the BO process due to their large discrepancies with respect to the HF source. 

\begin{algorithm}[!t]
    \SetAlgoLined
    \DontPrintSemicolon
    \textbf{Given:} Initial multi-fidelity data $\mathcal{D}^{k}=\{(\boldsymbol{x}^i, y^i\}_{i=1}^k$, black-box functions $f(\boldsymbol{x}; j)$ and their corresponding sampling costs $O(j)$ where $j = [1,...,ds]$\\
    \textbf{Goal:} Optimizing high-fidelity function (source $l \in [1,...,ds]$)\\ 
    \textbf{Define:} Utility functions (see Eq. \ref{eq: PI-improvement} and Eq. \ref{eq: LF-utility}) and stop conditions \\
    \textbf{Step $0$:} Train an \texttt{LMGP} and exclude highly biased low-fidelity sources based on its fidelity manifold\\
    \While{stop conditions not met}{
        \begin{enumerate}
            \item Train an \texttt{LMGP} using $\mathcal{D}^k$
            \item Define the multi-fidelity cost-aware acquisition function (see Equations \ref{eq: LF_AF}, \ref{eq: HF-AF}, and \ref{eq: MFCA-AF}):\\
            $\alpha_{M F C A}(\boldsymbol{x} ; j)= \begin{cases}\alpha_{L F}(\boldsymbol{x} ; j) / O(j) & j \in\{1, \cdots, d s\} \quad \& \quad j \neq l \\ \alpha_{H F}(\boldsymbol{x} ; l) / O(l) & j=l\end{cases}$
            \item Solve the auxiliary optimization problem:\\$[\boldsymbol{x}^{k+1},j^{k+1}]=\underset{\boldsymbol{x}, j}{\operatorname{argmax}} \hspace{2 mm} \alpha_{M F C A} (\boldsymbol{x} ; j)$
            \item Query $f(\boldsymbol{x};j)$ at point $\boldsymbol{x}^{k+1}$ from source $j^{k+1}$ to obtain $y^{k+1}$
            \item Update data: $\mathcal{D}^{k+1} \leftarrow \mathcal{D}^{k} \cup (\boldsymbol{x}^{k+1}, y^{k+1})$
            \item Update counter: $k \leftarrow k+1$
        \end{enumerate}       
    }
    \textbf{Output:} Updated data $\mathcal{D}^{k}=\{(\boldsymbol{x}^{i}, y^{i})\}_{i=1}^k$, \texttt{LMGP}
    
    \caption{Multi-fidelity Cost-aware Bayesian Optimization for Maximization}
    \label{alg: MFBO-algorithm}
\end{algorithm}

\cmt{
\noindent based on this algorithm, the first step is to fit a LMGP model on the initial data. Then the LF and HF data will be fed into their corresponding AFs and the next sample and its corresponding fidelity source will be found through the auxiliary optimization problem. Having the new sample, we query its function value from its corresponding fidelity source to augment the initial data. This process will be repeated by the time it reaches the termination condition. 

Defining termination metric is quite challenging in MFBO problems, since they are affected by many factors. In general, budget constraint has always been the main obstacle in optimization problems. Therefore, we define budget limitation to be the first stop condition of our proposed MFCA BO model. However, MF methods mostly sample from inexpensive LF data, so they never reach the stop condition or they need many excessive iterations to reach it. Therefore, a second stop condition is also defined in this paper. 
As mentioned in Sec.\ref{sec: proposed-AF}, \fbest ~is an approximation of the ground truth in our MFCA BO proposed approach; thus, the permanency of \fbest ~for large enough iterations can be a sign of convergence. So we defined the fixity of \fbest for 50\footnote{It is only a large enough number gained through trial and error.} iterations as our second stop condition.
}
    \section{Results and Discussions} \label{sec: results}
We compare the performance of the following four methods on four analytic and two real-world examples (Sections \ref{sec: analytical-examples} and \ref{sec: real-world example}, respectively):
\begin{itemize}
    \item \LMGP: Our proposed MFCA BO approach.
    \item \MEI: Single-fidelity BO whose emulator and AF are LMGP and EI, respectively. 
    \item \MPI: Single-fidelity BO whose emulator and AF are LMGP and PI, respectively. 
    \item \BOT: Multi-fidelity BO with BoTorch where STGP and KG are used as emulator and AF.
\end{itemize}

The first three methods are myopic and hence computationally much faster than \BOT ~which is lookahead (see Section \ref{sec: background-MFBO} for details on \BOT). Assuming the computational costs of BO (which are mainly associated with emulation and solving the auxiliary optimization problem) are negligible compared to querying any of the data sources, we compare the above methods in terms of cost and accuracy which are defined as, respectively, the overall data collection costs and the ability to find the global optimum of the HF source. 

Our rationales for comparing \LMGP ~against the above three methods are to demonstrate: 
$(1)$ the advantages of employing LF sources in BO, 
$(2)$ that our designed myopic AF (see Eq. \ref{eq: MFCA-AF}) improves both the sampling cost and accuracy compared to even lookahead methods such as \BOT, and
$(3)$ the importance of excluding highly biased LF sources from BO. We also note that the emulators in \BOT ~cannot handle categorical inputs and hence we compare \LMGP ~to \MEI ~and \MPI ~in Section \ref{sec: real-world example} where the two examples have categorical variables.  

For all the methods, we terminate the optimization process if either of the following conditions are met: $(1)$ the overall sampling cost reaches a pre-determined maximum budget, or $(2)$ the best HF sample (i.e., $y_l^*$ in Eq. \ref{eq: HF-AF}) does not improve over $50$ consecutive iterations.
These conditions are quite simple and straightforward; allowing us to focus on the effects of AF and the emulator on the performance. 

In Section \ref{sec: analytical-examples}, we set the maximum budget to $40000$ units for \LMGP, \MEI, and \MPI ~while for \BOT ~we choose $50000$ since it is, as demonstrated below, highly inefficient and requires more samples to provide reasonable accuracy. Additionally, in all of our examples an HF sample costs $1000$ so \MEI ~and \MPI ~are terminated based on the maximum budget condition. In Section \ref{sec: analytical-examples}, we set the maximum budget to $1000$ and $1800$ for the two examples since their data collection cost are much lower than the examples in Section \ref{sec: analytical-examples}.

\subsection{Analytic Examples} \label{sec: analytical-examples}
As detailed in Table \ref{table: analytic-formulation} in Appendix \ref{sec: appendix-table}, we consider four analytic examples (\potential, \rosen, \borehole, and \wing) whose input dimensionality ranges from $1$ to $10$. All examples are single-response and the number of data sources varies between $2$ and $5$. 
The source-dependent sampling costs and number of initial data points are also detailed in Table \ref{table: analytic-formulation}. To compare the robustness and effectiveness of the four BO methods described above, we use relatively small initial datasets (especially from the HF source) and consider various cost ratios (the maximum cost ratio between two sources equals $1000$). For each example, we quantify the effect of random initial data by repeating the optimization process $20$ times for each method (all initial data are generated via Sobol sequence). 

\begin{figure}[!b]
    \centering
    \begin{subfigure}{.5\textwidth}
        \centering
        \includegraphics[width=1\linewidth]{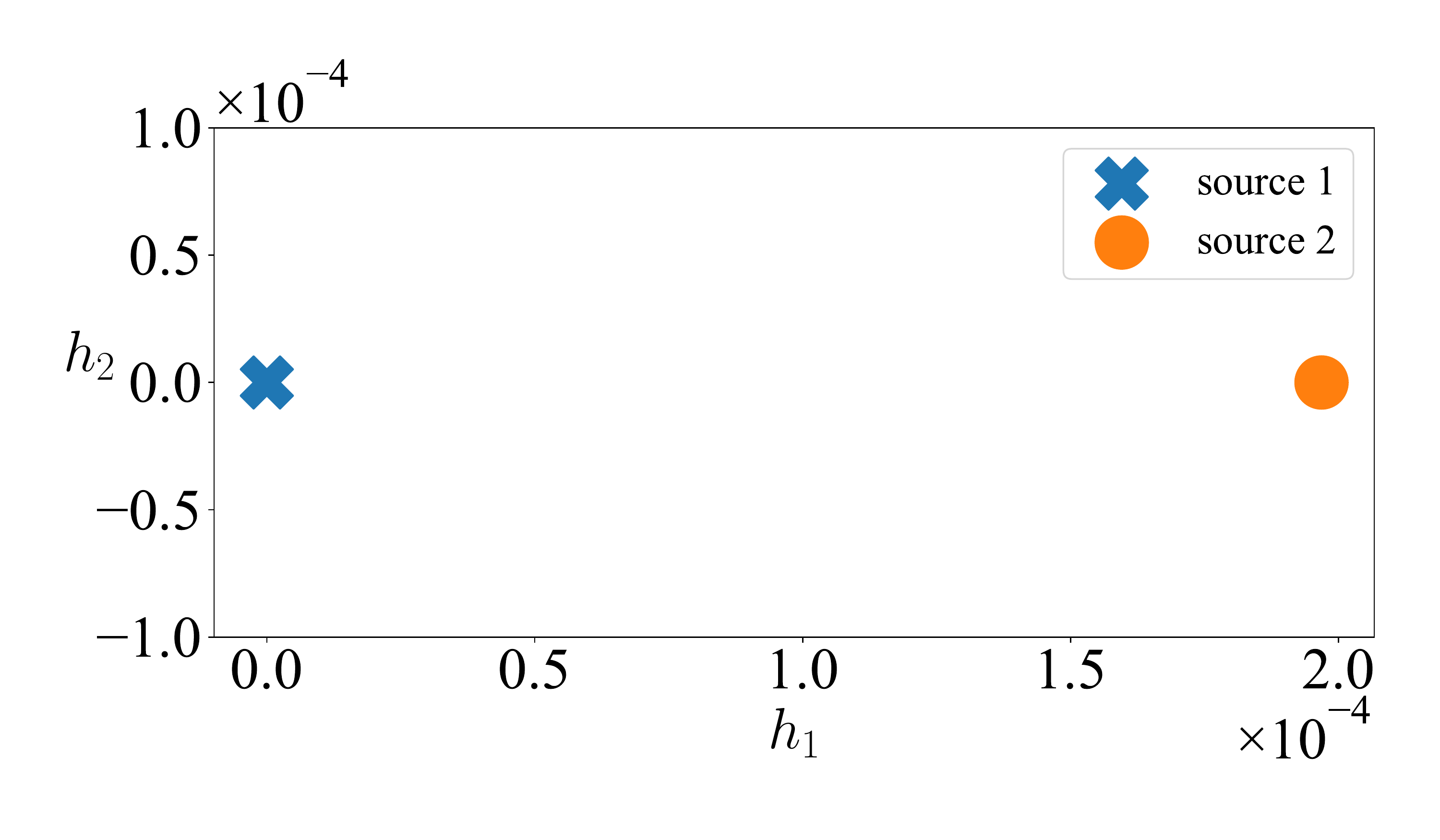}
        \caption{\potential}
    \end{subfigure}%
    \begin{subfigure}{.5\textwidth}
        \centering
        \includegraphics[width=1\linewidth]{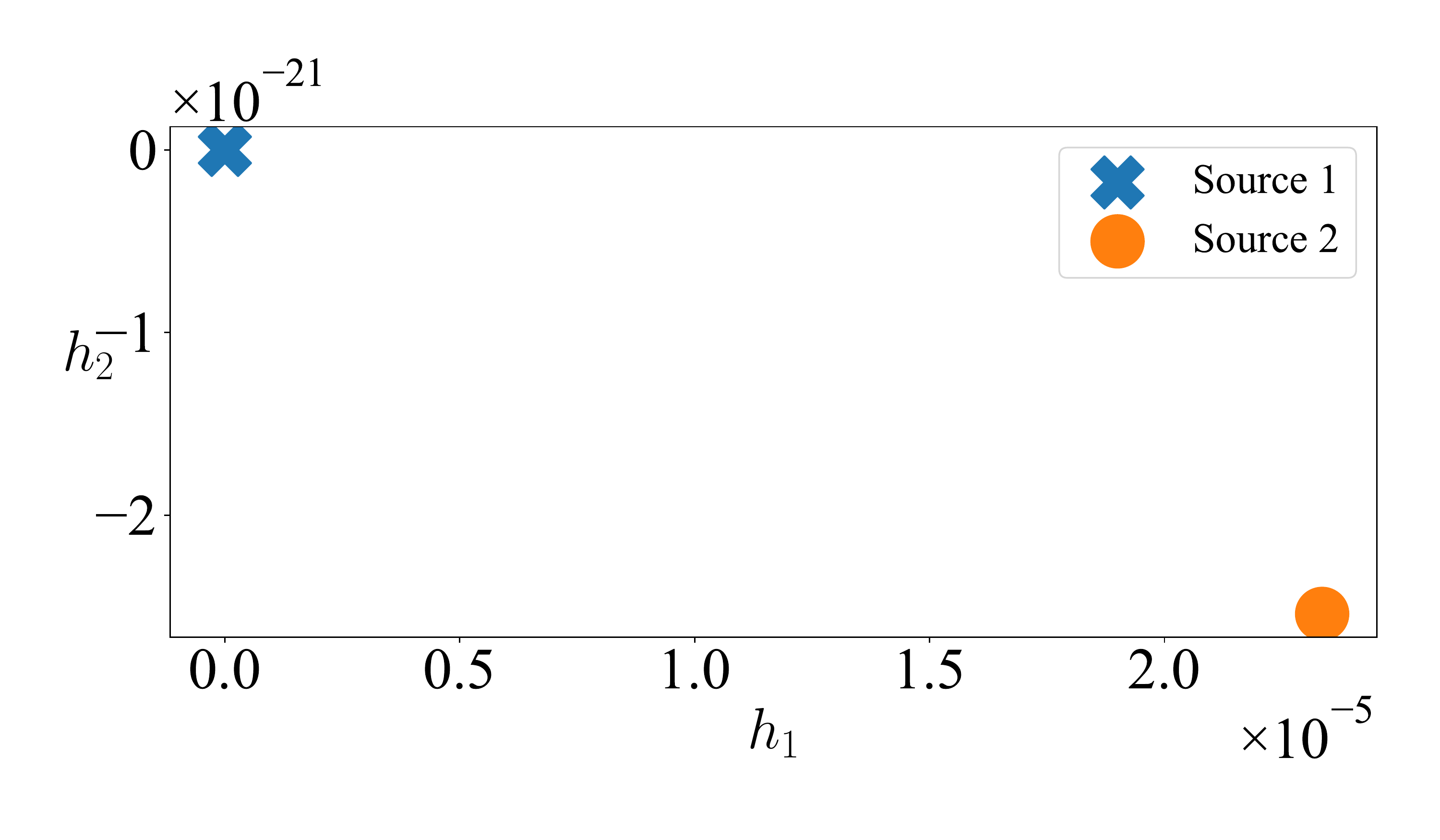}
        \caption{\rosen}
    \end{subfigure}
    \newline
    \begin{subfigure}{0.5\textwidth}
        \centering
        \includegraphics[width=1\linewidth]{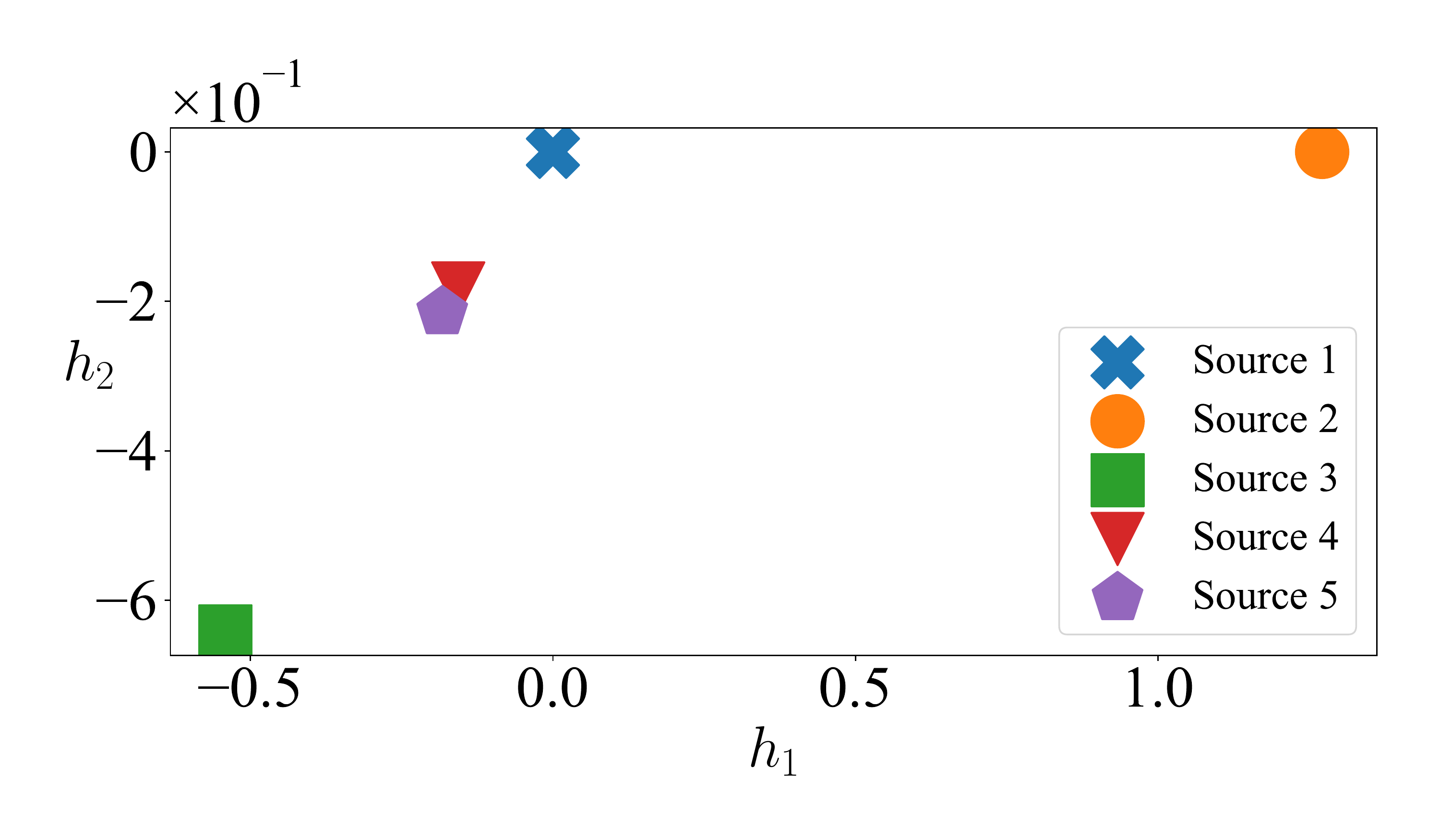}
        \caption{\borehole}
        \label{fig: Borehole_LS}
    \end{subfigure}%
    \begin{subfigure}{.5\textwidth}
        \centering
        \includegraphics[width=1\linewidth]{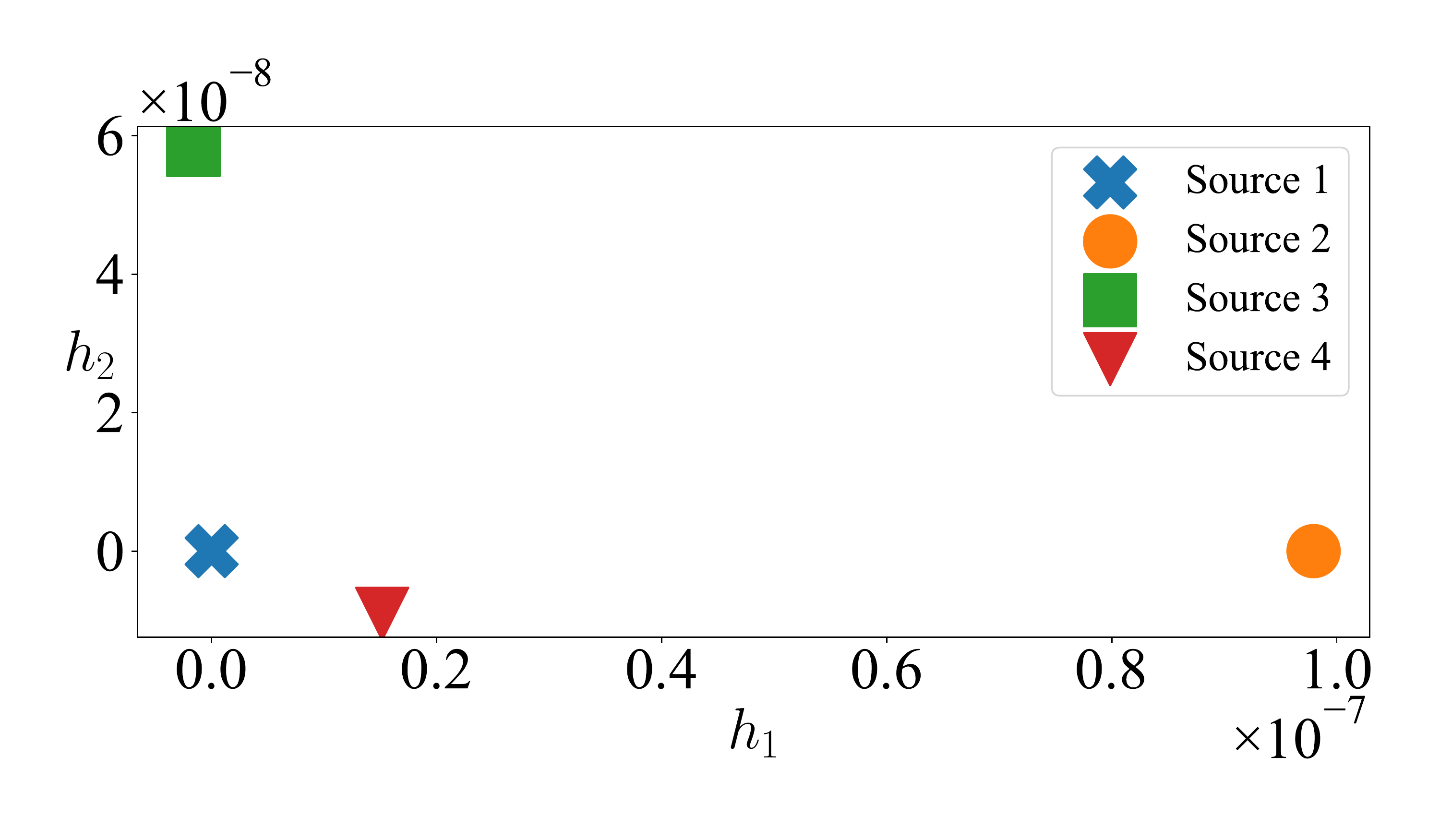}
        \caption{\wing}
    \end{subfigure}
    \caption{\textbf{Fidelity manifolds of analytic examples:} We train LMGPs on the initial multi-source data to (inversely) learn the relation between the sources. The encoding described in Section \ref{sec: BO-LMGP} is used and hence each source is represented with a point in the fidelity manifold of LMGP. For each example, the plot corresponds to a randomly selected case among the $20$ repetitions (only one plot is shown due to consistency across the repetitions). Based on the distances in the fidelity manifold (see also Eq. \ref{eq: LMGP-Corelation-extended2}), we conclude that the second and third sources are highly biased in \texttt{Borehole} and hence must be excluded from the MF BO process.}
    \label{fig: LMGP_ALL4_LS}
\end{figure}

Table \ref{table: analytic-formulation} also enumerates the relative accuracy of the LF sources of each example by calculating the relative root mean squared error (RRMSE) between them and the corresponding HF source based on $1000$ samples (these RRMSEs are not used in BO in any way). In the case of \borehole ~we observe that, unlike \wing, the source ID, true fidelity level (based on the RRMSEs), and sampling cost are not related. For instance, the first LF source is the least accurate and most expensive among all the LF sources in \borehole.

Per Step $0$ in Algorithm \ref{alg: MFBO-algorithm}, we always use the initial data to train an LMGP to identify the useful LF sources. Based on Figure \ref{fig: LMGP_ALL4_LS}, we expect the LF sources to be beneficial in all the examples except for \borehole ~since the latent points of the first and second sources are distant from the latent position of the HF source (we test this expectation in Appendix \ref{sec: bo-5source}). Hence, hereafter we exclude these highly biased LF sources, i.e., both \LMGP ~and \BOT ~use three sources (HF, LF3, and LF4 in Table \ref{table: analytic-formulation}) to optimize \borehole. We note that \BOT ~does not provide any mechanism to detect highly biased LF sources, but we also leverage this LMGP-based insight to boost the performance of \BOT ~and, in turn, better highlight the effectiveness of our proposed AF. 

\begin{figure}[!b]
    \centering
    \begin{subfigure}{.5\textwidth}
        \centering
        \includegraphics[width=1\linewidth]{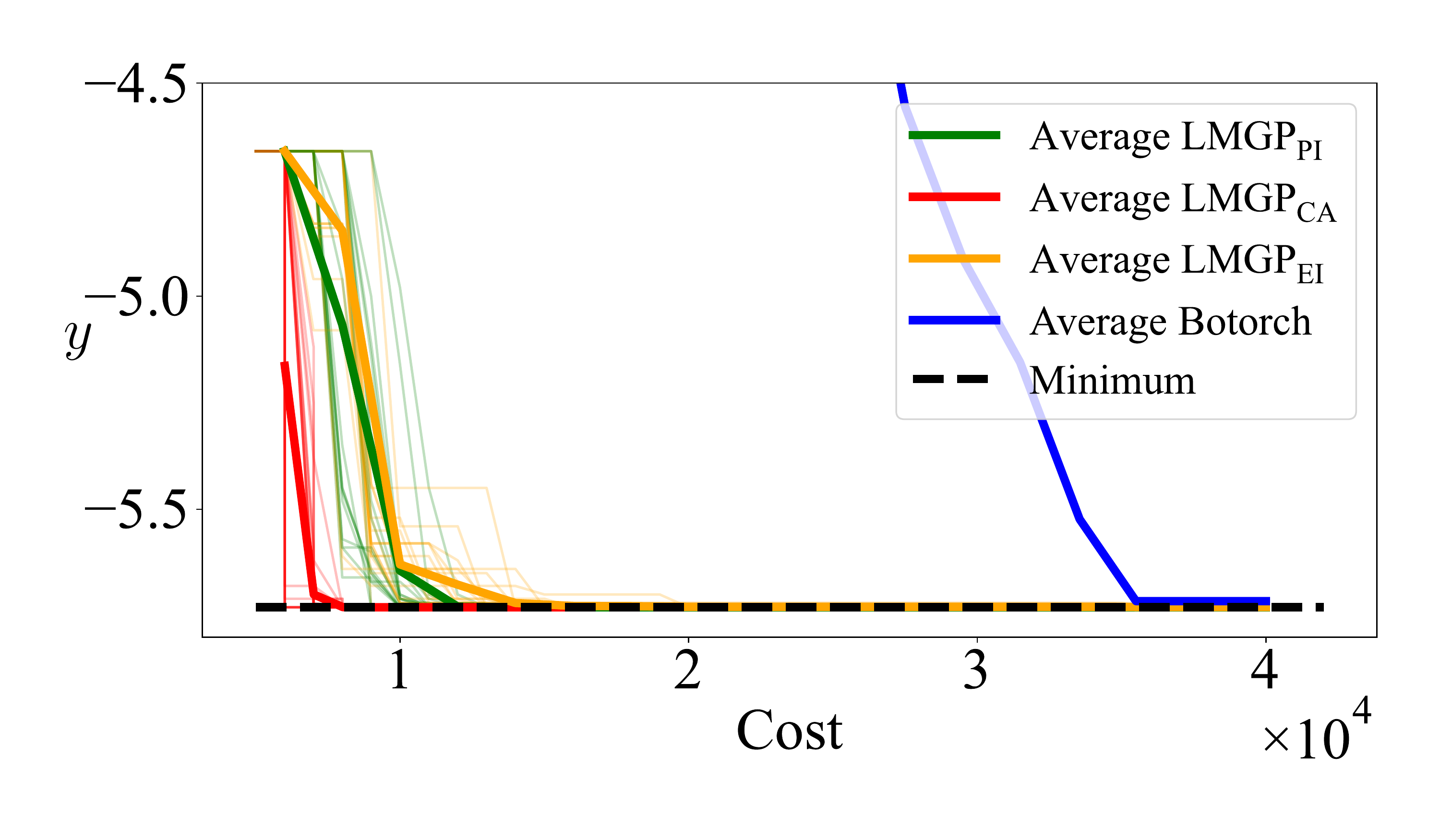}
        \caption{\potential}
        \label{fig: 1d-convergence_vs_cost}
    \end{subfigure}%
    \begin{subfigure}{.5\textwidth}
        \centering
        \includegraphics[width=1\linewidth]{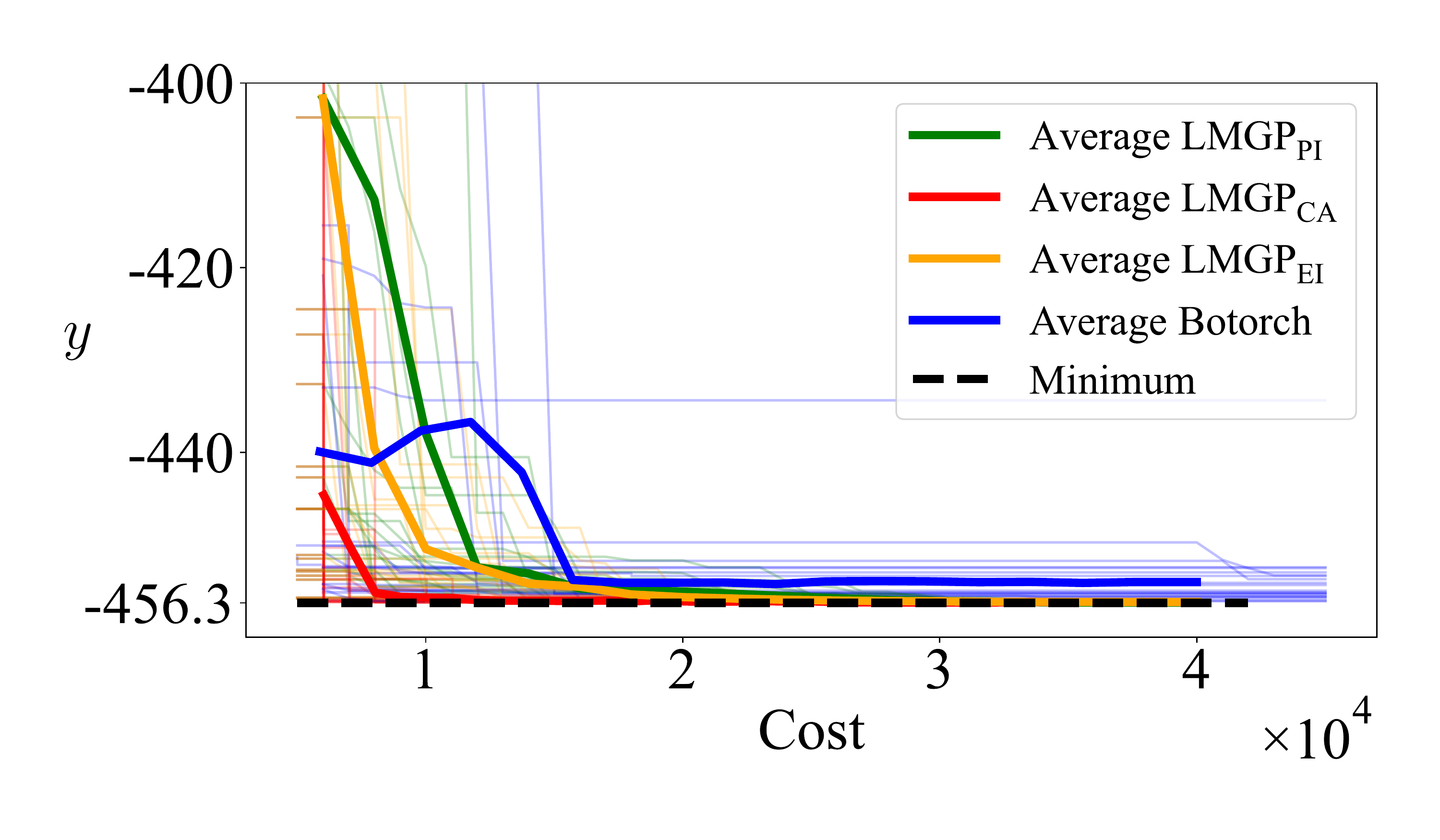}
        \caption{\rosen}
        \label{fig: rosen_convergence_vs_cost}
    \end{subfigure}
    \newline
    \begin{subfigure}{0.5\textwidth}
        \centering
        \includegraphics[width=1\linewidth]{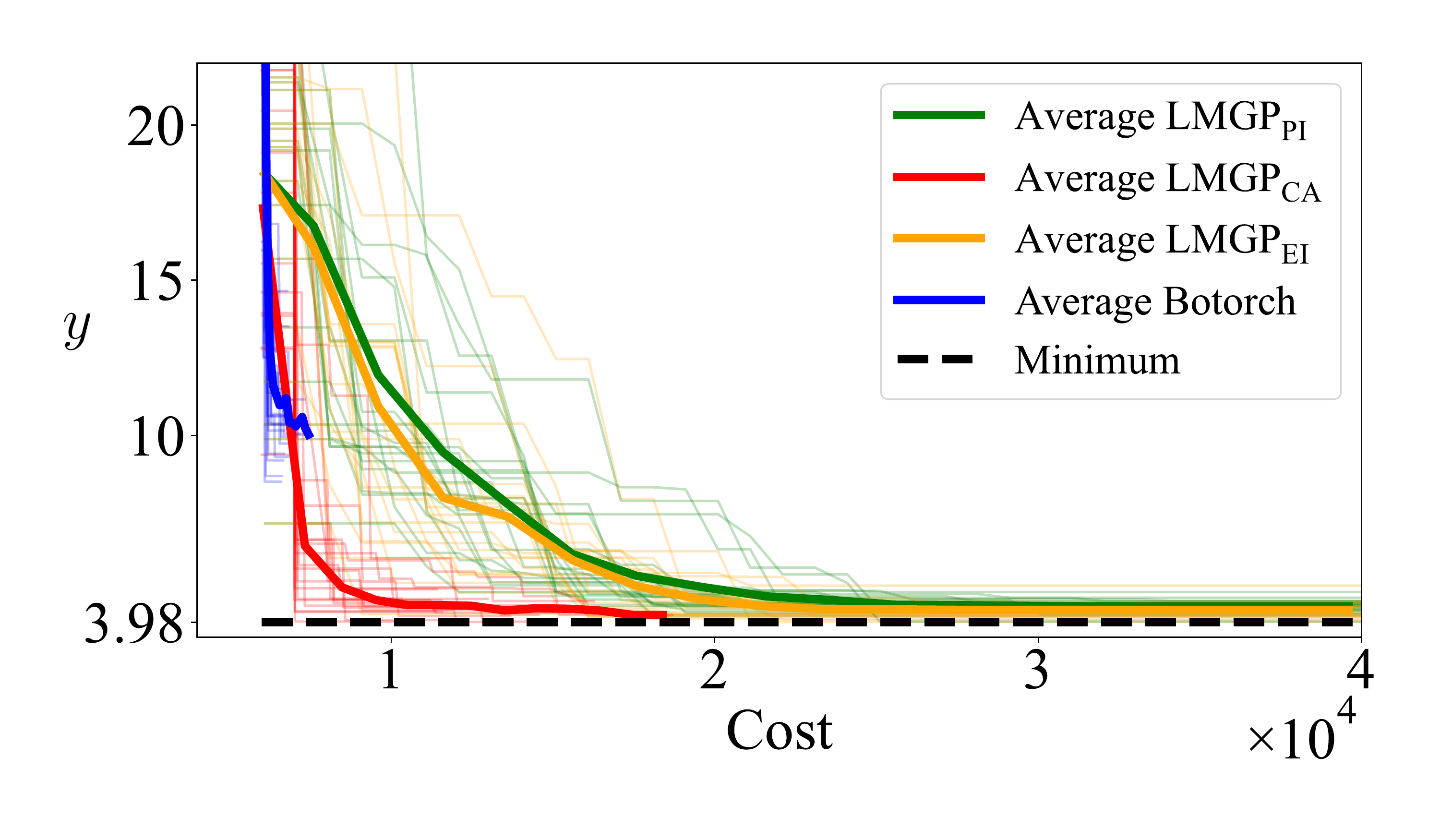}
        \caption{\borehole}
        \label{fig: Borehole_convergence_vs_cost}
    \end{subfigure}%
    \begin{subfigure}{.5\textwidth}
        \centering
        \includegraphics[width=1\linewidth]{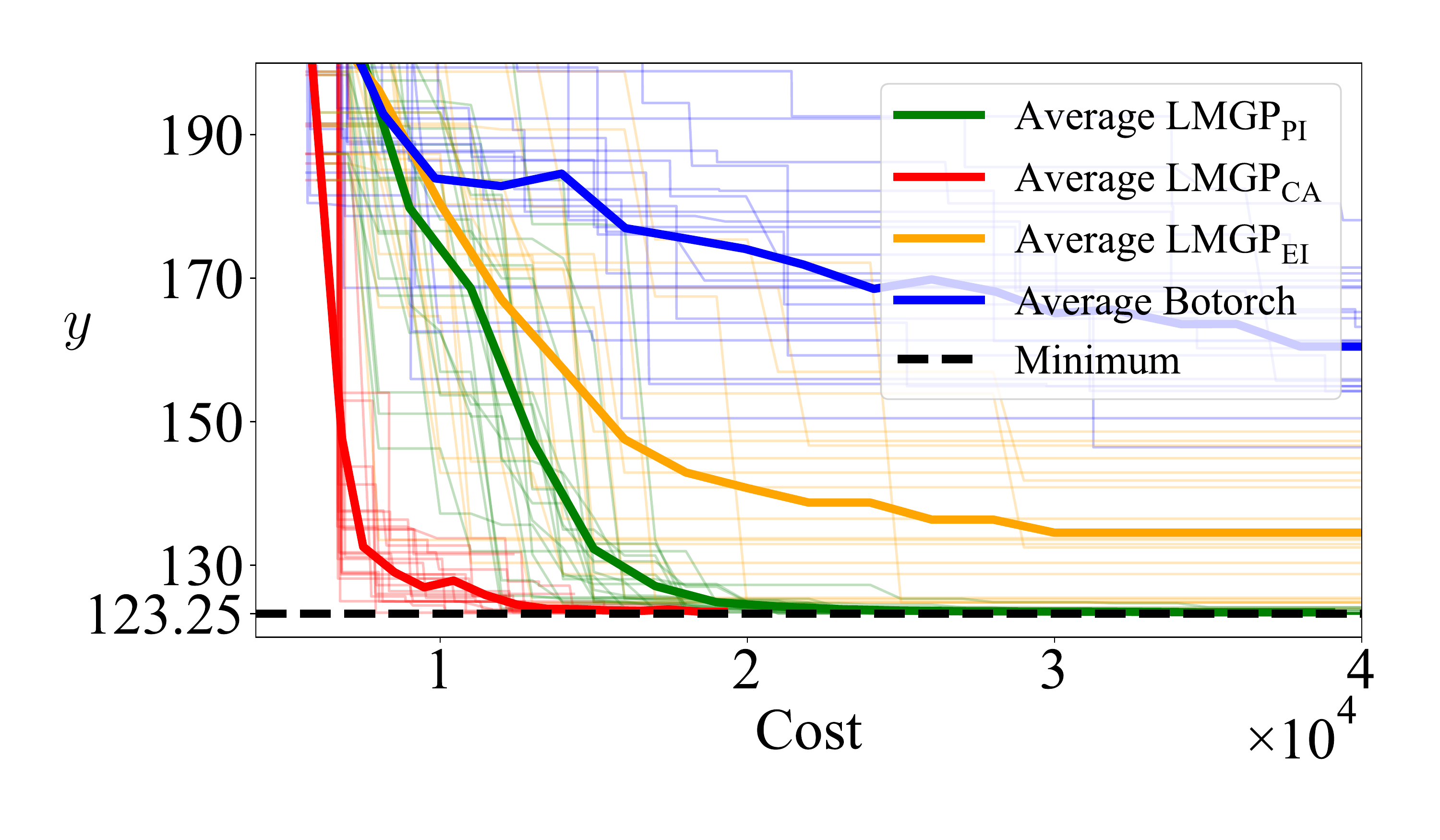}
        \caption{\wing}
        \label{fig: wing_convergence_vs_cost}
    \end{subfigure}
    \caption{\textbf{Convergence histories}: The plots illustrate the best HF sample (i.e., $y_l^*$ in Eq. \ref{eq: HF-AF}) found by each method as a function of sampling costs accumulated during the BO iterations (the cost of initial data is included). \LMGP ~consistently outperforms other methods in all the examples, especially in high-dimensional cases (i.e., \borehole ~and \wing). The solid thick curves indicate the average behavior across the $20$ repetitions (the variations associated with \BOT ~in \potential ~are extremely small and hence not visible). \LMGP ~and \BOT ~use three sources in \borehole. The second termination condition (i.e., maximum of $50$ BO iterations without improvement in $y_l^*$) is disabled for \BOT ~in \wing ~to illustrate its convergence trajectory.}
    \label{fig: convergence_vs_cost}
\end{figure}

Figure \ref{fig: convergence_vs_cost} summarizes the convergence histories by tracking the best HF estimate found by each method (i.e., $y_l^*$ in Eq. \ref{eq: HF-AF}) as a function of accumulated sampling costs. It is evident that \LMGP ~consistently outperforms all other methods across the four examples. In particular, \LMGP ~demonstrates the advantage of leveraging inexpensive LF sources in BO by accelerating the convergence without sacrificing the accuracy (compare \LMGP ~to \MEI ~and \MPI for any of the examples in Figure \ref{fig: convergence_vs_cost}). 
Additionally, unlike \BOT, the performance of \LMGP ~is robust to the input dimensionality and sampling costs. For instance, \BOT ~estimates the optimum as $y_l^* = 10.02$ in Figure \ref{fig: Borehole_convergence_vs_cost} while the ground truth is $3.98$. The reason behind this inaccuracy is that \BOT ~fails to find an HF sample whose information value is large enough to overcome its high sampling cost and, as a result, cheap LF sources are largely queried. 
However, these queries do not improve $y_l^*$ and hence the second strop condition (maximum number of repetitions without improvement of $y_l^*$) terminates \BOT ~after $50$ iterations. Removing this stop condition, while significantly increasing the number of iterations, does not improve the accuracy of \BOT. To demonstrate the effect of this removal, we only consider the maximum budget constraint for \BOT ~in \wing ~and observe that the ground truth is again not found, see Figure \ref{fig: wing_convergence_vs_cost}. 
These issues are resolved in \LMGP ~by the intricate interplay between $\Phi(z)$ and $\phi(z)$ as explained in Section \ref{sec: proposed-AF}.

\begin{figure}[!t]
    \centering
    \begin{subfigure}{.5\textwidth}
        \centering
        \includegraphics[width=1\linewidth]{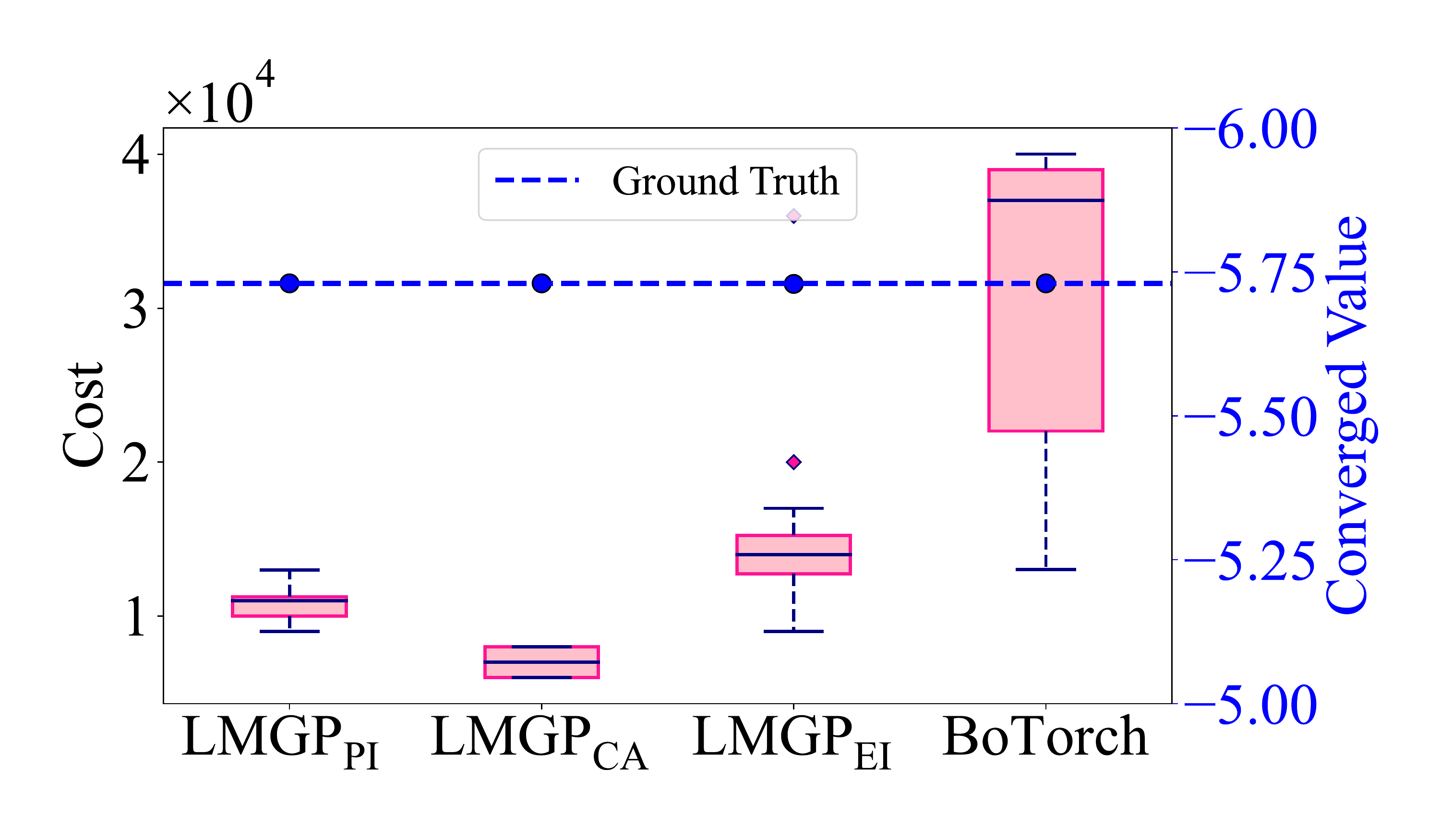}
        \caption{\potential}
        \label{fig: 1d-vox_cost}
    \end{subfigure}%
    \begin{subfigure}{.5\textwidth}
        \centering
        \includegraphics[width=1\linewidth]{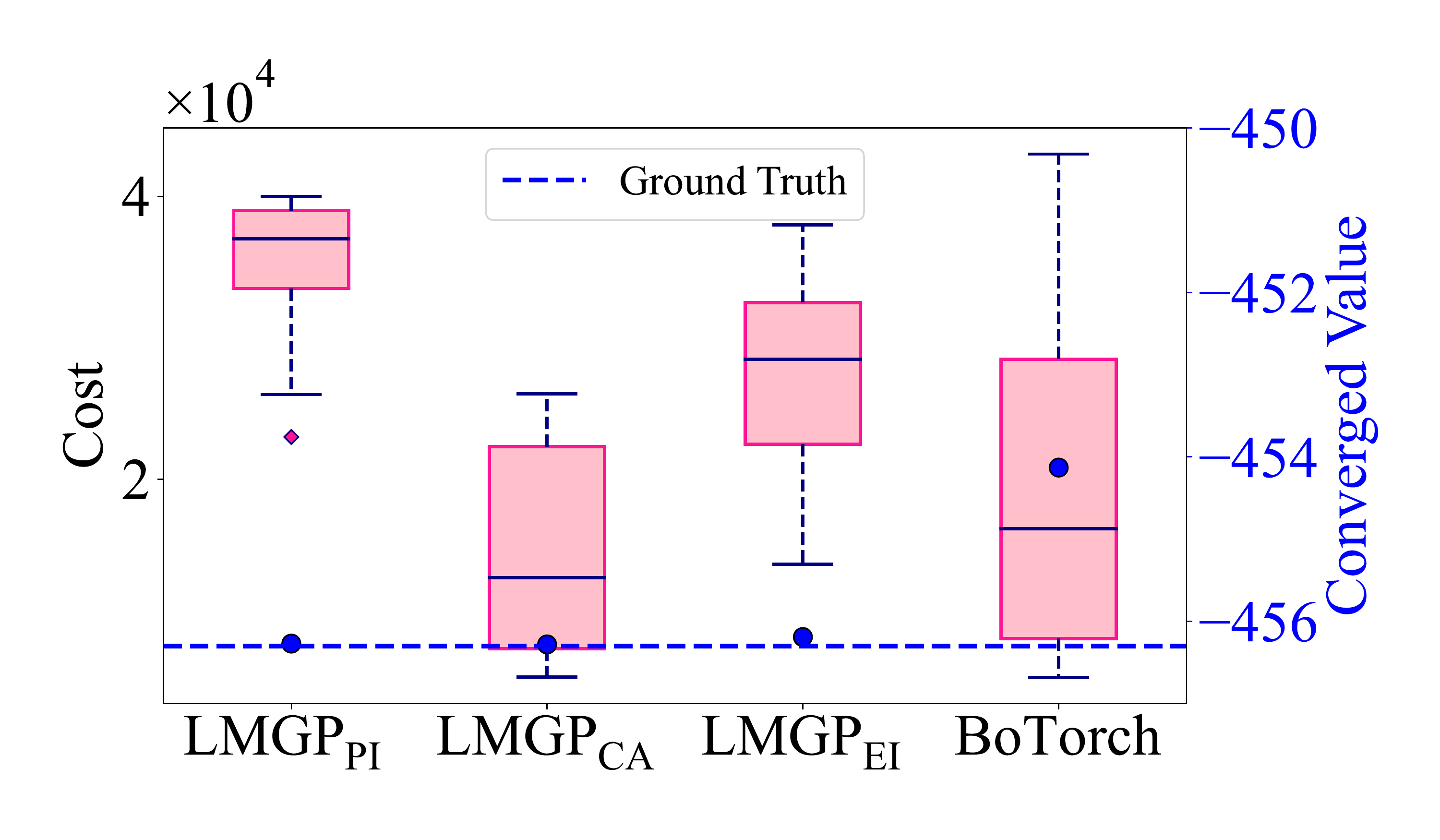}
        \caption{\rosen}
        \label{fig: rosen_box_cost}
    \end{subfigure}
    \newline
    \begin{subfigure}{0.5\textwidth}
        \centering
        \includegraphics[width=1\linewidth]{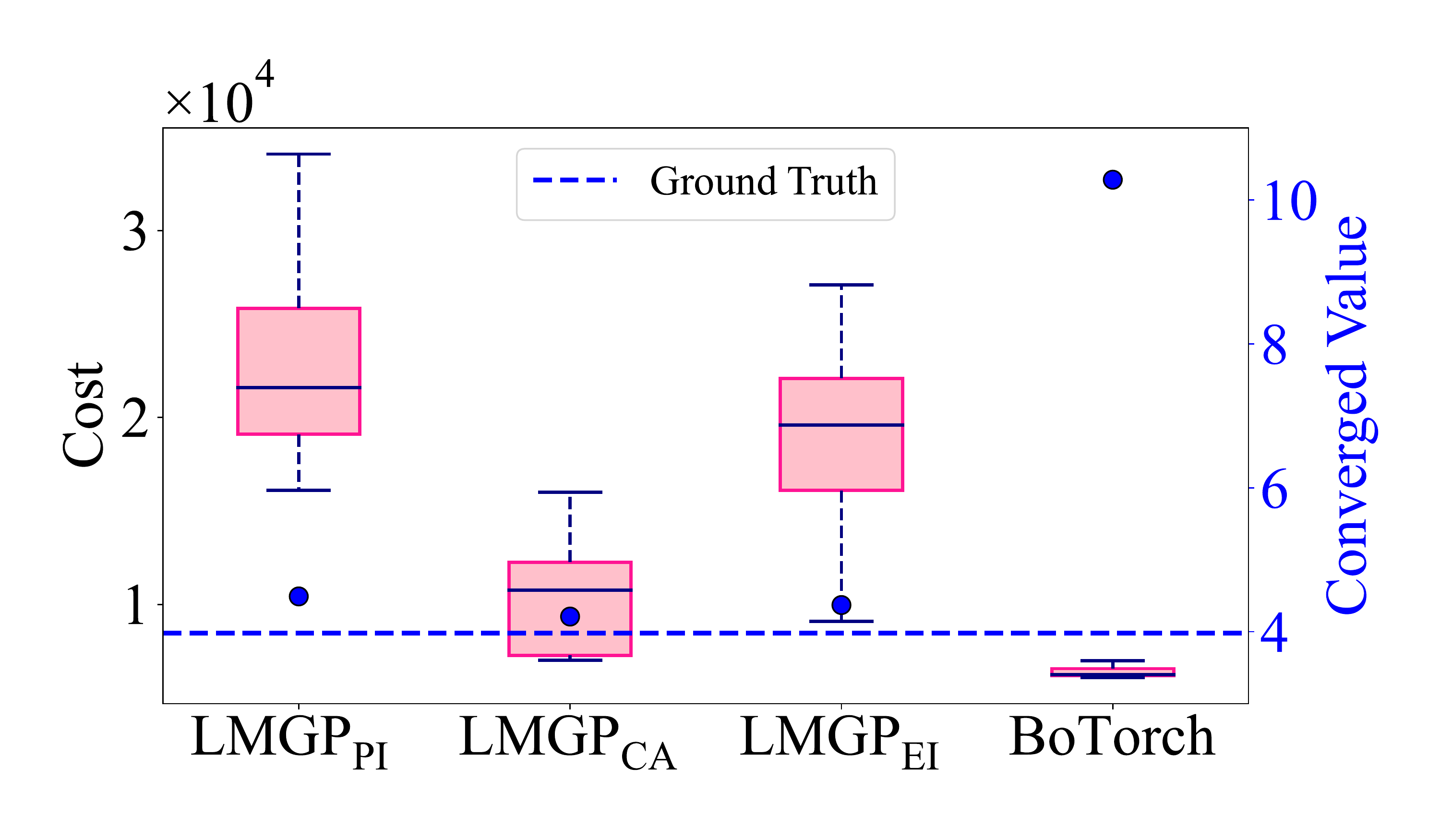}
        \caption{\borehole}
        \label{fig: Borehole_box-cost}
    \end{subfigure}%
    \begin{subfigure}{.5\textwidth}
        \centering
        \includegraphics[width=1\linewidth]{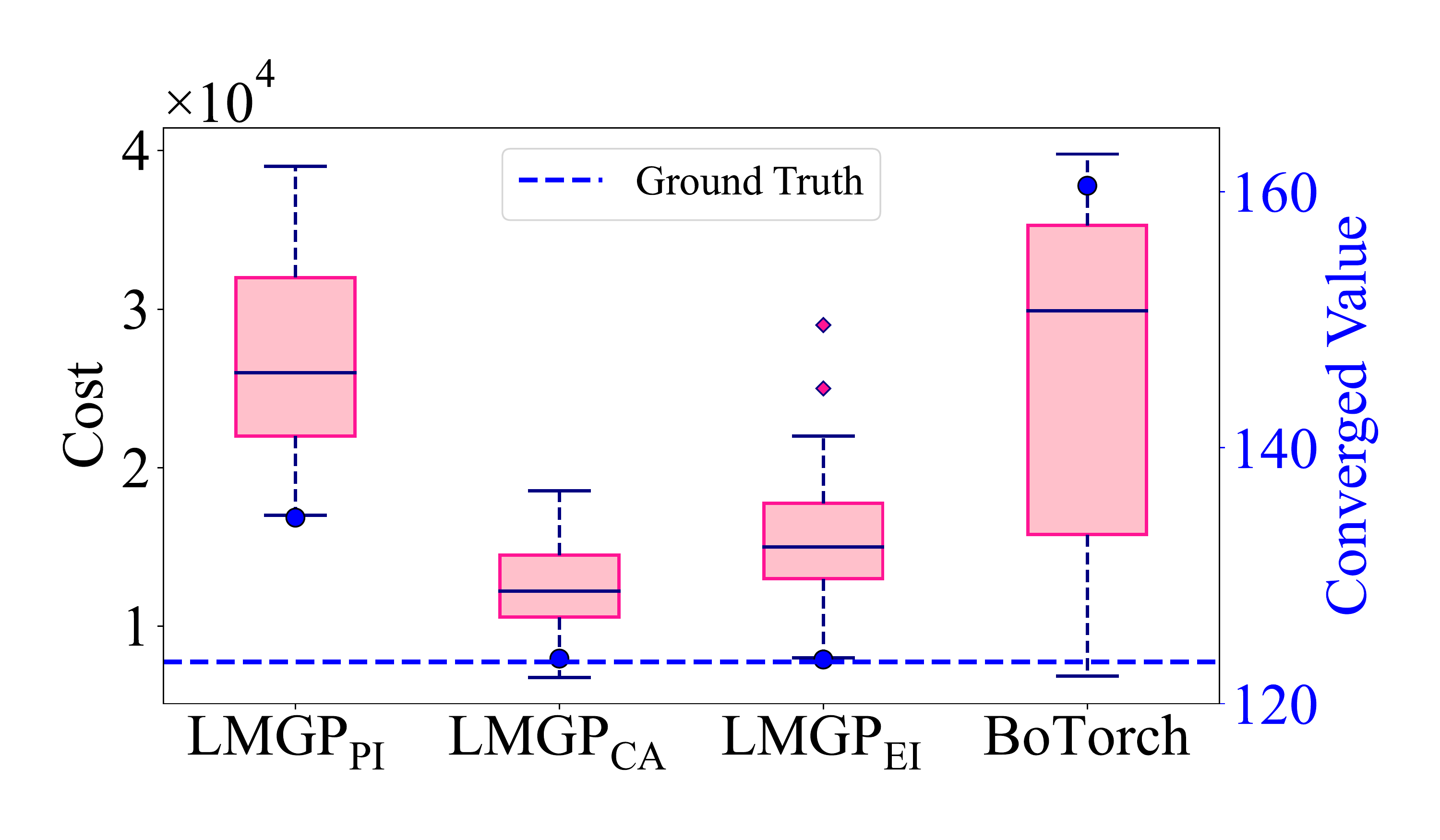}
        \caption{\wing}
        \label{fig: wing_box_cost}
    \end{subfigure}
    \caption{\textbf{Accumulated costs before improvements plateau:} The box-plots illustrate the accumulated costs up to and including the iteration at which the best HF sample is first obtained (i.e., these box-plots do not consider the two termination criteria). On the right axis, the converged solution (averaged across the $20$ repetitions) and ground truth are demonstrated via, respectively, the blue marker and the horizontal dashed line. In all four examples, \LMGP ~finds the optimum faster than other methods. Comparison between the axes indicates that small accumulated costs do not imply superior performance since the converged solution might be a local optimum, as is the case for \BOT ~in \ref{fig: Borehole_box-cost}.}
    \label{fig: box-plot cost}
\end{figure}

To exclude the effect of the termination criteria from the results, we provide the accumulated cost up to and including the iteration at which each method finds its best estimate, see Figure \ref{fig: box-plot cost}. In terms of finding the true optimum (compare the blue dots to the horizontal dashed line), \LMGP outperforms all other methods and is followed by \MEI ~and then \MPI (the high accuracy of the SF methods in finding the ground truth is expected since they only sample from the HF source and neither of the two termination criteria are stringent). However, \BOT ~either terminates at an incorrect solution or is even costlier than SF methods. This poor performance is expected since, even though \BOT ~is lookahead, its AF cannot handle the high cost-ratios across the sources and its emulator does not effectively learn the nonlinear relations between the HF and LF sources.

\begin{figure}[!t]
    \centering
    \begin{subfigure}{.5\textwidth}
        \centering
        \includegraphics[width=1\linewidth]{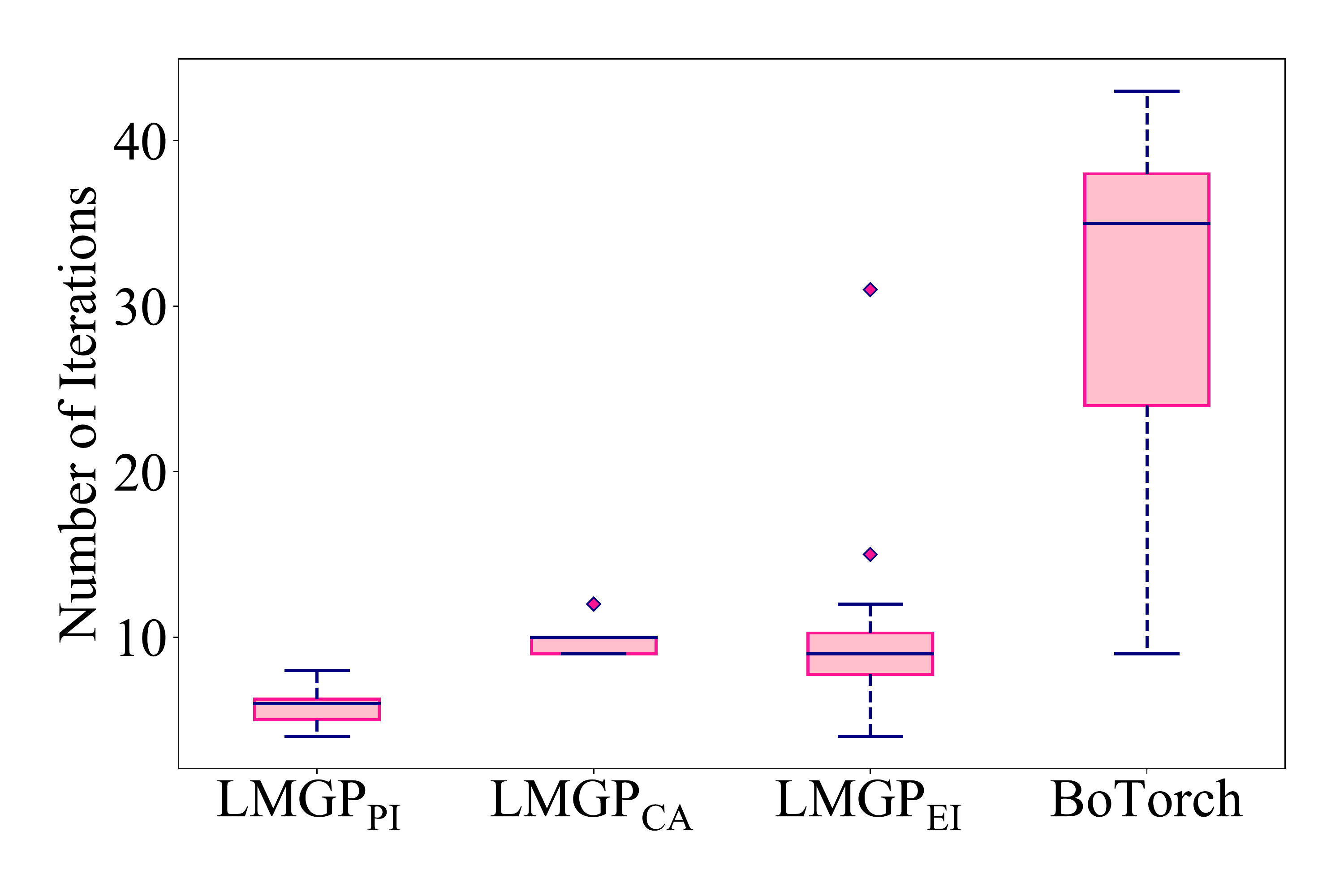}
        \caption{\potential}
        \label{fig: 1d-vox_iter}
    \end{subfigure}%
    \begin{subfigure}{.5\textwidth}
        \centering
        \includegraphics[width=1\linewidth]{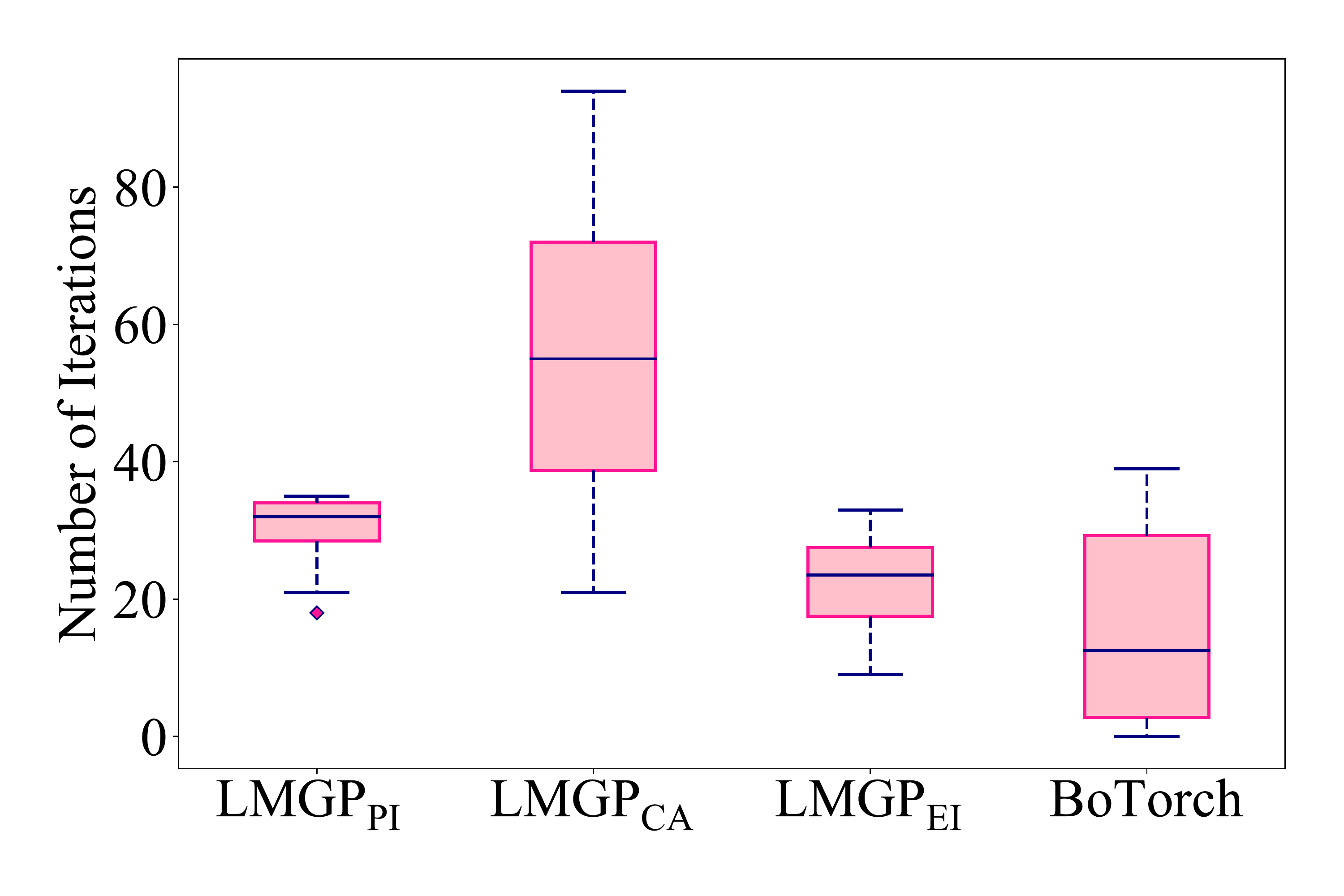}
        \caption{\rosen}
        \label{fig: rosen_box_iter}
    \end{subfigure}
    \newline
    \begin{subfigure}{0.5\textwidth}
        \centering
        \includegraphics[width=1\linewidth]{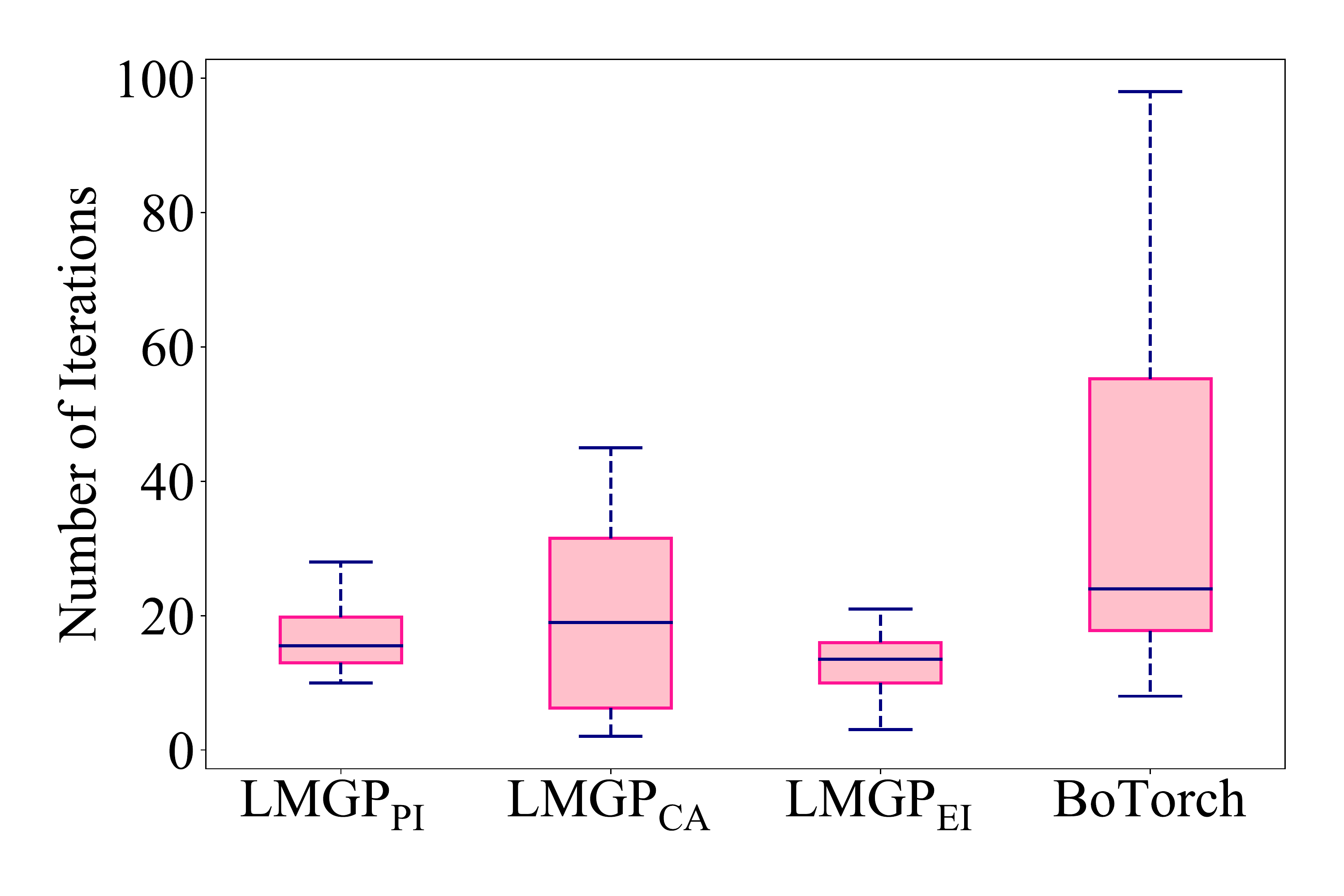}
        \caption{\borehole}
        \label{fig: Borehole_box-iter}
    \end{subfigure}%
    \begin{subfigure}{.5\textwidth}
        \centering
        \includegraphics[width=1\linewidth]{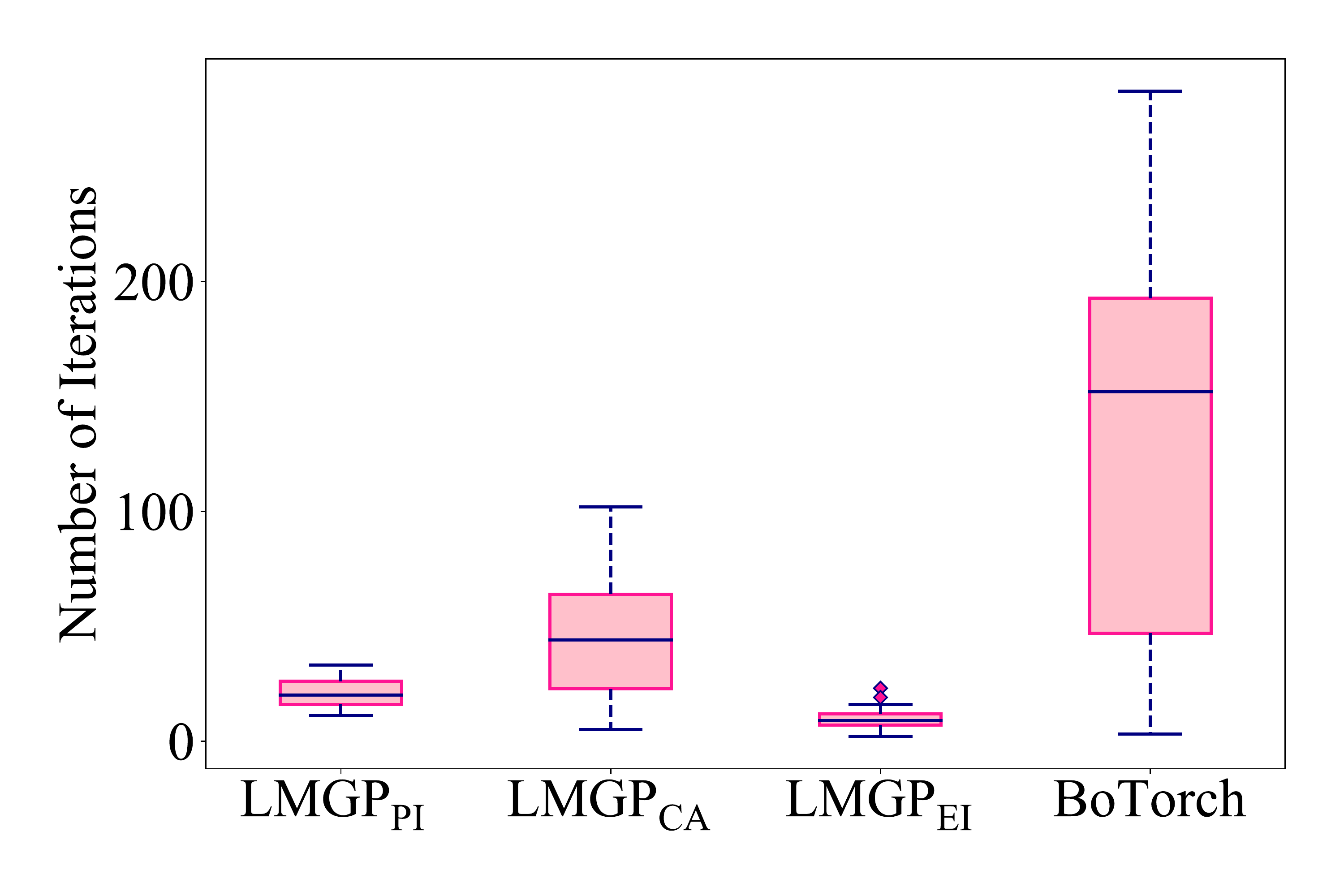}
        \caption{\wing}
        \label{fig: wing_box_iter}
    \end{subfigure}
    \caption{\textbf{Number of iterations at convergence:} As expected, \BOT ~and \LMGP ~need more iterations to converge compared to the SF methods. However, the difference in the case of \LMGP ~is very small as our method is quite efficient in leveraging the LF sources. It is noted that, since only one sample is obtained per iteration, these plots are also representative of the total number of samples collected via BO.}
    \label{fig: box-plot iteration}
\end{figure}

To provide more insight into the mechanics of MF BO methods, we also report the number of iteration at convergence, see Figure \ref{fig: box-plot iteration}. As expected, \BOT ~and \LMGP ~require more iterations as they aim to leverage cheap LF sources to reduce the overall costs. \LMGP ~needs fewer iterations for convergence than \BOT ~except in Figure \ref{fig: rosen_box_iter}. This behavior is a result of the termination criteria: the ground truth of \rosen ~is $-456.3$ while \LMGP ~rapidly converges to a very close solution ($-456$) and then takes highly cheap LF samples to improve the best sample. These improvements are quite small ($0.01$ per iteration) and hence many iterations are needed for convergence. 

\begin{figure}[!h]
    \centering
    \begin{subfigure}{1\textwidth}
        \centering
        \includegraphics[width=0.7\linewidth]{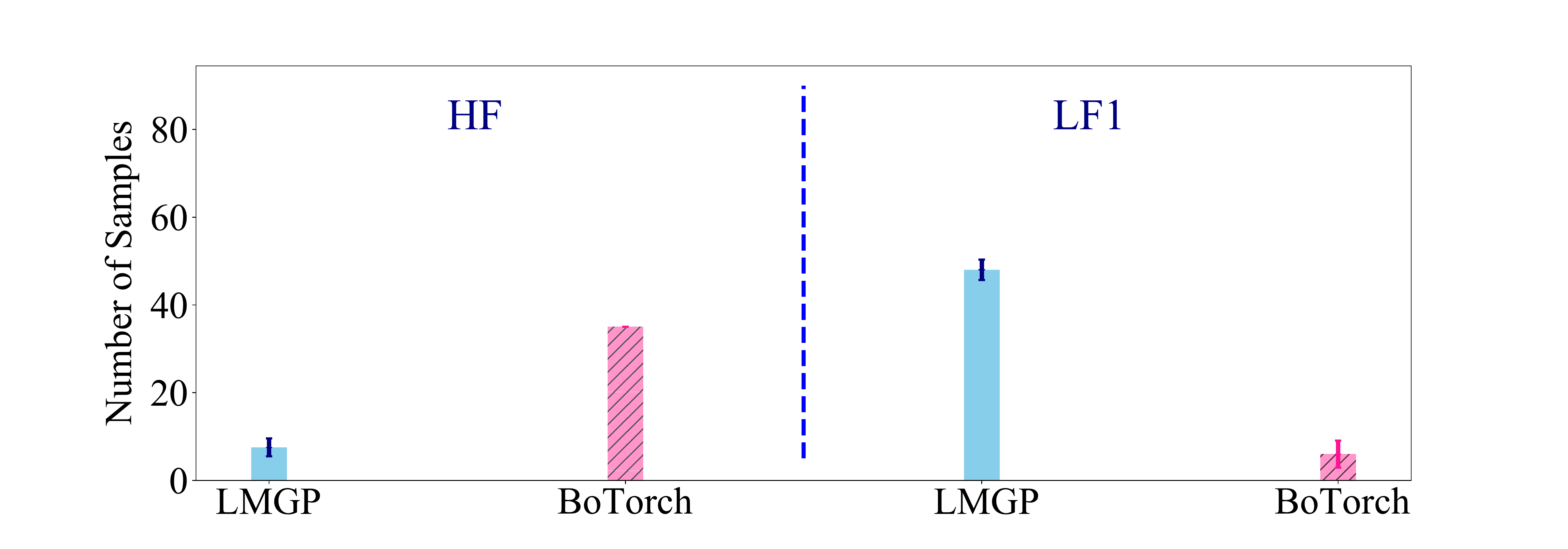}
        \caption{\potential}
        \label{fig: 1d-error_lmgp}
    \end{subfigure}%
    \newline
    \begin{subfigure}{1\textwidth}
        \centering
        \includegraphics[width=0.7\linewidth]{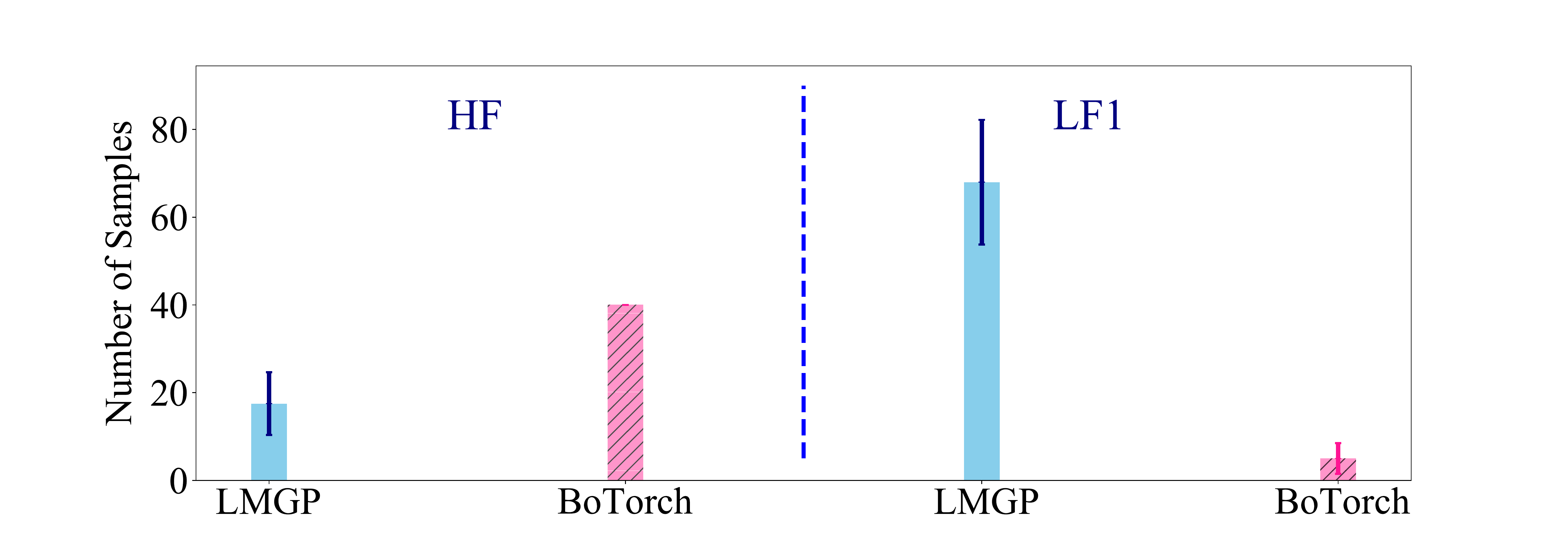}
        \caption{\rosen}
        \label{fig: rosen_error_lmgp}
    \end{subfigure}%
    \newline
    \begin{subfigure}{1\textwidth}
        \centering
        \includegraphics[width=0.7\linewidth]{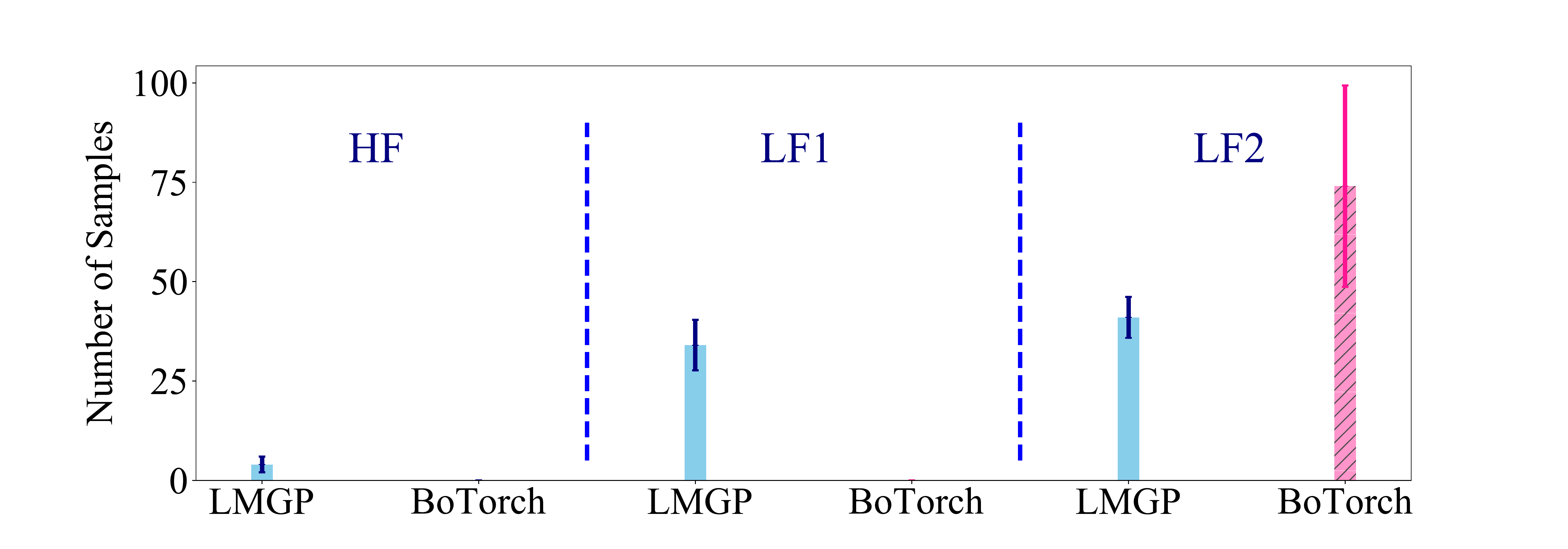}
        \caption{\borehole}
        \label{fig: borehole-error_lmgp}
    \end{subfigure}%
    \newline
    \begin{subfigure}{1\textwidth}
        \centering
        \includegraphics[width=0.7\linewidth]{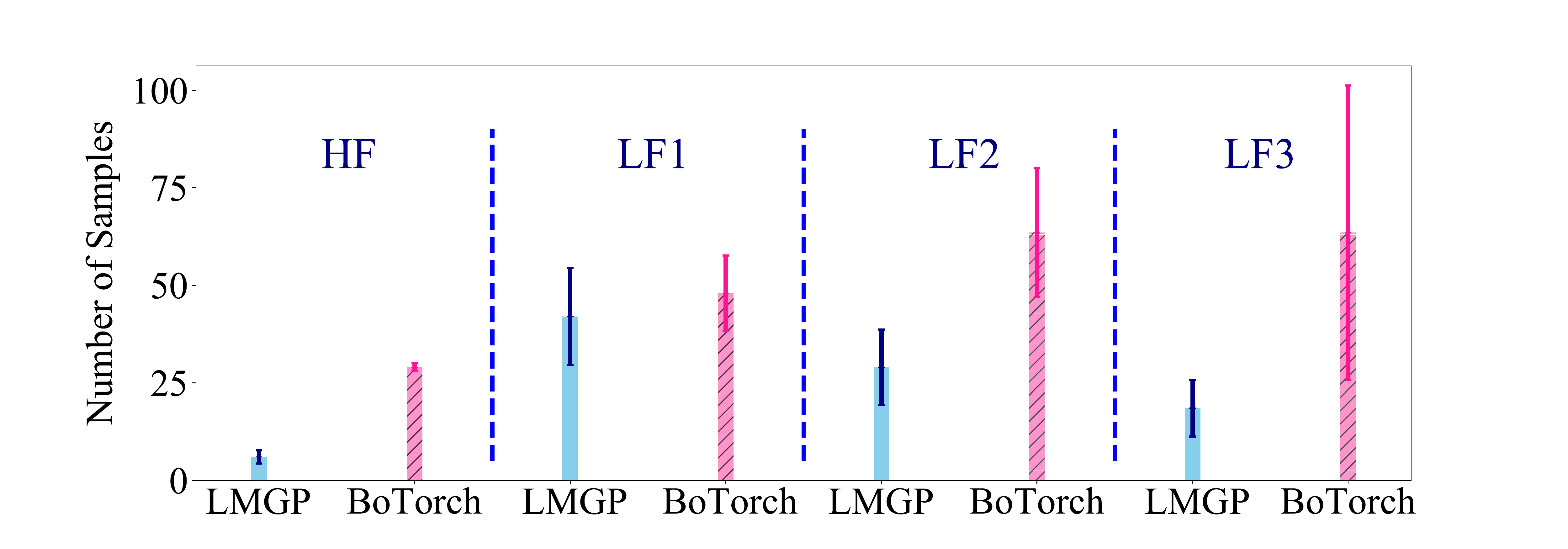}
        \caption{\wing}
        \label{fig: wing_error_lmgp}
    \end{subfigure}%
    \caption{\textbf{Number of samples taken from each source:} As opposed to \BOT, \LMGP ~optimally and automatically adjust the sampling frequency from each source. For instance, \BOT ~does not sample from two sources in \borehole ~since they are much more expensive than the second LF source. However, \LMGP ~not only samples from all sources, but also adjusts the sampling frequency from the LF sources based on their relative accuracy, initial data, and cost (note that LF2 is ten times cheaper to query than LF1 in \borehole).}
    \label{fig: source-wise sampling-analytic datasets}
\end{figure}

We highlight that, as long as the associated costs are low, a large number of iterations is not reflective of bad performance since the goal of MF BO is to reduce the overall sampling costs and not necessarily the total number of samples. To demonstrate this, we provide the per-source sampling frequencies for \LMGP ~and \BOT ~in Figure \ref{fig: source-wise sampling-analytic datasets} which demonstrates that \LMGP ~automatically adjusts its sampling mechanism based on the initial data and the relative accuracy of the LF sources (compared to the corresponding HF source) and their costs. For instance, unlike \BOT ~ which avoids querying the HF source in \borehole, \LMGP ~leverages all sources where the number of samples taken from each source depends on its cost and (in the case of LF sources) relative accuracy. In particular, we observe that \LMGP ~samples almost equally from the LF sources even though the second source is $10$ times cheaper (note that these LF sources correspond to the third and fourth sources in Table \ref{table: analytic-formulation}). This behavior may seem undesirable at the first glance especially since the LF sources have the same relative accuracy (see RRMSEs in Table \ref{table: analytic-formulation}) but a closer look indicates that it is primarily caused by the number of initial samples: since there are $10$ times more data points from the second LF source, the emulator of \LMGP ~correctly provides large prediction uncertainties which, in turn, results in a large expected utility for the first LF source, see Eq. \ref{eq: LF_AF}. 
These discussions also hold for \wing, see Figure \ref{fig: wing_error_lmgp} where, unlike \BOT, \LMGP ~samples from the LF sources based on their relative accuracy (which is learnt internally by its emulator) as well as cost.

\begin{figure*}[!t] 
    \centering
    \includegraphics[page=1, width = 1\textwidth]{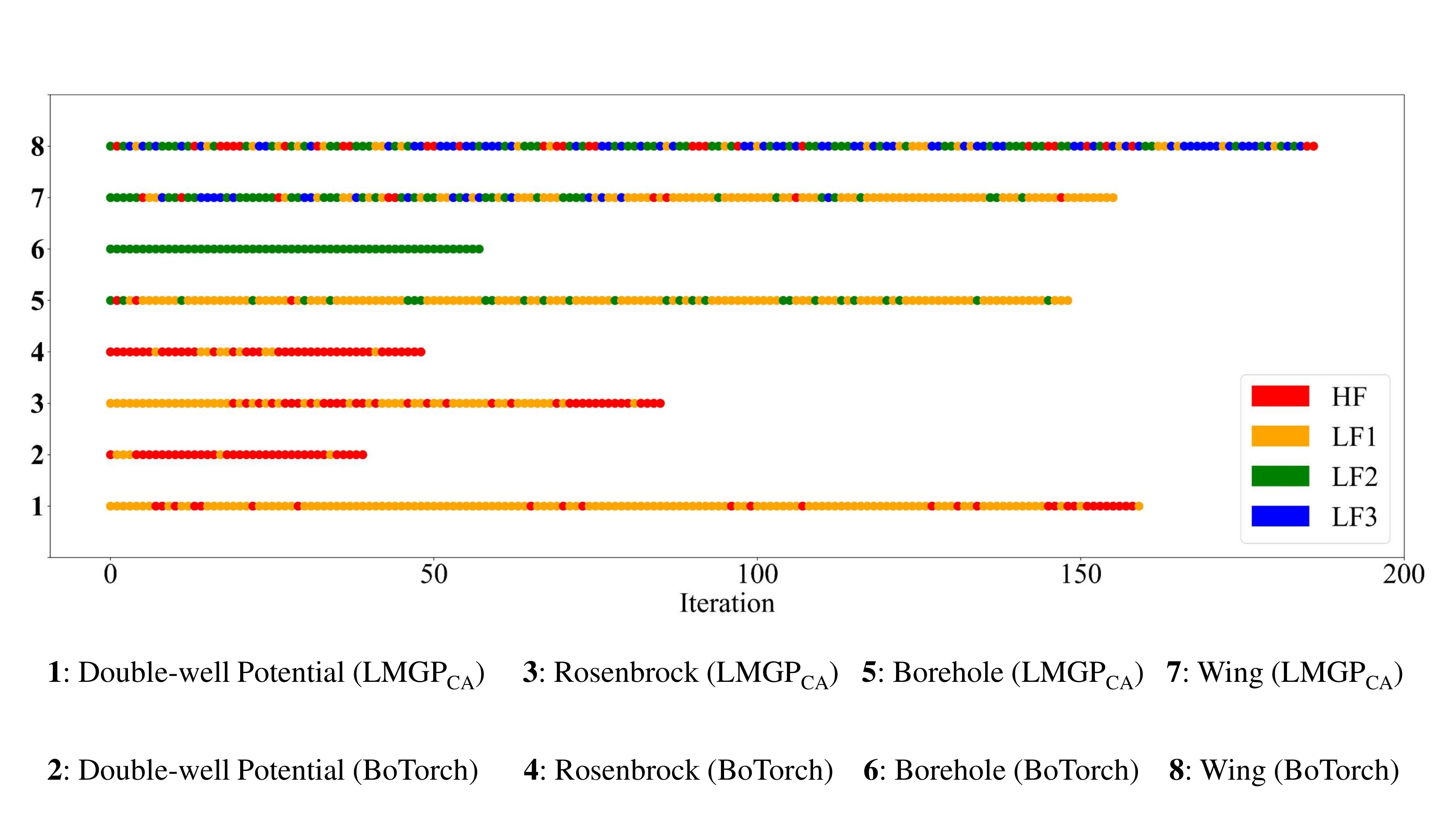}
    \vspace{-0.3cm}
    \caption{\textbf{Source-wise sampling orders:} 
    A repetition is randomly selected from each example to visualize the sampling orders made by \LMGP ~and \BOT. 
    The horizontal axis enumerates the number of BO iterations while the vertical axis denotes the example and the MF BO method used for optimization. There are $2, 2, 3 ~\text{and}~4$ data sources in each example from top to bottom. 
    Unlike \BOT, \LMGP ~balances exploration and exploitation throughout the optimization process.}
    \label{fig: color-coded-fidelity}
\end{figure*}

Finally, we demonstrate the performance of \LMGP ~in balancing exploration and exploitation. To this end, source-wise sampling orders made by \LMGP ~ and \BOT ~are visualized for a randomly selected repetition in each example, see Figure \ref{fig: color-coded-fidelity}. As it can be observed, \LMGP ~alternates between all the sources: in each example the majority of the samples are from the LF sources which are queried based on exploration (see Eq. \ref{eq: LF_AF} and Eq. \ref{eq: MFCA-AF}) while the expensive HF source is typically used with much lower frequency. 
The sampling orders are particularly interesting for \rosen ~and \borehole ~where, unlike \BOT ~which struggles to alternate between the sources, \LMGP ~effectively uses all sources during the optimization. 

\subsection{Real-world Datasets} \label{sec: real-world example}
In this Section, we study two materials design problems where the goal is to identify the composition that optimizes the property of interest. Unlike the examples in Section \ref{sec: analytical-examples}, these two problems are noisy and have categorical inputs. We compare the performance of \LMGP ~only against the SF methods since \BOT ~does not accommodate categorical inputs. In both examples, the HF and LF data are obtained via simulations (based on the density functional theory) which have different fidelity levels. 

The first problem is bi-fidelity and aims to find the member of the nanolaminate ternary alloy (\nta) family with the largest bulk modulus \cite{yeom2019performance}. The HF and LF datasets each have $224$ samples, one response, and $10$ features ($7$ quantitative and $3$ categorical where the latter have $10$, $12$, and $2$ levels). In our studies, we presume a cost ratio of $10$ to $1$ between the sources and proceed as follows: we exclude the composition with the largest bulk modulus from the HF dataset, take $20$ and $10$ samples from, respectively, the HF and LF datasets (SF methods only use HF samples), and then initiate the BO methods. We repeat this process $15$ times for each BO method to quantify its robustness to the random initial data. 

Our second problem concerns hybrid organic–inorganic perovskite (\hoip) crystals where the goal is to find the compound with the smallest inter-molecular binding energy. There are one HF and two LF datasets in this problem and their sampling costs are $15$, $10$, and $5$, respectively\footnote{We assign these sampling costs randomly to the LF sources as we do not know which one is more accurate.}. The three datasets have similar dimensionalities ($1$ output and $3$ categorical inputs with $10$, $3$, and $16$ levels) but are of different sizes: the HF dataset has $480$ samples, while the first and second LF sources have $179$ and $240$ samples, respectively. We apply the three BO methods to this problem as follows: we first exclude the best compound from the HF dataset and build the initial MF data by randomly sampling from the three datasets. Then, we launch the BO process where \MEI ~ and \MPI ~only use the HF samples. We set the size of the initial datasets to $(15, 20, 15)$ for the HF and LF sources, respectively, and repeat the BO process $15$ times to assess the repetition-size variability. 

\begin{figure}[!t]
    \centering    
    \begin{subfigure}{0.5\textwidth}
        \centering
        \includegraphics[width=1\linewidth]{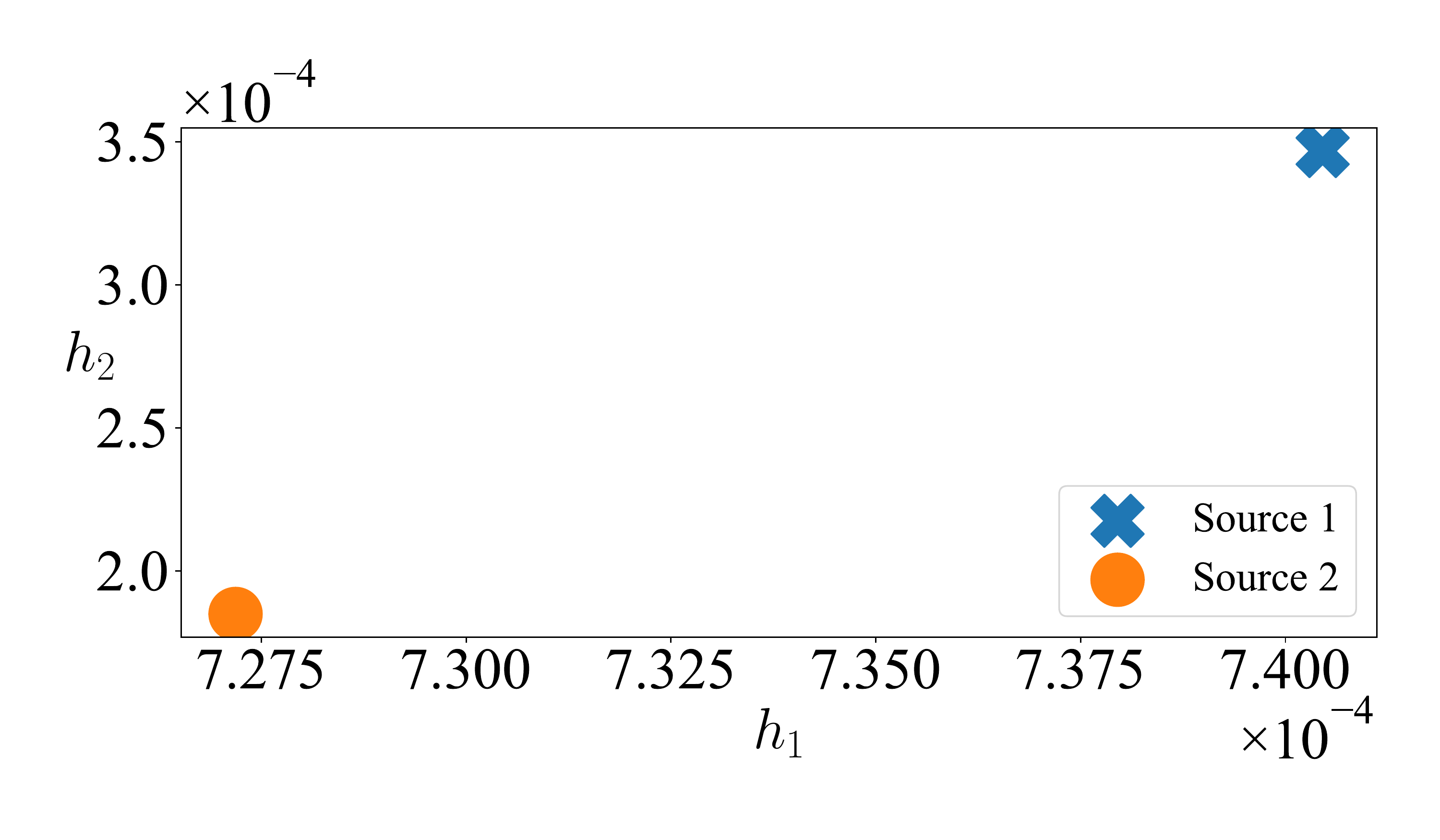}
        \caption{\nta ~(initial data)}
        \label{fig: fidelity_manifold_nta1}
    \end{subfigure}%
    \begin{subfigure}{0.5\textwidth}
        \centering
        \includegraphics[width=1\linewidth]{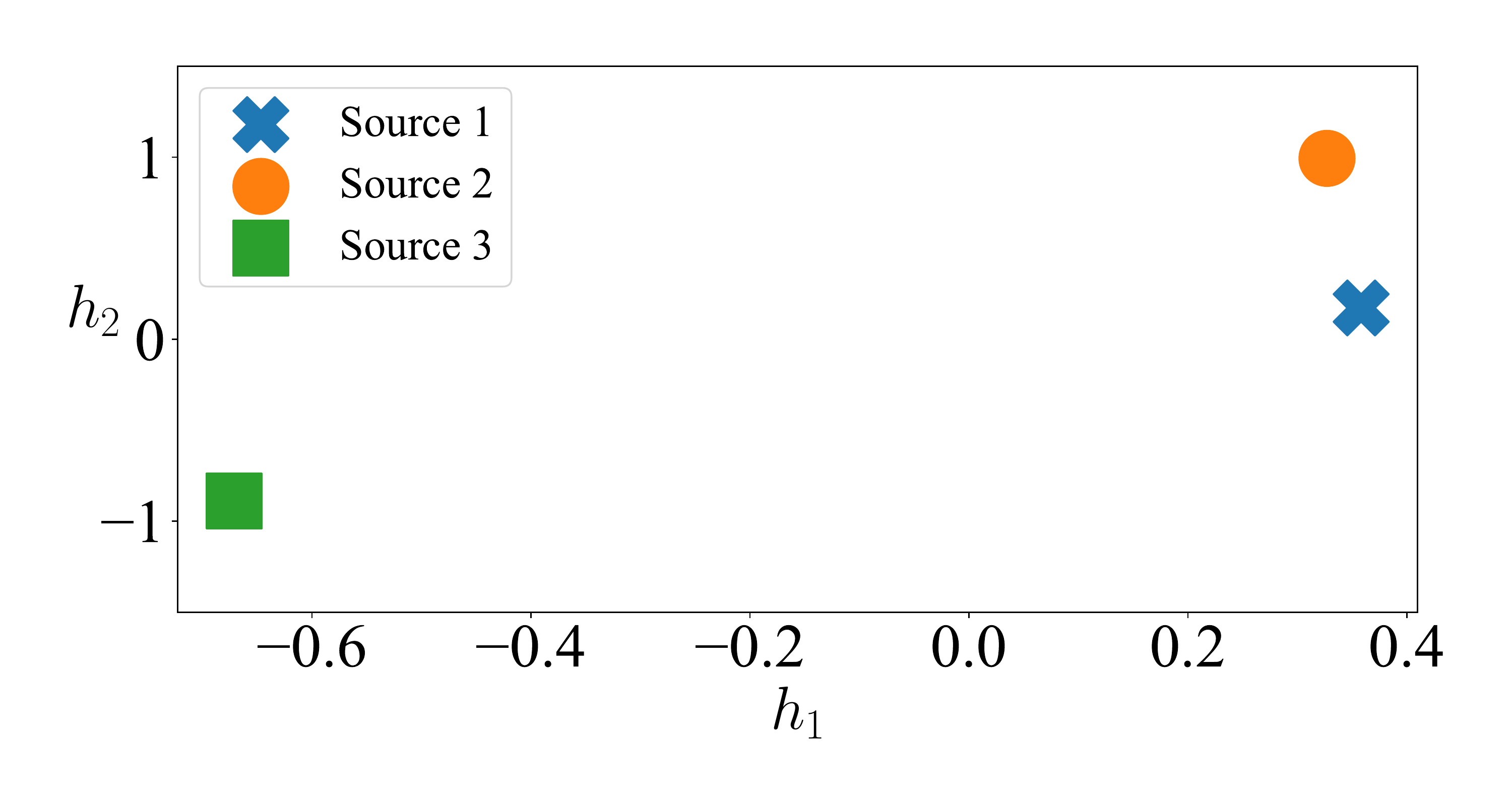}
        \caption{\hoip ~(initial data)}
        \label{fig: fidelity_manifold_hoip1}
    \end{subfigure}%
    \newline
    \centering    
    \begin{subfigure}{0.5\textwidth}
        \centering
        \includegraphics[width=1\linewidth]{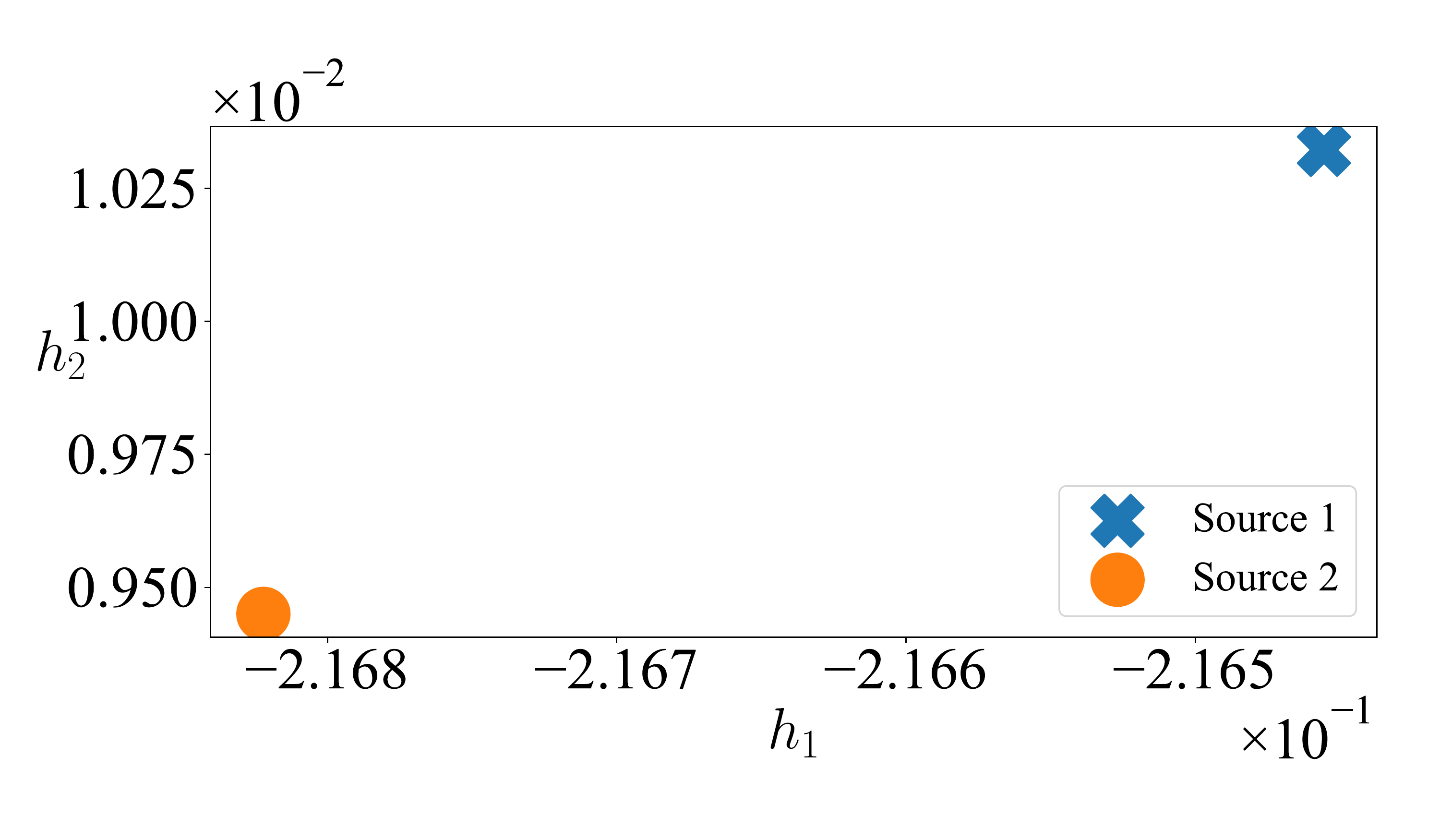}
        \caption{\nta ~(all data)}
        \label{fig: fidelity_manifold_nta2}
    \end{subfigure}%
    \begin{subfigure}{0.5\textwidth}
        \centering
        \includegraphics[width=1\linewidth]{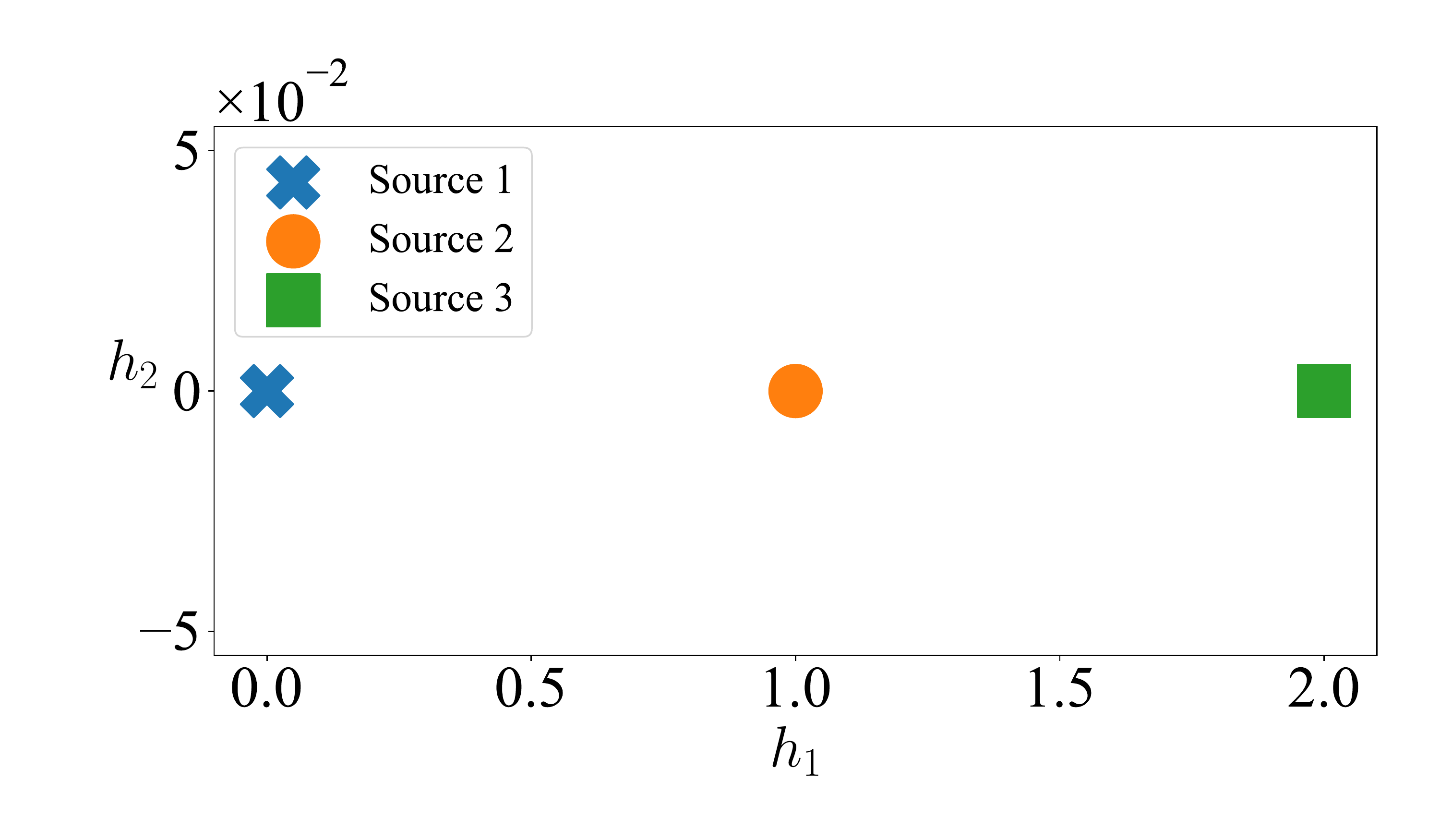}
        \caption{\hoip ~(all data)}
        \label{fig: fidelity_manifold_hoip2}
    \end{subfigure}%
    \caption{\textbf{Fidelity manifolds:} The manifolds in the top row are learned via the initial data and indicate that the LF sources should not be used in BO for \hoip ~because the latent points corresponding to them are positioned far from the point encoding the HF source. The manifolds in the second row are built using the entire MF data in each example. The similarity between the two fidelity manifolds of each example indicates that LMGP can effectively learn source-wise discrepancies via small data.
    }
    \label{fig: HOIP_Material-convergence}
\end{figure}

Per Step $0$ in Algorithm \ref{alg: MFBO-algorithm}, we train an LMGP to the initial data in each problem to determine whether any of the LF sources must be excluded from BO. As demonstrated in Figure \ref{fig: fidelity_manifold_nta1}, the LF and HF sources in \nta ~are highly correlated since their corresponding latent points are very close in the learnt fidelity manifold of LMGP. However, the latent points in Figure \ref{fig: fidelity_manifold_hoip1} are quite distant and hence we exclude both LF sources from the BO process. It is noted that $(1)$ we provide these manifolds for a randomly selected repetition in each example since they insignificantly change across the repetitions (most changes are due to rotation and translation of all the latent points which do not affect the relative distances), and $(2)$ even though small initial data is used in training the LMGPs, the resulting manifolds provide trustworthy representations of the relative fidelities. To test this second point, we fit an LMGP to the entire data in each example and visualize the resulting manifolds, see Figures \ref{fig: fidelity_manifold_nta2} and \ref{fig: fidelity_manifold_hoip2}. As it can be observed, while the manifolds do not match exactly, the relative distances between the latent points are similar. 

\begin{figure}[!t]
    \centering
    \begin{subfigure}{.5\textwidth}
        \centering
        \includegraphics[width=1\linewidth]{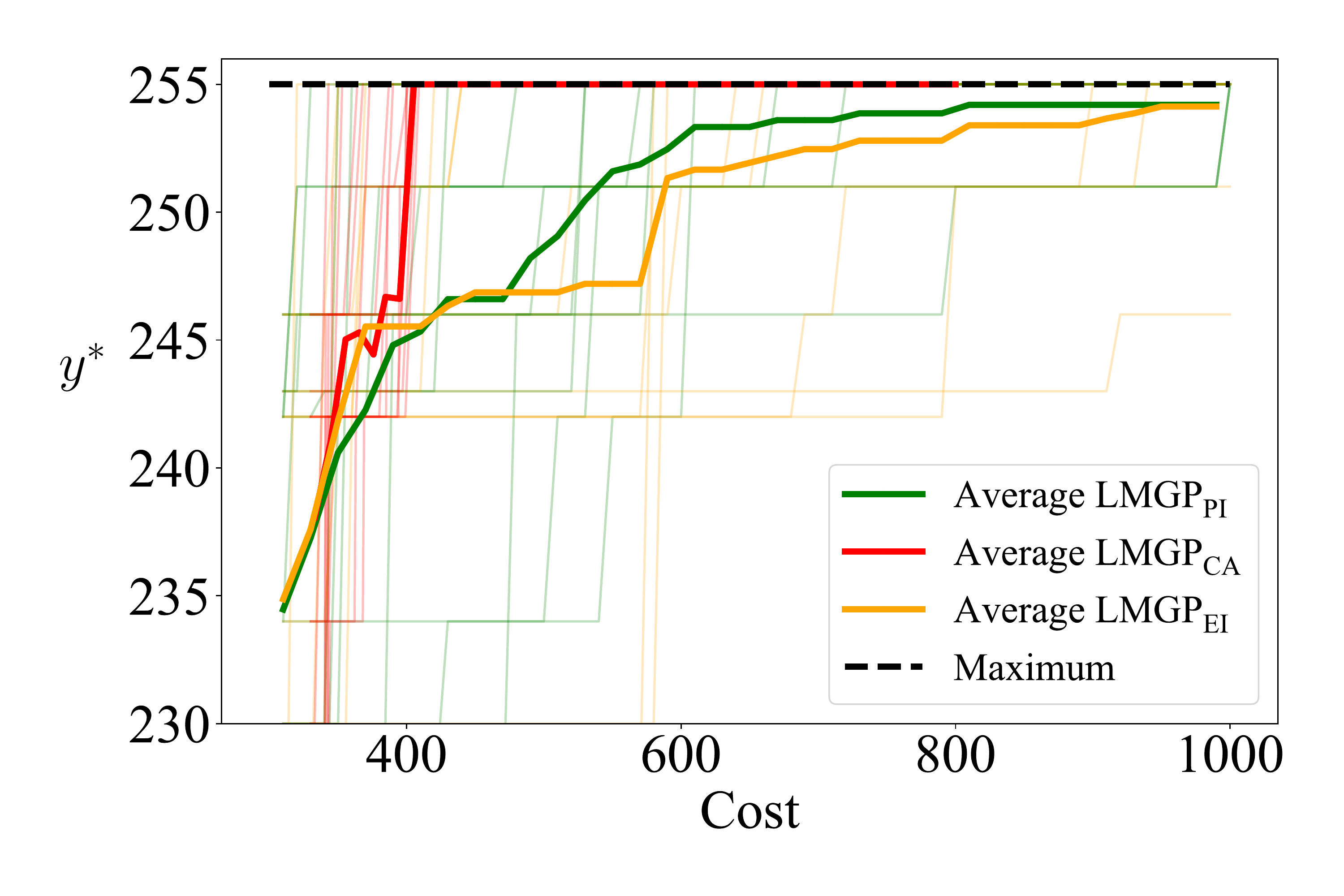}
        \caption{\nta}
        \label{fig: materila_convergence}
    \end{subfigure}%
    \begin{subfigure}{.5\textwidth}
    \centering
        \includegraphics[width=1\linewidth]{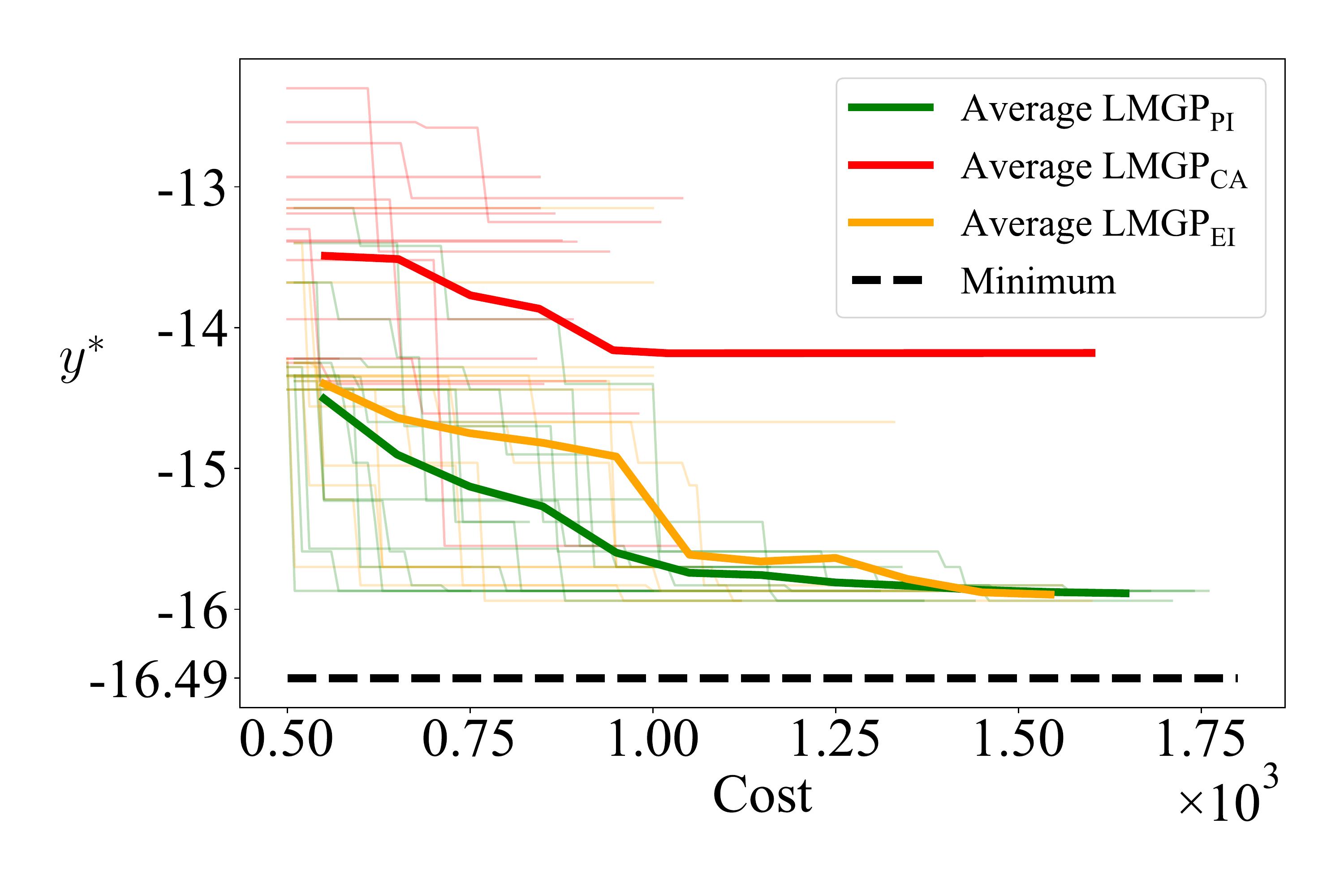}
        \caption{\hoip}
        \label{fig: HOIP convergnece}
    \end{subfigure}
    \caption{\textbf{Convergence history:} The plots illustrate the best HF sample (i.e., $y_l^*$ in Eq. \ref{eq: HF-AF}) found by each method as a function of sampling costs accumulated during the BO iterations (the cost of initial data is included). As expected, \LMGP ~outperforms the single-fidelity methods only in \nta ~since the LF sources of \hoip ~have major discrepancies. The solid thick curves indicate the average behavior across the $20$ repetitions.}
    \label{fig: HOIP_Material-convergence}
\end{figure}

Figure \ref{fig: HOIP_Material-convergence} summarizes the convergence histories by tracking the best HF estimate found by each method (i.e., $y_l^*$ in Eq. \ref{eq: HF-AF}) as a function of the accumulated costs. As expected, \LMGP ~outperforms the two SF methods in \nta ~but not in ~\hoip. In Figure \ref{fig: materila_convergence} we observe that \MEI ~and \MPI ~cannot find the optimum compound before the convergence criteria terminate the optimization process. 
However, these two methods perform quite well in \hoip ~and converge to a value that is very close to the ground truth (the small difference can be eliminated by relaxing the convergence metrics). As expected, \LMGP ~finds a sub-optimal compound in \hoip ~since the highly biased LF sources steer the search process in the wrong direction. 

\begin{figure}[!b]
    \centering
    \begin{subfigure}{.5\textwidth}
        \centering
        \includegraphics[width=1\linewidth]{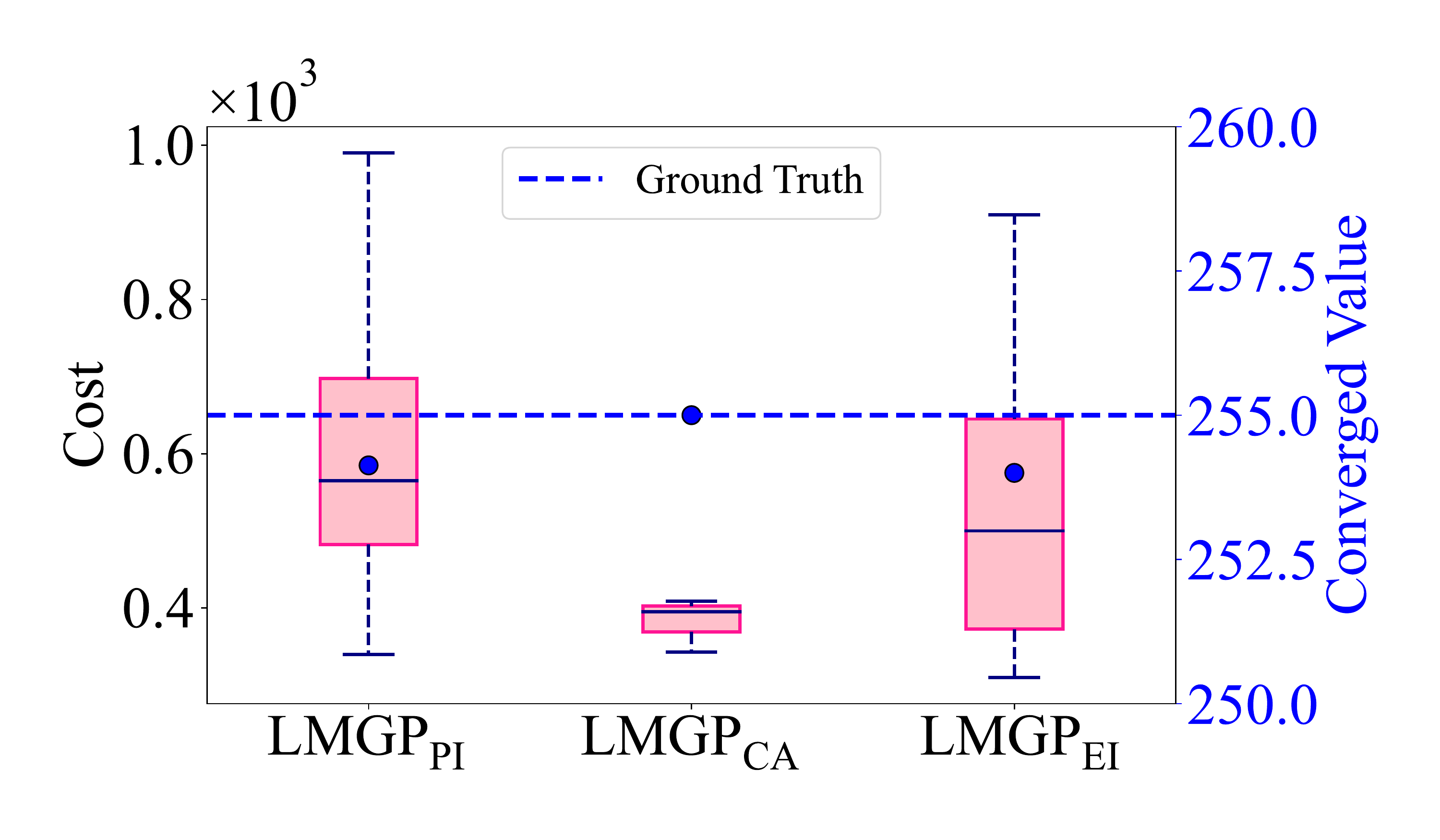}
        \caption{\nta}
        \label{fig: materila_box-cost}
    \end{subfigure}%
    \begin{subfigure}{.5\textwidth}
        \centering
        \includegraphics[width=1\linewidth]{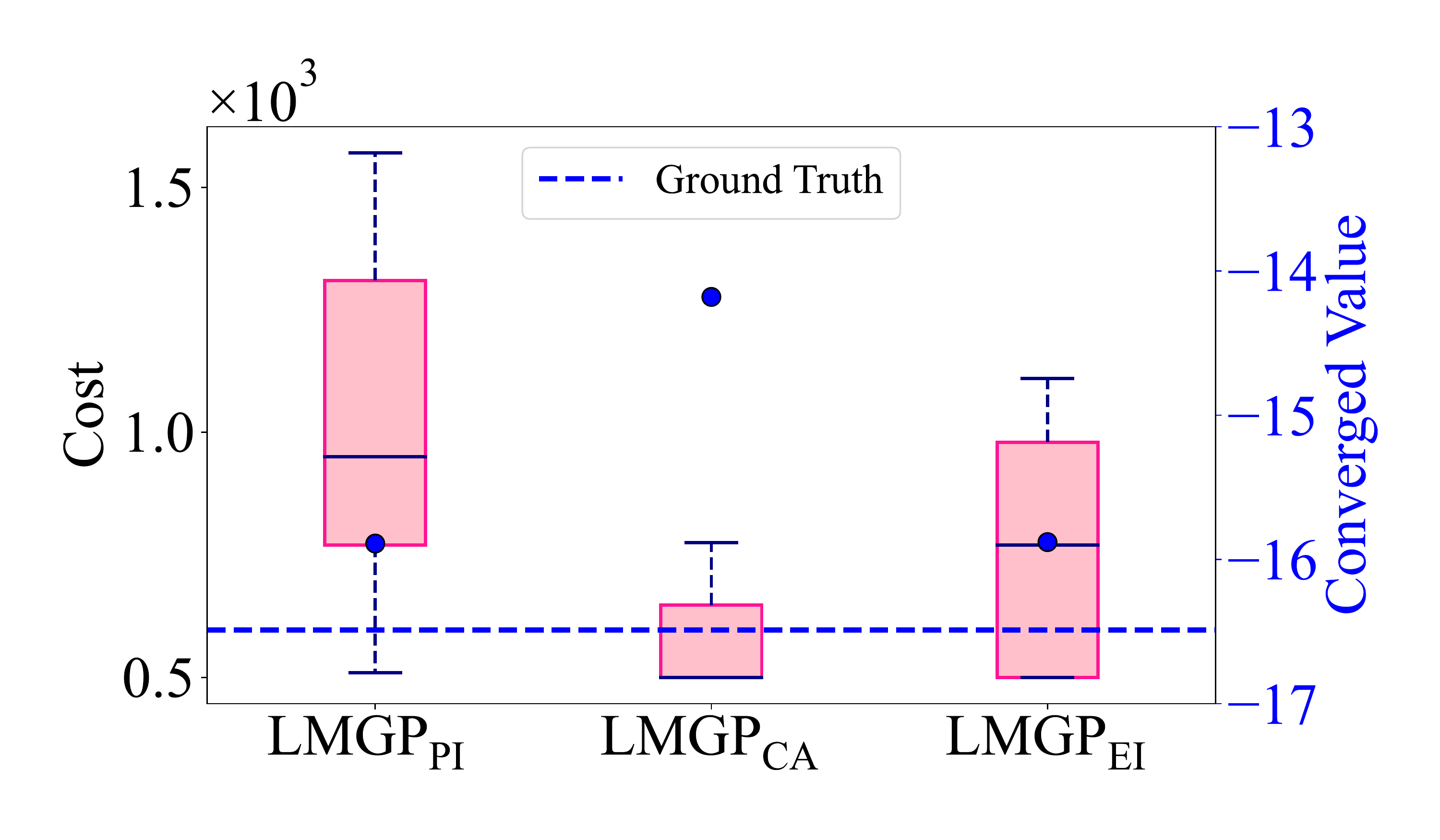}
        \caption{\hoip}
        \label{fig: HOIP-box-cost}
    \end{subfigure}
    \caption{\textbf{Accumulated costs before improvements plateau:} The box-plots illustrate the accumulated costs up to and including the iteration at which the best HF sample is first obtained (i.e., these box-plots do not consider termination criteria). On the right axis, the converged solution (averaged across the $20$ repetitions) and ground truth are demonstrated via, respectively, the blue marker and the horizontal dashed line.}
    \label{fig: HOIP_Material-box-cost}
\end{figure}

Similar to Section \ref{sec: analytical-examples} we also provide the accumulated cost up to and including the iteration at which each method finds its best compound (which may not correspond to the ground truth) in each example, see Figure \ref{fig: HOIP_Material-box-cost}. In the case of \nta, \LMGP ~outperforms both \MEI ~and \MPI ~in terms of both accuracy (i.e., finding the ground truth - compare the blue dots to the horizontal dashed line) and consistency (i.e., showing small variations across the repetitions - compare the box heights). In the case of \hoip, however, \LMGP ~provides lower accuracy than the SF methods since it is using highly biased LF sources. Even though \LMGP ~is more robust to variations in the initial data, the lower accuracy does not justify its use for \hoip. 

\begin{figure}[!b]
    \centering
    \begin{subfigure}{.5\textwidth}
        \centering
        \includegraphics[width=1\linewidth]{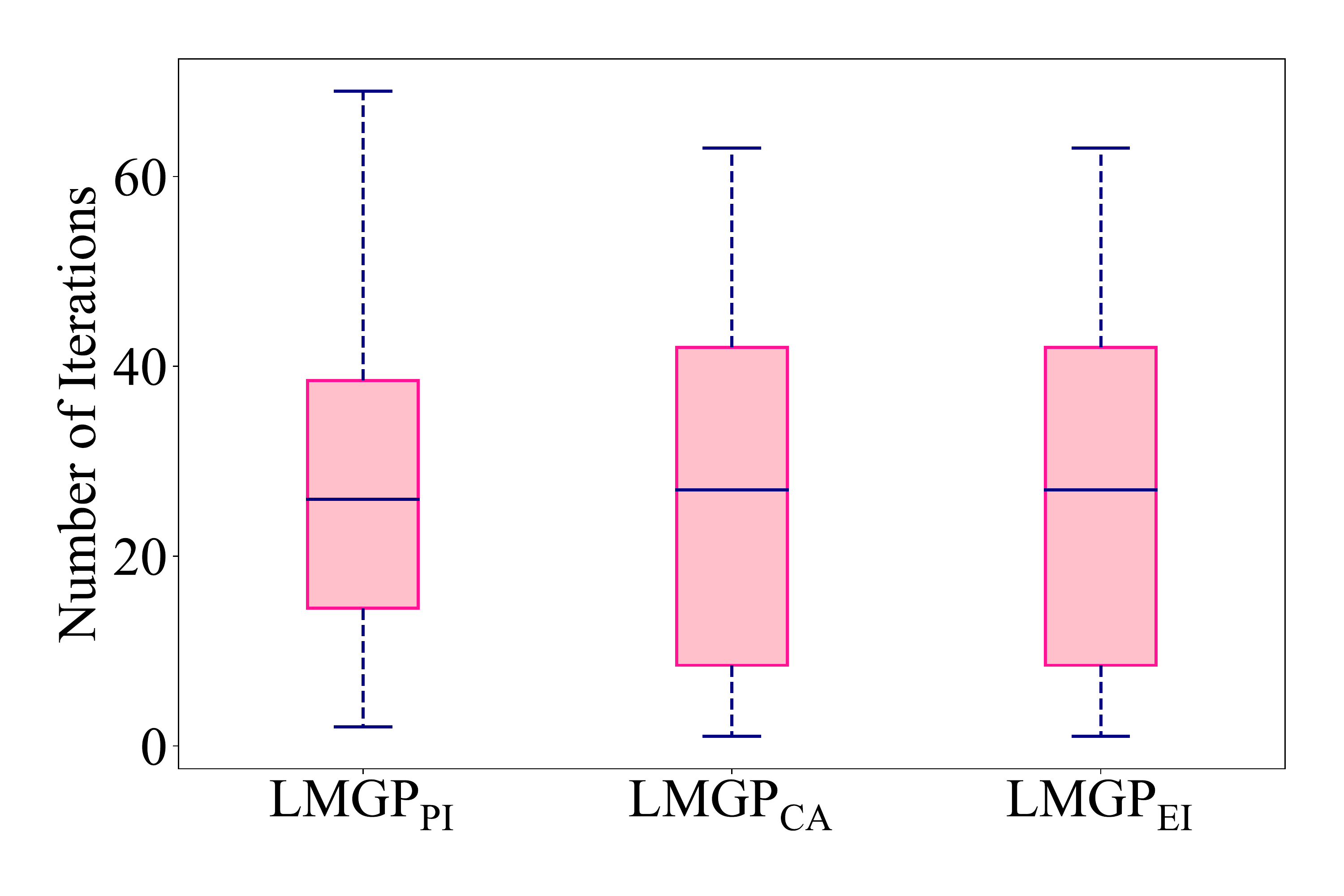}
        \caption{\nta}
        \label{fig: materila_box-iteration}
    \end{subfigure}%
    \begin{subfigure}{.5\textwidth}
        \centering
        \includegraphics[width=1\linewidth]{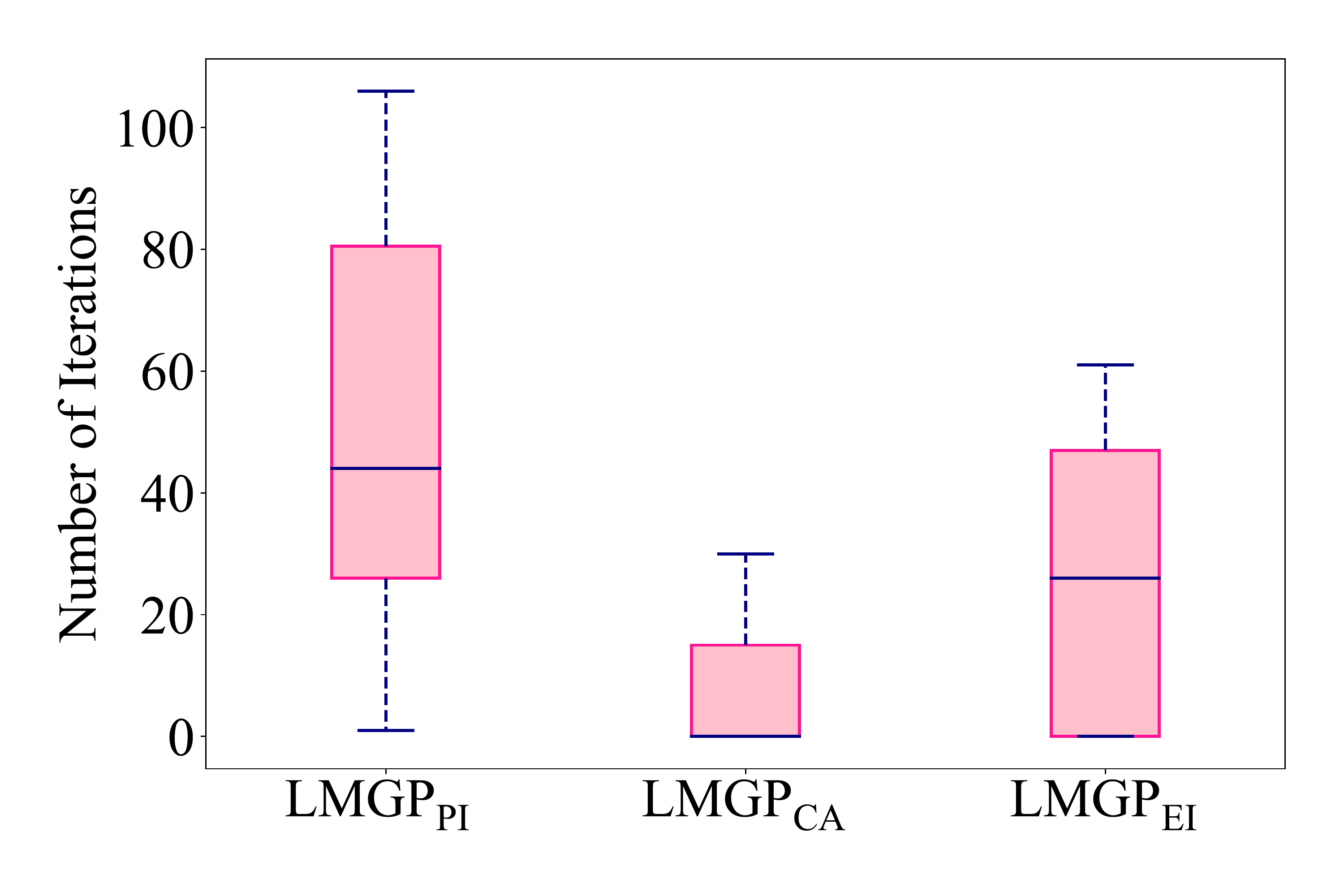}
        \caption{\hoip}
        \label{fig: HOIP-box-iter}
    \end{subfigure}
    \newline
    \centering
    \begin{subfigure}{.5\textwidth}
        \centering
        \includegraphics[width=1\linewidth]{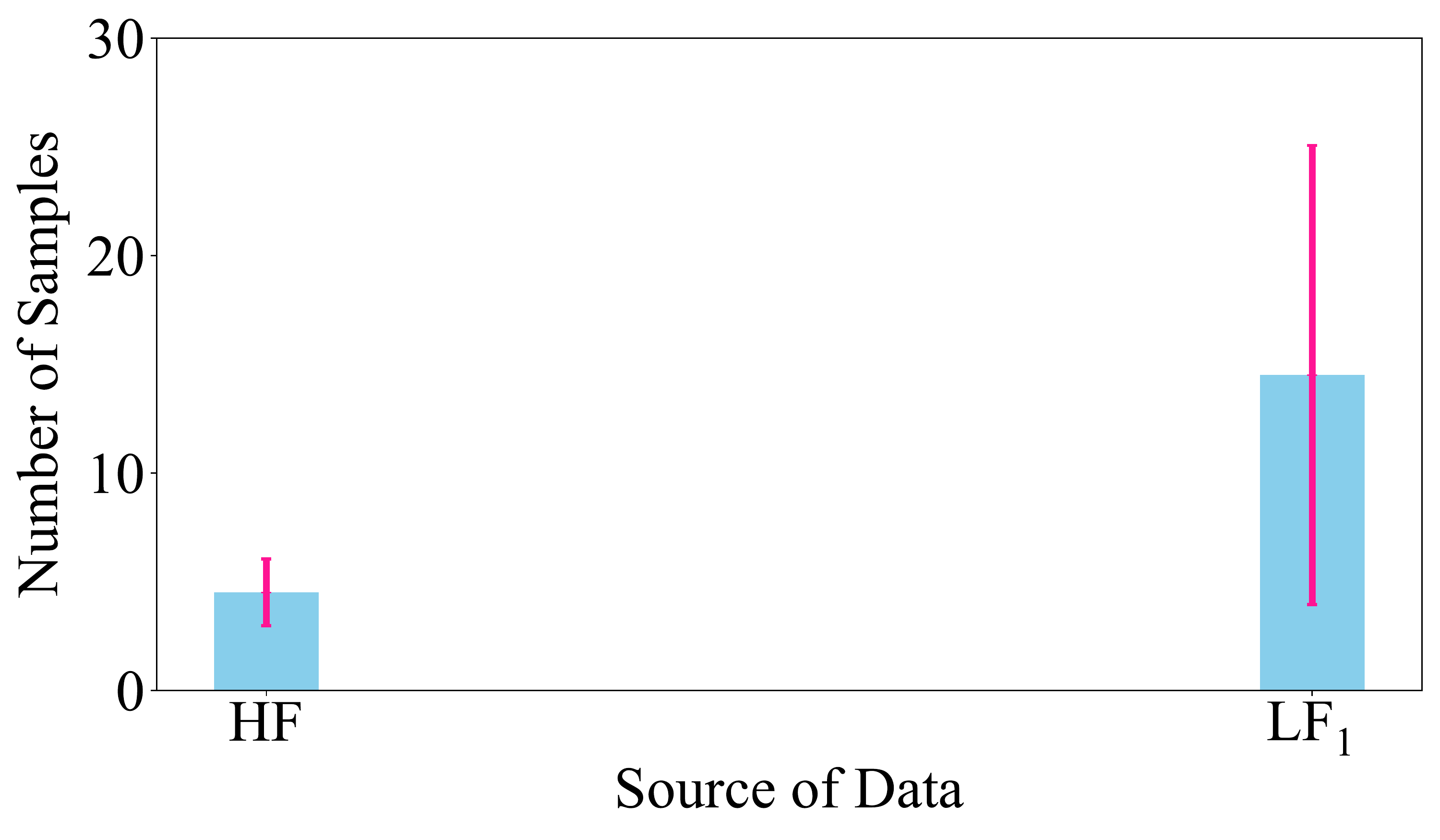}
        \caption{\nta}
        \label{fig: error-bar-materail}
    \end{subfigure}%
    \begin{subfigure}{.5\textwidth}
        \centering
        \includegraphics[width=1\linewidth]{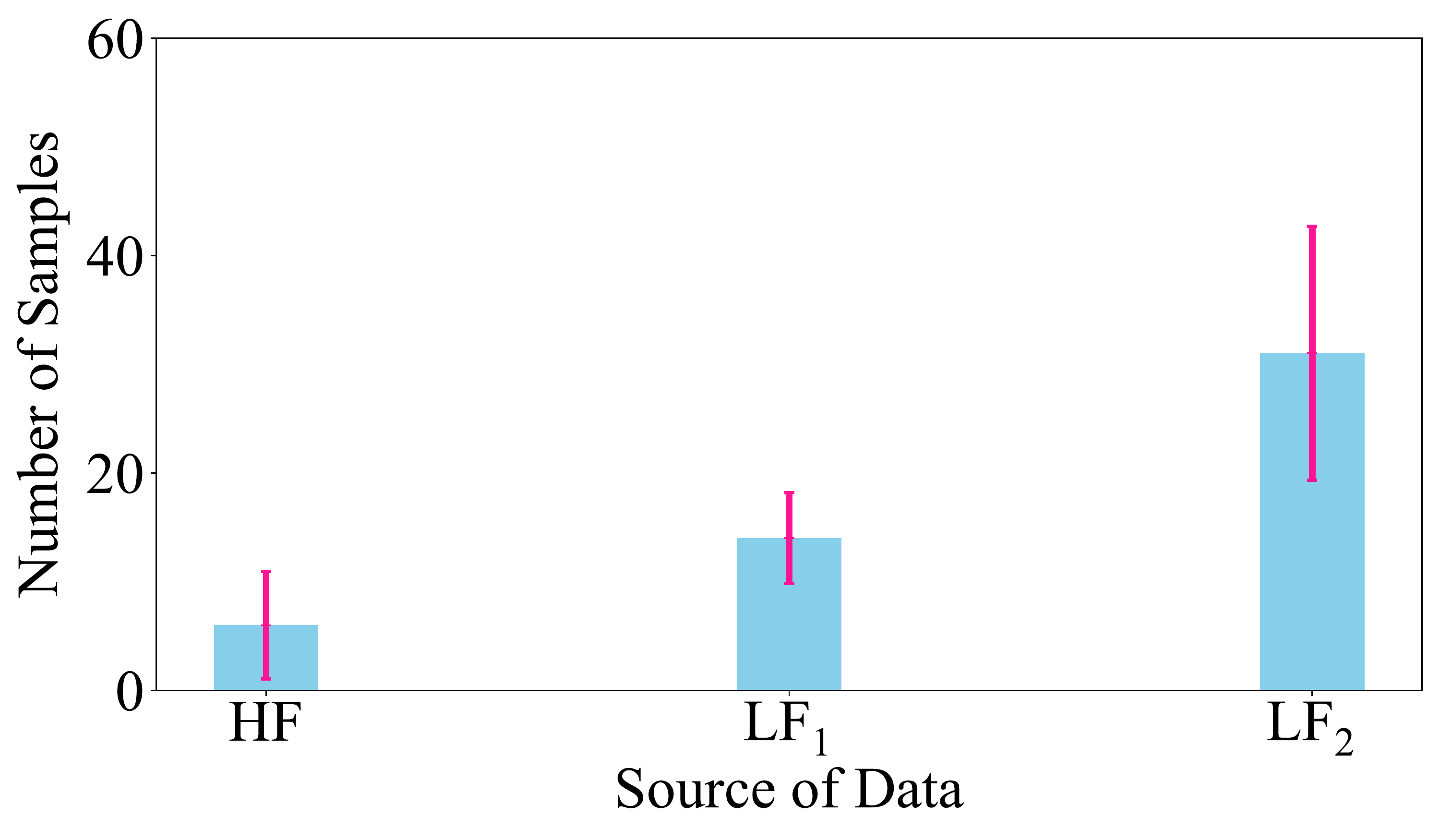}
        \caption{\hoip}
        \label{fig: error-bar-HOIP}
    \end{subfigure}  
    \newline
    \begin{subfigure}{.75\textwidth}
        \centering
        \includegraphics[width=1\linewidth]{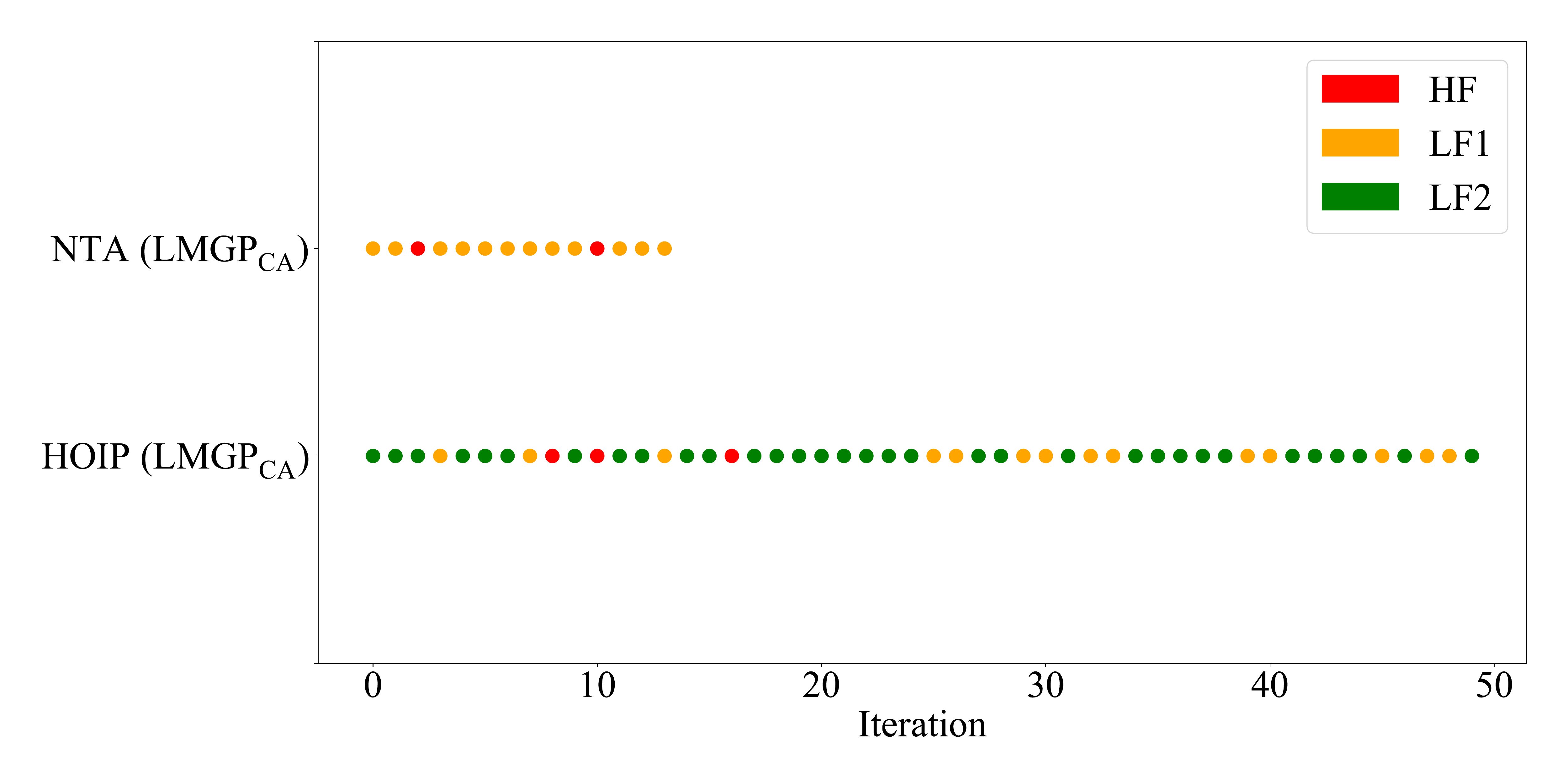}
        \caption{\hoip}
        \label{fig: sampling-order-realworld-data}
    \end{subfigure}      
    \caption{\textbf{Sampling behavior of single- and multi-fidelity BO:} In these examples \LMGP ~takes no more iterations (i.e., total number of additional samples of any fidelity) than either \MPI ~or \MEI. In both examples, most of these samples are from LF datasets which is desirable in MF BO as long as the LF sources are sufficiently correlated with the HF source.}
    \label{fig: HOIP_Material-box-iter}
\end{figure}

We now investigate the resource allocation behavior of \LMGP. As shown in Figures \ref{fig: materila_box-iteration} and \ref{fig: HOIP-box-iter}, \LMGP ~takes equal or fewer iterations to converge (note that since one sample is taken per iteration, this means that \LMGP ~takes fewer overall samples). In the case of \nta, this behavior is desirable especially since most samples are taken from the LF source which is cheaper to query, see Figure \ref{fig: error-bar-materail}. However, in the case of \hoip, this seemingly desirable behavior results in convergence to an incorrect solution. Hence, we emphasize the importance of Step $0$ in Algorithm \ref{alg: MFBO-algorithm}: while \LMGP ~can effectively allocate resources based on the initial dataset sizes and data collection costs (see Figures \ref{fig: error-bar-HOIP} and \ref{fig: sampling-order-realworld-data}), highly biased LF sources can steer the search in the wrong direction and, in turn, result in convergence to an incorrect solution.

    \section {Conclusion} \label{sec: conclusion}
We introduce a multi-fidelity cost-aware framework for Bayesian optimization of expensive black-box functions. Compared to single-source BO, our framework provides improved accuracy and convergence rate by leveraging inexpensive LF sources during the optimization. Unlike existing MF BO techniques, our method accommodates an arbitrary number of LF sources and can effectively balance exploration and exploitation regarding both the search space and source utilization. We demonstrate these benefits on both analytic and engineering examples and argue that they are the results of our new acquisition function as well as integrating LMGPs with BO.

One of the major outcomes of our work is determining (only via the initial data) if using LF sources in BO improves the performance. Currently, we make this decision by inspecting the learnt fidelity manifold of LMGP: if the point representing an LF source is far from the point which encodes the HF source, then that LF source should not be used in MF BO. This distance is directly related to the global correlation between an LF and the HF sources and we use this relation to judge whether the discrepancy is large enough. While this simple approach works quite well, it may provide sub-optimal results and hence we plan to improve it in two major directions. Firstly, we envision developing a local metric which enables LF sources to contribute to BO even if they are only correlated with the HF source on a small portion of the search space. Secondly, we plan to integrate the fidelity metrics with the AFs to scale the information values based on the sample fidelity. With these additions, all LF sources are kept in the loop since they may provide locally useful predictions.  

Our new AF does not have any calibration parameters but one can certainty scale its individual components to prioritize (based on, e.g., prior knowledge) sampling from specific sources. There is also potential in designing new utility functions that, in addition to (in lieu of) expected and probability of improvement, are inspired by other AFs such as upper confidence bound.
The examples of this papers do not explore these options since we observe high performance (which is much better than the competing methods). 
In addition, in our studies we use a very simple mechanism for encoding the fidelity via LMGP and assume the data collection costs are given and fixed but these choices and parameters can be adjusted based on the application. We plan to study these directions in our future works. 

\section*{Acknowledgments}
We appreciate the support from National Science Foundation (award numbers OAC-2211908 and OAC-2103708) and the Early Career Faculty grant from NASA’s Space Technology Research Grants Program (award number 80NSSC21K1809).
    \section*{Code and Data Availability} \label{sec: availability}
The datasets corresponding to analytic and engineering examples are available on \href{https://gitlab.com/S3anaz/multi-fidelity-cost-aware-bayesian-optimization/-/tree/main/data}{on this GitLab repository}.
The source codes for reproducing the results of \BOT ~are also available \href{https://gitlab.com/S3anaz/multi-fidelity-cost-aware-bayesian-optimization/-/tree/main/Notebooks}{here}.
 
\appendix
\addcontentsline{toc}{section}{Appendices}
\section*{Appendices}

\setcounter{equation}{0}
\renewcommand{\theequation}{\thesection-\arabic{equation}}

\section{Formulation of EI and PI} \label{sec: appendix_PI}
We first derive the AF for PI and then follow a similar procedure for EI. We insert PI's utility function in Eq. \ref{eq: general-AF}:
\begin{equation} 
    \begin{split}
        \alpha_{\mathrm{PI}}(\boldsymbol{x})=\mathbb{E}[I(\boldsymbol{x}) \mid \mathcal{D}]=\int_{-\infty}^{\infty} \operatorname{Pr}(I(\boldsymbol{x})) I(\boldsymbol{x})\mathrm{d}  x
    \end{split}
    \label{eq: PI-utiliy-defenition}
\end{equation}
\noindent Eq. \ref{eq: PI-improvement} demonstrates that $I(\boldsymbol{x})$ is zero for $y(\boldsymbol{x}) < y^*$, so:
\begin{equation} 
    \begin{split}
        \alpha_{\mathrm{PI}}(\boldsymbol{x})=\mathbb{E}[I(\boldsymbol{x}) \mid \mathcal{D}]=\int_{y^*}^{\infty} \operatorname{Pr}(I(\boldsymbol{x})>0) \mathrm{d}  y 
    \end{split}
    \label{eq: PI-integral}
\end{equation}
\noindent Hence, to calculate $\alpha_{PI}$ we only need to find $\operatorname{Pr}(I(\boldsymbol{x})>0)$ which is a function of the random variable $y(\boldsymbol{x})$.
In a GP, the response $y(\boldsymbol{x})$ follows a normal distribution with mean $\mu{(\boldsymbol{x})}$ and variance $\sigma^2(\boldsymbol{x})$: 
\begin{equation} 
    \begin{split}
        y(\boldsymbol{x}) \sim \mathcal{N}\left(\mu(\boldsymbol{x}), \sigma^2(\boldsymbol{x})\right)
    \end{split}
    \label{eq: PI-f(x)}
\end{equation}
\noindent We now apply the reparameterization trick to $y(\boldsymbol{x})$ to calculate PI. Considering $z \sim \mathcal{N}(0,1) \text {, then } y(\boldsymbol{x})=\mu(\boldsymbol{x})+\sigma(\boldsymbol{x}) z$ is a normal distribution with mean $\mu{\boldsymbol{(x)}}$ and variance $\sigma^2(\boldsymbol{x})$. Then:
\begin{equation} 
    \begin{split}
        \operatorname{Pr}(I(\boldsymbol{x})>0) \Leftrightarrow \operatorname{Pr}(y^*<y(\boldsymbol{x}))=\operatorname{Pr}(\frac{y^*-\mu(\boldsymbol{x})}{\sigma(\boldsymbol{x})}<\frac{y(\boldsymbol{x})-\mu(\boldsymbol{x})}{\sigma(\boldsymbol{x})}) 
    \end{split}
    \label{eq: PI-probability}
\end{equation}
\noindent defining $z_0=\frac{y^*-\mu(\boldsymbol{x})}{\sigma(\boldsymbol{x})}$ which follows $\mathcal{N}\left(0, \sigma^2\right)$ simplifies the above as: 
\begin{equation} 
    \begin{split}
        \operatorname{Pr}(\frac{y^*-\mu(\boldsymbol{x})}{\sigma(\boldsymbol{x})}<z)=1-\operatorname{Pr}(z \leq \frac{y^*-\mu(\boldsymbol{x})}{\sigma(\boldsymbol{x})})=1-\Phi(z)=\Phi(-z)=\Phi(\frac{\mu(\boldsymbol{x})-y^*}{\sigma(\boldsymbol{x})})
    \end{split}
    \label{eq-app: PI-AF}
\end{equation}
\noindent where $\Phi$ is the CDF of the standard normal variable \cite{jones2001taxonomy, gutmann2001radial}.

In the case of EI, we follow similarly and use the reparameterization trick to define $y(\boldsymbol{x})=\mu(\boldsymbol{x})+\sigma(\boldsymbol{x}) z$ to rewrite Eq. \ref{eq: EI-utility} as $I(\boldsymbol{x})=\mu(\boldsymbol{x})+\sigma(\boldsymbol{x}) z-y^*$ where $z$ is the standard normal random variable. We now insert this utility function into Eq. \ref{eq: general-AF}:
\begin{equation} 
    \begin{split}
        \alpha_{EI}(\boldsymbol{x}) \equiv \mathbb{E}[I(\boldsymbol{x})|D]=\int_{-\infty}^{\infty} \max(\mu(\boldsymbol{x})+\sigma(\boldsymbol{x}) z-y^*,0) \phi{(z)} \mathrm{d} z
    \end{split}
    \label{eq: EI-expectation}
\end{equation}
\noindent To  eliminate the $max$ operator and simplify the integration, we divide $\mu(\boldsymbol{x})+\sigma(\boldsymbol{x}) z-y^*$ into two negative and positive components by finding the switch point:
\begin{equation} 
    \begin{split}
        y(\boldsymbol{x})=y^* \Rightarrow \mu(\boldsymbol{x})+\sigma(\boldsymbol{x}) z=y^* \Rightarrow z_{0}=\frac{y^*-\mu(\boldsymbol{x})}{\sigma(\boldsymbol{x})}
    \end{split}
    \label{eq: EI-switch}
\end{equation}
\noindent choosing $z_{0}$ as our switch point converts the integration to:
\begin{equation} 
    \begin{split}
        \alpha_{EI}(\boldsymbol{x})=\underbrace{\int_{-\infty}^{z_0} (\mu(\boldsymbol{x})+\sigma(\boldsymbol{x}) z-y^*)\phi(z) \mathrm{d} z}_{\text {zero since } z<z_0, I(\boldsymbol{x})=0}+\int_{z_0}^{\infty} (\mu(\boldsymbol{x})+\sigma(\boldsymbol{x}) z-y^*) \phi(z) \mathrm{d} z
    \end{split}
    \label{eq: EI-2-com-integral}
\end{equation}
\noindent substituting $\phi(z)=\frac{1}{\sqrt{2 \pi}} \exp (-z^2 / 2)$ in Eq. \ref{eq: EI-2-com-integral} the integrating provides the AF:
\begin{equation} 
    \begin{split}
  \begin{aligned}
        \alpha_{EI}(\boldsymbol{x}) &=\int_{z_0}^{\infty}\left(\mu(\boldsymbol{x})-y^*\right) \phi(z) \mathrm{d} z+\int_{z_0}^{\infty} \sigma(\boldsymbol{x}) z \frac{1}{\sqrt{2 \pi}} e^{-z^2 / 2} \mathrm{~d} z \\
        &=\left(\mu(\boldsymbol{x})-y^*\right) \underbrace{\int_{z_0}^{\infty} \phi(z) \mathrm{d} z}_{1-\Phi\left(z_0\right) \equiv 1-\mathrm{CDF}\left(z_0\right)}+\frac{\sigma(\boldsymbol{x})}{\sqrt{2 \pi}} \int_{z_0}^{\infty} z e^{-z^2 / 2} \mathrm{~d} z \\
        &=\left(\mu(\boldsymbol{x})-y^*\right)\left(1-\Phi\left(z_0\right)\right)-\frac{\sigma(\boldsymbol{x})}{\sqrt{2 \pi}} \int_{z_0}^{\infty}\left(e^{-z^2 / 2}\right)^{\prime} \mathrm{d} z \\
        &=\left(\mu(\boldsymbol{x})-y^*\right)\left(1-\Phi\left(z_0\right)\right)-\frac{\sigma(\boldsymbol{x})}{\sqrt{2 \pi}}\left[e^{-z^2 / 2}\right]_{z_0}^{\infty} \\
        &=\left(\mu(\boldsymbol{x})-y^*\right) \underbrace{\left(1-\Phi\left(z_0\right)\right)}_{\Phi\left(-z_0\right)}+\sigma(\boldsymbol{x}) \phi\left(z_0\right)
        \end{aligned}
    \end{split}
    \label{eq: EI-AF-uncomp}
\end{equation}
or:
\begin{equation} 
    \begin{split}
        \alpha_{EI}(\boldsymbol{x})=(\mu(\boldsymbol{x})-y^*)\Phi(\frac{\mu(\boldsymbol{x})-y^*}{\sigma(\boldsymbol{x})})+\sigma(\boldsymbol{x}) \phi(\frac{\mu(\boldsymbol{x})-y^*}{\sigma(\boldsymbol{x})})
    \end{split}
    \label{eq: EI-AF-compl-ap}
\end{equation}

\section{Fidelity Kernels of Single-Task Multi-Fidelity GP} \label{sec: single-GP}
\setcounter{equation}{0}
\renewcommand{\theequation}{\thesection-\arabic{equation}}

Single-task multi-fidelity GP (STGP) uses two fidelity features $(1)$ the data fidelity parameter, $x_a$, which distinguishes between different fidelity sources, and $(2)$ iteration fidelity parameter, $x_b$, which is optional and usually exists in hyperparameter tuning problem. These two features are used in $e_i(\cdot)$ which are user-defined functions that model the cross-source correlations in Eq. \ref{eq: Singletaskgp-Cov}. The formulation of these functions is as follows:
\begin{equation} 
    \begin{split}
        e_1(x_a, x^\prime_{a})=(1-x_{a})(1-x^\prime_{a})(1+x_{a} x^\prime_{a})^p
    \end{split}
    \label{eq: e1}
\end{equation}
\noindent where $p$ is the degree of polynomial (which needs to be estimated) and has a Gamma prior. $e_3$ is defined similarly but for the second fidelity:
\begin{equation} 
    \begin{split}
        e_3(x_b, x^\prime_{b})=(1-x_{b})(1-x^\prime_{b})(1+x_{b} x^\prime_{b})^p
    \end{split}
    \label{eq: e3}
\end{equation}
\noindent Finally, $e_2$ is the interaction term with four deterministic terms and one polynomial kernel: 
\begin{equation} 
    \begin{split}
        e_2([x_a, x_b]^T,[x_a^{\prime}, x_b^{\prime}]^T)=(1-x_{b})(1-x^\prime_{b})(1-x_{a})(1-x^\prime_{a})(1+[x_a, x_b]^T[x_a^{\prime}, x_b^{\prime}]^T)^p
    \end{split}
    \label{eq: e2}
\end{equation}

\section{Table of Numerical Examples} \label{sec: appendix-table}
\setcounter{equation}{0}
\renewcommand{\theequation}{\thesection-\arabic{equation}}

Table \ref{table: analytic-formulation} lists the analytic functions studied in section \ref{sec: analytical-examples}. The error of each LF source with respect to the corresponding HF source is calculated via relative root mean squared error (RRMSE):
\begin{equation} 
    RRMSE = \sqrt{\frac{(\boldsymbol{y}_l - \boldsymbol{y}_h)^T(\boldsymbol{y}_l - \boldsymbol{y}_h)}{10000 \times var(\boldsymbol{y}_h)}}
    \label{eq: rrmse}
\end{equation}
\noindent where $\boldsymbol{y}_l$ and $\boldsymbol{y}_h$ are vectors of size $10000 \times 1$ that store random samples taken from the LF and HF sources, respectively.

\begin{table}[!h]
    \setlength{\extrarowheight}{0pt}
    \setlength{\tabcolsep}{0pt}
    {\renewcommand{\arraystretch}{2}
    \centering
    \begin{tabular}{| >{\centering}m{0.3\linewidth} | >{\centering}m{0.13\linewidth} | >{\centering}m{0.37\linewidth} | >{\centering}m{0.035\linewidth} | >{\centering}m{0.1\linewidth} | c |}
        \hline
        \textbf{Name} &
        \textbf{Source ID} &
        \textbf{Formulation} &
        $n$ &
        \textbf{RRMSE} &
        \textbf{Cost} 
        \\ \hline
        \multirow{2}{*}
            \potential &
            HF &
            $ 0.6x^4-0.3 x^3-3x^2+2x$ &
            $5$ &
            $-$ &
            $1000$ 
            \\ \cdashline{2-6}
            &
            LF &
            $0.6x^4-0.3 x^3-3x^2-1.2x$ &
            $0$ &
            $1.14$ &
            $1$ 
        \\ \hline
        \multirow{2}{*}
            \rosen &
            HF &
            $(1-x_1)^2+100 (x_2- x_1^2)^2-456.3$ &
            5 &
            $-$ &
            $1000$ 
            \\ \cdashline{2-6}
            &
            LF &
            $(1-x_1)^2+100$&
            $10$ &
            $1.42$ &
            $1$ 
        \\ \hline
        \multirow{5}{*}
        \borehole &
        HF &
        $\frac{2 \pi T_{u}(H_u-H_l)}{\ln (\frac{r}{r_w})(1+\frac{2 L T_u}{\ln (\frac{r}{r w}) r_w^2 k_w}+\frac{T_{u}}{T_l})}$ &
        $5$ &
        $-$ &
        $1000$ 
        \\ \cdashline{2-6}
        &
        LF1 &
        $\frac{2 \pi  T_{\mathrm{u}} (H_{\mathrm{u}}-0.8 H_l)} {\ln (\frac{r}{r_w})(1+\frac{1 L T_{\mathrm{u}}}{\ln (\frac{r}{r_w}) r_w^2 k_w}+\frac{T_{u}}{T_l})}$ &
        $5$ &
        $4.40$ &
        $100$ 
        \\ \cdashline{2-6}
        &
        LF2 &
        $\frac{2 \pi T_u(H_u- H_l)}{\ln (\frac{r}{r_w})(1+\frac{8 L T_u}{\ln (\frac{r}{r w}) r_w^2 k_w}+0.75 \frac{T_{u}}{T_l})}$ &
        $50$ &
        $1.54$ &
        $10$ 
        \\ \cdashline{2-6}
        &
        LF3 &
        $\frac{2 \pi T_{\mathrm{u}}(1.09 H_u-H_l)}{\ln (\frac{4 r}{r_w})(1+\frac{3 L T_{u}}{\ln (\frac{r}{r_w}) r_w^2 k_w}+\frac{T_u}{T_l})}$ &
        $5$ &
        $1.30$ &
        $100$ 
        \\ \cdashline{2-6}
        &
        LF4 &
        $\frac{2 \pi T_{\mathrm{u}}(1.05 H_u-H_l)}{\ln (\frac{2 r}{r_w})(1+\frac{3 L T_{i u}}{\ln (\frac{r}{\tau w}) r_w^2 k_{\mathrm{W}}}+\frac{T_{\mathrm{u}}}{T_l})}$ &
        $50$ &
        $1.3$ &
        $10$ 
        \\ \hline
        \multirow{4}{*}
        \wing &
        HF &
        $0.36 s_w^{0.758} w_{f w}^{0.0035}(\frac{A}{\cos ^2(\Lambda)})^{0.6} q^{0.006} \times \lambda^{0.04}(\frac{100 t_c}{\cos (\Lambda)})^{-0.3}(N_z W_{d g})^{0.49}+s_w w_p$ &
        $5$ &
        $-$ &
        $1000$ 
        \\ \cdashline{2-6}
        &
        LF1 &
        $0.36 s_w^{0.758} w_{f w}^{0.0035}(\frac{A}{\cos ^2(\Lambda)})^{0.6} q^{0.006} \times \lambda^{0.04}(\frac{100 t_c}{\cos (\Lambda)})^{-0.3}(N_z W_{d g})^{0.49}+w_p$ &
        $5$ &
        $0.19$ &
        $100$ 
        \\ \cdashline{2-6}
        &
        LF2 &
        $0.36 s_w^{0.8} w_{f w}^{0.0035}(\frac{A}{\cos ^2(\Lambda)})^{0.6} q^{0.006} \times \lambda^{0.04}(\frac{100 t_c}{\cos (\Lambda)})^{-0.3}(N_z W_{d g})^{0.49}+w_p$ &
        $10$ &
        $1.14$ &
        $10$ 
        \\ \cdashline{2-6}
        &
        LF3 &
        $0.36 s_w^{0.9} w_{f w}^{0.0035}(\frac{A}{\cos ^2(\Lambda)})^{0.6} q^{0.006} \times \lambda^{0.04}(\frac{100 t_c}{\cos (\Lambda)})^{-0.3}(N_z W_{d g})^{0.49}$ &
        $50$ &
        $5.75$ &
        $1$    
        \\ \hline
    \end{tabular}
    
    \caption{\textbf{List of analytic functions}: The examples have a diverse degree of dimensionality, number of sources, and complexity. $n$ denotes the number of initial samples and the relative root mean squared error (RRMSE) of an LF source is calculated by comparing its output to that of the HF source at $10000$ random points, see Eq. \ref{eq: rrmse}. For \borehole, LF3 and LF4 become the first and second LF sources, respectively, once LMGP identifies that the listed LF1 and LF2 in this table are highly biased.}        
    \label{table: analytic-formulation}
    }
\end{table}

\section{Effect of Highly Biased Low-fidelity Sources} \label{sec: bo-5source}
\begin{figure*}[!h] 
    \centering
    \begin{subfigure}{.5\textwidth}
        \centering
        \includegraphics[width = 1\textwidth]{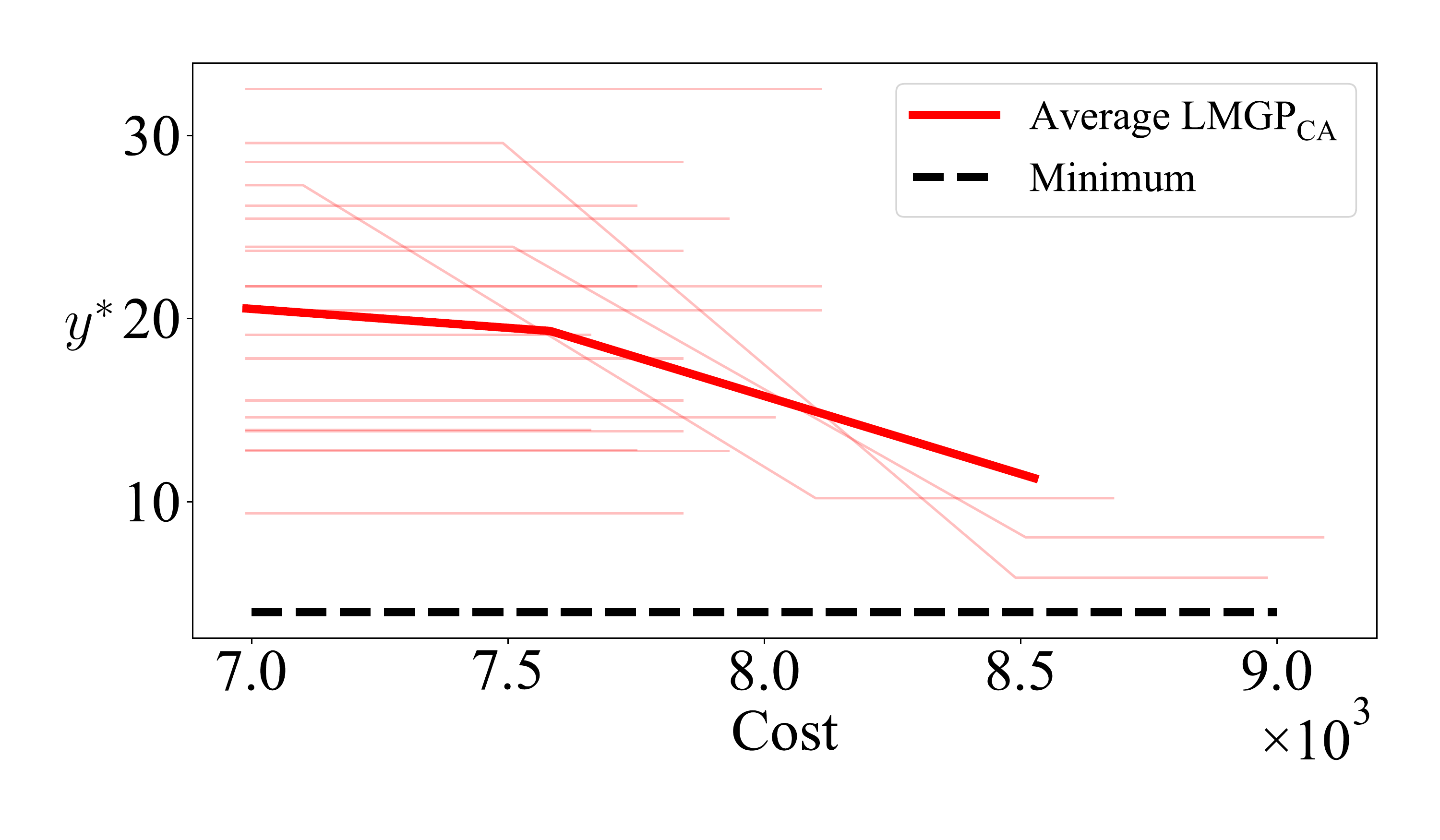}
        \caption{Convergence history}
        \label{fig: borehole-5source-convergence}
    \end{subfigure}%
    \begin{subfigure}{.5\textwidth}
        \centering
        \includegraphics[width = 1\textwidth]{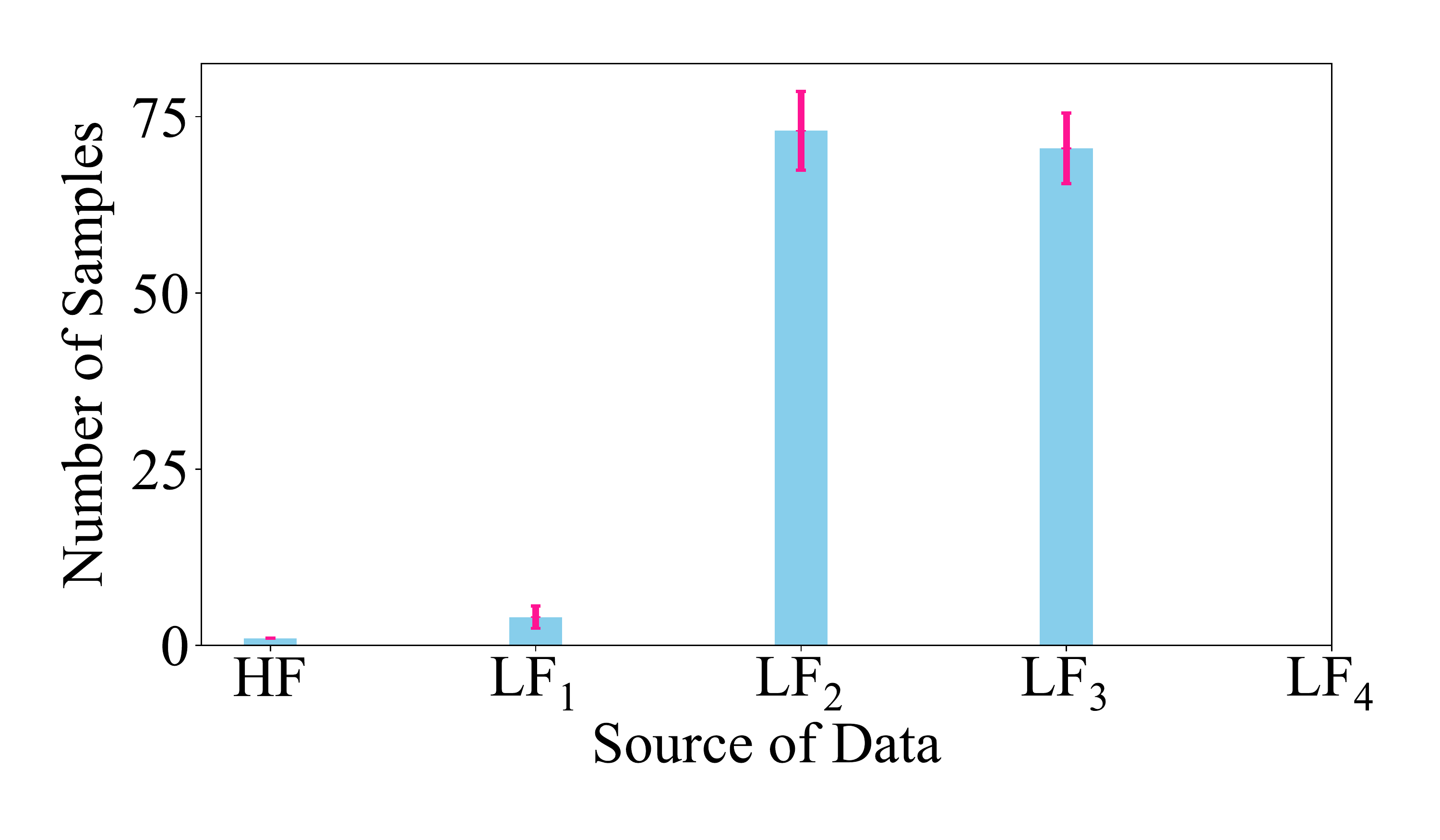}
        \caption{Frequency of source utilization}
        \label{fig: borehole-5source-sampling}
    \end{subfigure}
    \caption{\textbf{Effect of highly biased and inexpensive low-fidelity sources:} While \LMGP ~effectively samples from all sources (considering their costs and contribution to the initial data), it converges to an incorrect solution since LF1 and LF2 are highly biased. The initial data are not included in \ref{fig: borehole-5source-sampling}.}
    \label{fig: borehole-5source}
\end{figure*}

\setcounter{equation}{0}
\renewcommand{\theequation}{\thesection-\arabic{equation}}
In Section \ref{sec: analytical-examples}, we exclude two LF sources from \borehole ~due to their high discrepancy with respect to the HF source, see Figure \ref{fig: LMGP_ALL4_LS}. Below, we summarize the performance of \LMGP ~on this problem without removing these two sources. As it can be observed in Figure \ref{fig: borehole-5source-convergence}, the optimization is terminated based on the second convergence metric which caps the maximum number of iterations without improvement in the best HF sample (i.e., $y_l^*$ in Eq. \ref{eq: HF-AF}). 

\section{List of Abbreviations and Symbols} \label{sec: nomenclature}

\begin{table}[H]
\begin{tabular}{ll}
\textbf{Abbreviation} & \textbf{Explanation}                                                   \\
AF                                                 & Acquisition Function                                                   \\
BO                                                & Bayesian optimization                                                  \\
\BOT                                               & Multi-fidelity BO with BoTorch                                         \\
Co-K                                               & Co-Kriging                                                             \\
EI                                                 & Expected Improvement                                                   \\
GP                                                 & Gaussian Process                                                     \\
HF                                                 & High-fidelity                                                          \\
HOIP                                                & Hybrid Organic–Inorganic Perovskite\\                                                         
KG                                                 & Knowledge Gradient                                                     \\
LF                                                 & Low-Fidelity                                                           \\
LMGP                                                & Latent Map Gaussian Process                                          \\
\LMGP                                               & Proposed MFCA BO approach                                              \\
\MEI                                               & Single-fidelity BO whose emulator and AF are LMGP and EI, respectively \\
\MPI                                               & Single-fidelity BO whose emulator and AF are LMGP and PI, respectively \\
MF                                                 & Multi-Fidelity                                                         \\
MFCA                                               & Multi-Fidelity Cost-Aware                                              \\
MLE                                                & Maximum Likelihood Estimation                                          \\
MLEI                                               & Most Likely Expected Improvment                                        \\
MSE                                                & Mean Squared Error                                                     \\
NTA                                               & Nanolaminate Ternary Alloy                                            
\\
PI                                                 & Probability of Improvement                                             \\
RRMSE                                               & Relative Root Mean Squared Error                                       \\
SF                                                 & Single-Fidelity                                                        \\
STGP                                               & Single-Task Multi-Fidelity Gaussian  \hspace{20cm}                                                                                                             \\
                                                                   
\end{tabular}
\end{table}

\begin{table}[H]
\begin{tabular}{ll}
\textbf{Symbol}                                                         & \textbf{Description}                                                                               \\
$\boldsymbol{A}$                                                        & Rectangular matrix that maps $\boldsymbol{\zeta(t)}$ to $\boldsymbol{z(t)}$                        \\
$c(\boldsymbol{x}, \boldsymbol{x}^\prime)$                              & Covariance function                                                                                     \\

$ds$                                                                    & Number of data sources                                                                             \\
$dt$                                                                   & Dimension of categorical inputs                                                                    \\
$dz$                                                                    & Dimension of the latent map                                                                        \\
$dx$                                                                    & Dimension of numerical inputs                                                                      \\

$\boldsymbol{h}(s) = [h_1, \cdots, h_{dh}]^T$                           & Latent representation of data source $s$                                                           \\
$I(\boldsymbol{x})$                                                     & Utility function                                                                      \\
$'\boldsymbol{j}'$                                                      & Categorical vector of size ${n_j \times 1}$ whose elements are all set to $'j'$                    \\
$\mathcal{N}\left(\mu(\boldsymbol{x}), \sigma^2(\boldsymbol{x})\right)$ & Normal distribution with mean $\mu(\boldsymbol{x})$ and standard deviation $\sigma^2(\boldsymbol{x})$ \\
$n_j$                                                                   & Number of samples obtained from $s(j)$ (i.e., source $j$)                                          \\
$ l_i $                                                                 & Number of distinct levels in ${i}^{th}$ categorical input                                          \\

$\boldsymbol{R}$                                                        & Correlation matrix                                                                                 \\
$r(.,.)$                                                                & Parametric correlation function                                                                    \\
$s=\{'1', \cdots, 'ds'\}$                                               & Categorical variable whose $j^{th}$ element corresponds to data source $j$              \\
$ \boldsymbol{t}$                                                       & Categorical inputs (all except source indicator)                                                                                \\
$\boldsymbol{U}_j$                                                      & $n_j \times (dx + dt)$ feature matrix obtained from $s(j)$                                         \\
$\boldsymbol{u}$                                                        & Mixed inputs                                                                                       \\
$\boldsymbol{x}$                                                        & Input vector                                                                                       \\
$y(\boldsymbol{x})$                                                     & Output/response                                                                                    \\
$\boldsymbol{y}_j$                                                      & ${n_j \times 1}$ vector of responses obtained from $s(j)$                                          \\
$\boldsymbol{z(t)}$                                                     & Points on latent map corresponding to combination $\boldsymbol{t}$ of the categorical variables    \\

$\alpha(\boldsymbol{x})$                                                & Acquisition Function                                                                               \\
$\boldsymbol{\zeta}(\boldsymbol{t})$                                    & Unique prior vector representation of $\boldsymbol{t}$                                             \\
$\xi(\boldsymbol{x})$                                                   & Zero-mean GP                                                                                       \\

$\sigma^{2}$                                                            & Variance of process                                                                                \\
$\Phi(z)$                                                               & Cumulative density function (CDF)                                                                  \\
$\phi(z)$                                                               & Probability density function (PDF)                                                                 \\
$\boldsymbol{\Omega}$                                                   & $\operatorname{diag}(\boldsymbol{\omega})$                                                         \\
$\boldsymbol{\omega}$                                                   & Scale parameters                                                                                   \\
$\otimes$                                                               & Kronecker product \hspace{20cm}                                                                                  \\                                                                   
\end{tabular}
\end{table}

    \pagebreak    
    \printbibliography
\end{document}